\documentclass{article}

\usepackage[utf8]{inputenc} 
\usepackage[T1]{fontenc}    
\usepackage{url}            
\usepackage{booktabs}       
\usepackage{amsfonts}       
\usepackage{nicefrac}       
\usepackage{microtype}      
\usepackage{xcolor}         
\usepackage[toc,page,header]{appendix}
\usepackage{wrapfig}
\usepackage{tikz}
\usetikzlibrary{fit} 
\usepackage{subcaption}
\usepackage{marvosym}

\usetikzlibrary{bayesnet, calc, arrows.meta, bending}
\usetikzlibrary{decorations.pathreplacing,angles,quotes,shapes.multipart, calligraphy}

\usepackage{hyperref}

\hypersetup{
    colorlinks,
    linkcolor={blue!50!black},
    citecolor={blue!50!black},
    urlcolor={blue!50!black}
}

\tikzset{>=Stealth}

\usepackage{amsmath}

\usepackage[accepted]{icml2024}

\usepackage{amsmath}
\usepackage{amssymb}
\usepackage{mathtools}
\usepackage{amsthm}
\usepackage{enumerate}
\usepackage{tabularx}
\newcolumntype{L}[1]{>{\raggedright}p{#1}}

\usepackage[most]{tcolorbox}
\usepackage{titlesec} 

\usepackage[capitalize,noabbrev]{cleveref}

\usepackage{amsmath,amsfonts,bm}

\def\eqref#1{equation~\ref{#1}}

\def\1{\bm{1}}

\def\bmu{\boldsymbol{\mu}}
\def\bsigma{\boldsymbol{\sigma}}
\def \bSigma{\boldsymbol{\Sigma}}

\def\bpi{\boldsymbol{\pi}}

\def\rva{{\mathbf{a}}}

\def\rvc{{\mathbf{c}}}

\def\rvs{{\mathbf{s}}}

\def\rvv{{\mathbf{v}}}

\def\rvx{{\mathbf{x}}}
\def\rvy{{\mathbf{y}}}
\def\rvz{{\mathbf{z}}}

\def\mA{{\bm{A}}}

\def\mH{{\bm{H}}}

\def\mP{{\bm{P}}}

\def\mW{{\bm{W}}}

\DeclareMathAlphabet{\mathsfit}{\encodingdefault}{\sfdefault}{m}{sl}
\SetMathAlphabet{\mathsfit}{bold}{\encodingdefault}{\sfdefault}{bx}{n}

\def\gC{{\mathcal{C}}}

\def\gF{{\mathcal{F}}}

\def\gH{{\mathcal{H}}}

\def\gN{{\mathcal{N}}}
\def\gO{{\mathcal{O}}}
\def\gP{{\mathcal{P}}}

\def\gS{{\mathcal{S}}}
\def\gT{{\mathcal{T}}}

\def\gV{{\mathcal{V}}}

\def\gX{{\mathcal{X}}}

\theoremstyle{plain}
\newtheorem{theorem}{Theorem}[section]
\newtheorem{example}{Example}

\newtheorem{lemma}[theorem]{Lemma}

\theoremstyle{definition}
\newtheorem{definition}[theorem]{Definition}
\newtheorem{assumption}[theorem]{Assumption}
\theoremstyle{remark}
\newtheorem{remark}[theorem]{Remark}

\newenvironment{sproof}{\proof}{\endproof}

\usepackage[textsize=tiny]{todonotes}

\usepackage{amsmath}
\usepackage[inline]{enumitem}

\usepackage{booktabs}
\usepackage{multirow}
\icmltitlerunning{Identifiable Object Representations under Spatial Ambiguities}

\begin{document}
\twocolumn[
\icmltitle{Identifiable Object Representations under Spatial Ambiguities}



\icmlsetsymbol{equal}{*}

\begin{icmlauthorlist}
\icmlauthor{Avinash Kori}{yyy}
\icmlauthor{Francesca Toni}{yyy}
\icmlauthor{Ben Glocker}{yyy}
\end{icmlauthorlist}

\icmlaffiliation{yyy}{Department of Computing, Imperial College London, London, UK}

\icmlcorrespondingauthor{Avinash Kori}{a.kori21@imperial.ac.uk}

\icmlkeywords{Spatial ambiguities, object centric learning, occlusions}

\vskip 0.3in
]



\printAffiliationsAndNotice{}  

\begin{abstract}
Modular object-centric representations are essential for \emph{human-like reasoning} but are challenging to obtain under spatial ambiguities,  \emph{e.g. due to occlusions and view ambiguities}. 
However, addressing challenges presents both theoretical and practical difficulties. 
We introduce a novel multi-view probabilistic approach that aggregates view-specific slots to capture \emph{invariant content} information while simultaneously learning disentangled global \emph{viewpoint-level} information. 
Unlike prior single-view methods, our approach resolves spatial ambiguities, provides theoretical guarantees for identifiability, and requires \emph{no viewpoint annotations}. Extensive experiments on standard benchmarks and novel complex datasets validate our method's robustness and scalability.
\end{abstract}

\section{Introduction}
\label{section: introduction}

The ability to capture the notion of \emph{objectness} in learned representations is considered to be a critical aspect for developing situation-aware AI systems with human-like reasoning capabilities~\citep{scholkopf2022statistical, lake2017building}. 
Objectness can be characterised as understanding the environment from the perspective of its building blocks. These can further be divided into object-part composition \cite{hinton1979some, hinton2022represent}, which might be a potential reason why humans generalise across environments with few examples to learn from~\cite{tenenbaum2011grow}.
Recent advances in object-centric representation learning (OCL) have shown great potential in segregating objects in observed scenes \citep{locatello2020object, kori2023grounded, lowe2024rotating}.
Indeed, the goal of OCL is to enable agents to learn representations of \emph{objects} in an observed scene in the context of their environment, as opposed to learning global representations as in the case of traditional generative models such as variational auto-encoders \citep{kingma2013auto}. 
OCL approaches enable agents to learn spatially disentangled representations, which is an important step in compositional scene generation \citep{bengio2013representation, lake2017building, battaglia2018relational, greff2020binding} and understanding of causal (and physical) interactions between the objects \citep{marcus2003algebraic, gerstenberg2021counterfactual, gopnik2004theory}.

\begin{figure}[!t]
    \centering
    \subfloat[]{\includegraphics[width=.74\columnwidth]{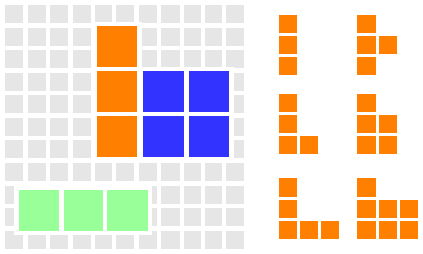}}\hfill
    \subfloat[]{\includegraphics[width=.25\columnwidth]{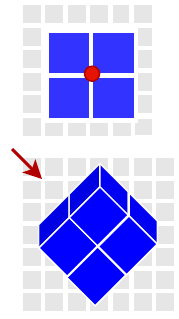}}\hfill
    \caption{
    \textbf{(a) Occlusion Ambiguity:} 
    the \textcolor{orange}{orange} object, which is occluded by the \textcolor{blue}{blue} object, could be any of the six plausible objects shown on the right. \textbf{(b) View Ambiguity:} 
    the \textcolor{blue}{blue} object is observed from two different viewpoints (represented with a \textcolor{red}{red} arrow and a dot), leading to a change in its overall shape. 
    In general, identifiable representations resolve ambiguities by determining the most plausible object under occlusion and correct object properties in case of view transformation by leveraging information from multiple viewpoints.
    }
    \label{fig:motivation}
\end{figure}

\begin{figure}[!t]
    \centering
    \includegraphics[width=0.8\columnwidth, trim={1cm 0 2cm 0},clip]{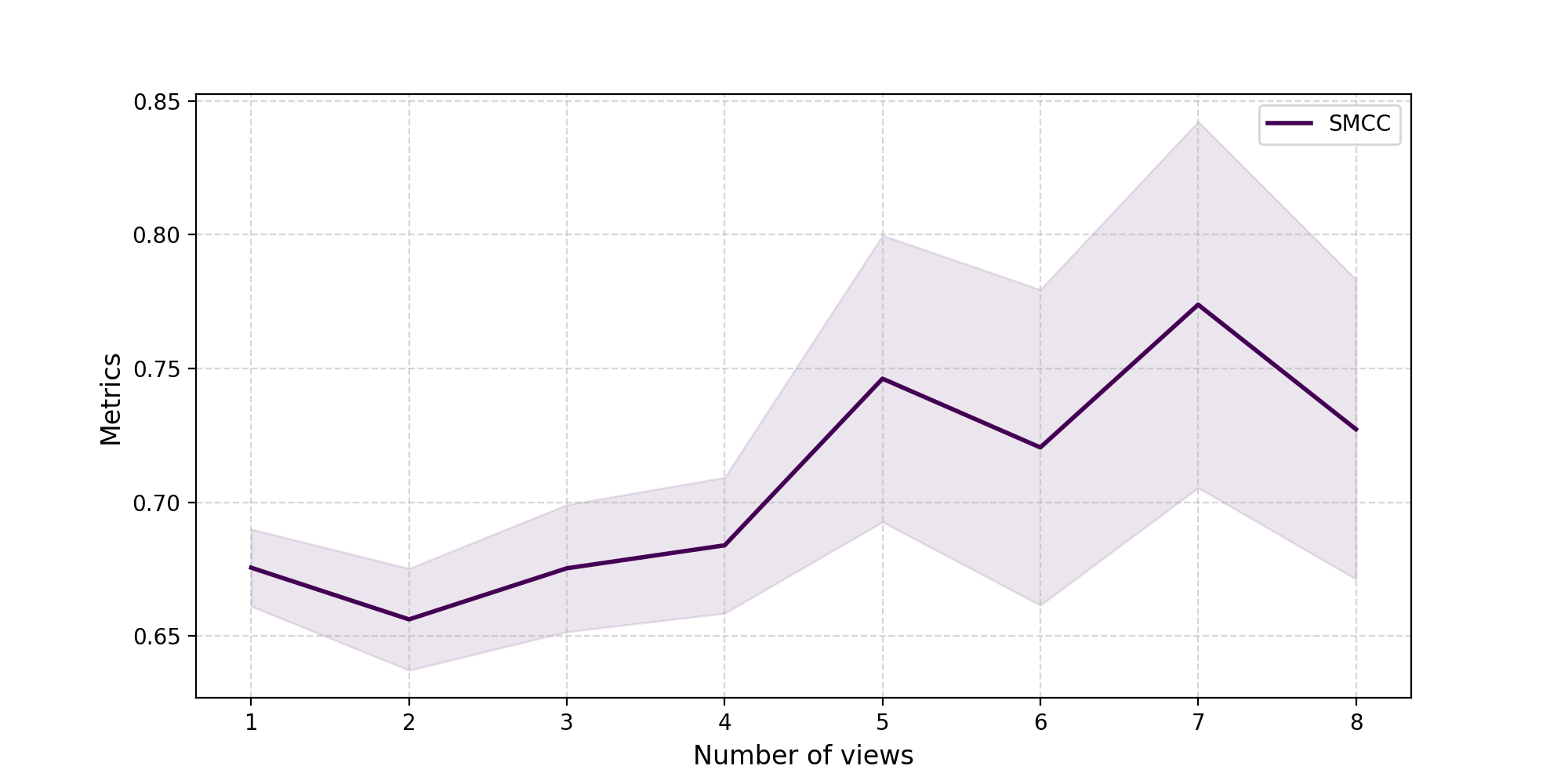}
    \caption{
    Identifiability across a number of views measured with Slot Mean Correlation Coefficient (SMCC). 
    }
    \label{fig:motivation2}
\end{figure}

Recent progress in OCL has been limited to learning scene representations from single-viewpoints \citep{locatello2020object, engelcke2021genesis, singh2021illiterate, kori2023grounded, chang2022object, seitzer2022bridging, lowe2024rotating}.
Although these approaches can learn meaningful object-specific representations, they encounter significant challenges stemming from spatial ambiguities such as occlusion and view ambiguities (see Fig.~\ref{fig:motivation} for examples). 
Additionally, while it has been hypothesised that these models learn un-occluded object representations even in the case of occlusions.
Learning from a single viewpoint fails to capture effective object representations, due to the presence of multiple plausibilities of partially or fully occluded objects and the effects of view transformations, as demonstrated in Fig. \ref{fig:motivation} and highlighted by the results in  Fig. \ref{fig:motivation2} (we will revisit these results later in section \ref{sec:experiments}).
Another example of 
spatial 
ambiguities can be observed in Fig. \ref{fig:overview}, where object $\gO_4$ in $\rvx^1$ and $\rvx^2$ can be interpreted as a cube, but only after considering $\rvx^3$ we can conclude that being a pyramid. 

A handful of approaches, including \textsc{MulMON}\citep{li2020learning}, \textsc{DyMON}\citep{li2021objectdymon}, \textsc{OCLOC}\citep{yuan2024unsupervisedocloc}, have considered multiple viewpoints for extracting object representations. 
Additionally, methods 
such as~\citep{liu2025slotlifter, chen2021roots, luo2024unsuperviseduocf} effectively use \textsc{NeRF}\citep{mildenhall2021nerf} for constructing a 3D environment from multi-viewpoint images, where the occlusions are addressed by construction. 
Among these methods, \textsc{MulMON}, \textsc{DyMON}, and all \textsc{NeRF} based approaches assume that the
viewpoint annotations are known, which 
simplifies the problem of learning to disentangle object representations conditioned on viewpoint information.

The problem setting in this work aligns with \textsc{OCLOC}, 
in that, our aim is to learn invariant object representations while simultaneously learning global view information with respect to an \emph{implicit global coordinate frame}. 
This eliminates the requirement for paired viewpoint-image data. While \textsc{OCLOC} introduces an innovative approach for learning global view information independently of the scene, its primary focus is on achieving \emph{object-consistency} unconditional to views rather than explicitly learning view-invariant object representations.
Additionally, learning global unconditional view representations does not guarantee learning identifiable view/object representations, which was not studied for \textsc{OCLOC}.  
In this work, we 
provide a 
novel model, where object representations satisfy view-invariance and view representations satisfy \emph{approximate equivariance} properties
, allowing us to exploit objects' inherent geometry and semantics to establish correspondences across views.

In single-view OCL, \citet{kori2024identifiable, brady2023provably, lachapelle2023additive} make an effort in rigorously formalising the underpinning, explicit and implicit assumptions and provide conditions under which models result in learning identifiable slot representations, leaving out ambiguous scenarios.  
Unlike them, our approach resolves spatial ambiguities, provides theoretical guarantees for identifiability, and requires no viewpoint annotations. 
To the best of our knowledge, this is the first work addressing explicit formalisations of assumptions and theory required for achieving \emph{identifiable} object representations under occlusions with multi-view observational data.
To this end, we make use of the spatial Gaussian mixture models(GMM) in latent distribution across viewpoints to encourage identifiability without additional auxiliary data.
Our main contributions in this work can be summarised as follows:
\begin{enumerate}[label=(\roman*)]
    \item We propose a probabilistic slot attention variant, \emph{View-Invariant Slot Attention (VISA)}
    for learning identifiable object-centric representations from multiple viewpoints, resolving spacial ambiguities such as 
    occlusions and view ambiguities (Section~\ref{sec:formalism}).
    \item We prove that our object-centric representations are identifiable in the case of partial or full occlusions without additional view information up to an equivalence relation with a mixture model specification (Section~\ref{sec:identifiability}).
    \item We provide conclusive evidence of our identifiability results, including visual verification on synthetic datasets; we also demonstrate the scalability of the proposed method on two new, carefully designed complex datasets \textsc{mvMoVi-C} and \textsc{mvMoVi-D} (Section~\ref{sec:experiments}).
\end{enumerate}

\begin{figure}[!t]
    \centering
    \includegraphics[width=1.0\columnwidth]{figures/overview2.png}
    \caption{
        The figure illustrates a scene with four objects $\gO^s = \{\gO_1, \gO_2, \gO_3, \gO_4\}$, observed from three different viewpoints, each described with a set of clearly visible objects: $\gO^1 = \{\gO_3, \gO_4\}, \gO^2 = \{\gO_1, \gO_3, \gO_4\}, \gO^3 = \{\gO_1, \gO_2, \gO_3, \gO_4\}$. 
        The corresponding images are passed through view and content encoders, and sampled global view vector $\rvv$ is used to estimate transformation function $\gT_{\theta^v}$ given by parameters $\theta^v$ predicted using a localisation network.   
        We apply a view-specific inverse $\gT^{-1}_{\theta^v}$ on respective images projecting them to an \emph{implicit} space, which is used to learn view conditioned slot posterior corresponding to GMMs represented by $q(\rvs^v \mid \gT^{-1}_{\theta^v}(\rvx^v))$, which are further aggregated to marginalize viewpoint information, resulting in a content posterior, also a GMM $q(\rvc \mid \{\rvs^1, \dots, \rvs^V\})$, which is further accumulated across all samples resulting in optimal prior $p(\rvc)$.
    }
    \vspace{-10pt}
    \label{fig:overview}
\end{figure}

\section{Related Works}

\paragraph{Identifiable Object-centric learning.} 
Extending nonlinear Independent Component Analysis (ICA) from representation learning to object-specific representational learning has been heavily explored before \citep{burgess2019monet, engelcke2019genesis, greff2019multi} by employing an iterative variational inference approach~\citep{marino2018iterative}, whereas \citet{van2020investigating, lin2020space} adopt more of a generative perspective, studied the effect of object binding and scene composition empirically. 
Recently, the use of iterative attention mechanisms has gained a significant interest~\citep{locatello2020object, engelcke2021genesis, singh2021illiterate, wang2023slot, singh2022neural, emami2022slot}. 
Most of these works operate in a single-view setting, which causes fundamental issues of viewpoint ambiguities in terms of occlusions and uncertainties in binding. 
Recent methods, including \cite{eslami2018neural, arsalan2017synthesizing, tobin2019geometry, wu2016learning} consider a single object from multiple views to tackle this particular problem. Additionally, \cite{kosiorek2018sequential, hsieh2018learning, li2020learning} explore multi-object binding in videos and multiple views, tackling object binding issues across frames. 
Despite their empirical effectiveness, most of these works lack formal identifiability guarantees.
In line with recent efforts analysing theoretical guarantees in object-centric representations \citep{lachapelle2023additive, brady2023provably, kori2024identifiable}, we formally investigate the modelling assumptions and their implications for achieving identifiability guarantees in the context of multi-object, multiview object-centric representation learning settings.

\paragraph{Multiview nonlinear ICA.}
It has been noted that addressing the challenge of nonlinear ICA can involve incorporating a learnable clustering task within the latent representations, thereby imposing asymmetry in the latent distribution \citep{willetts2021don, kivva2022identifiability}. 
Moreover, 
\citet{gresele2020incomplete} delve into multiview nonlinear ICA, particularly in scenarios involving corrupted observations, where they aim to recover invariant representations while accounting for certain ambiguities. 
Along similar lines, \citet{daunhawer2023identifiability, von2021self} explore the concept of style-content identification using contrastive learning, focusing on addressing the multiview nonlinear ICA problem.
Here, we work along similar lines by emphasising the learning of invariant content and identifiable object-centric representations. 
We achieve this by formulating a reconstruction objective where the enforced invariance and equivariance stem from the underlying probabilistic graphical model rather than relying on a contrastive learning objective. 
Similar to the noiseless setting in \cite{gresele2020incomplete}, we demonstrate the recovery of invariant content representations using different subsets of viewpoints. 

\section{Preliminaries}

\textbf{Probabilistic Slot Attention (PSA)} as introduced by \citet{kori2024identifiable}, presents a probabilistic interpretation of the slot attention algorithm \citep{locatello2020object}. 
In PSA, a set of feature embeddings $\mathbf{z} \in \mathbb{R}^{N \times d}$ per input $\mathbf{x}$ is taken as input, and an iterative Expectation Maximization (EM) algorithm is applied over these embeddings. 
This process results in a Gaussian Mixture Model (GMM) characterized by mean ($\bmu \in \mathbb{R}^{K \times d}$), variance ($\bsigma^2 \in \mathbb{R}^{K \times d}$), and mixing coefficients ($\bpi \in [0, 1]^{K \times 1}$). 
In summary, PSA employs the initial mean sampled from the prior distribution and initial variance initialized with unit vector, then iteratively updates the mean based on assignment probabilities ($A_{nk}$) using Eqn. \ref{eqn:PSA_EM_attn}, and adjusts the mean and variance accordingly as described in Eqn. \ref{eqn:PSA_EM}, for $T$ iterations. 

\begin{align}
    A_{nk} = \frac{\bpi(t)_k\gN\left(\rvz_n; \bmu(t)_k, \bsigma(t)_k^2\right)}{\sum_{j=1}^{K}\bpi(t)_j\gN\left(\rvz_n;\bmu(t)k, \bsigma(t)_j^2\right)}; \label{eqn:PSA_EM_attn}\\
    \hat{A}_{nk} = A_{nk}/\sum_{l=1}^NA_{lk}; \; \bpi(t+1)_k = \sum_{n=1}^N A_{nk} / N ;\\
    \bmu(t+1)_k = \sum_{n=1}^N \hat{A}_{nk}\rvz_n; \\  
    \bsigma(t+1)_k^2 = \sum_{n=1}^N \hat{A}_{nk} \left(\rvz_n - \bmu(t+1)_k\right)^2 &&
    \label{eqn:PSA_EM}
\end{align}

\textbf{Identifiable representations.} A model is considered identifiable when different training iterations yield consistent latent distributions, thereby resulting in identical model parameters \citep{khemakhem2020variational, khemakhem2020ice}. 
In the context of a parameter space $\Theta$ and a family of mixing functions $\gF$, identifiability of the model on the dataset $\gX$ is established if, for any $\theta_1, \theta_2 \sim \Theta$ and $f_{\theta_1}, f_{\theta_2} \sim \gF$, the condition $p(f_{\theta_1}^{-1}(\mathbf{x})) = p(f_{\theta_2}^{-1}(\mathbf{x}))$ holds for all $\rvx \in \gX$, implying $\theta_1 = \theta_2$. 
However, in practical scenarios, exact equality or ``strong'' identifiability is often unnecessary, as establishing relationships to transformations, which can be manually recovered, proves equally effective, resulting in a notion of weak identifiability, where relationships are recovered up to an affine transformation \citep{khemakhem2020ice, kivva2022identifiability}. 
Similar identifiability relations have been elucidated for OCL in \citep{brady2023provably, lachapelle2023additive, kori2024identifiable, mansouri2023object}. 
The notion of $\sim_s$ equivalence relation is elaborated in Dfn. \ref{dfn:sequivalence}.

\begin{definition}($\sim_s$ equivalence \citep{kori2024identifiable})
Let $f_{\boldsymbol{\theta}}: \gS \rightarrow  \gX$ denote a mapping from slot representation space $\mathcal{S}$ to image space $\mathcal{X}$ (satisfying Assumption \ref{ass:weak_inj}), the equivalence relation $\sim_s$ w.r.t. to parameters $\boldsymbol{\theta} \in \boldsymbol{\Theta}$ is defined as: $ \boldsymbol{\theta}_1 \sim_{s} \boldsymbol{\theta}_2 \Leftrightarrow $
    \begin{align}
        \exists \mP, \mH, \mathbf{c} : f_{\boldsymbol{\theta}_1}^{-1}(\rvx; \rvv) = \mP (f_{\boldsymbol{\theta}_2}^{-1}(\rvx; \rvv)\mH + \rva), \forall \rvx \in \gX, \nonumber
    \end{align}
    where $\mP \in \gP \subseteq \{0, 1\}^{K \times K}$ is a permutation matrix, $\mH \in \mathbb{R}^{d \times d}$ is an affine matrix, and $\rva \in \mathbb{R}^d$.
    \label{dfn:sequivalence}
\end{definition}

\section{VISA Formalism}
\label{sec:formalism}

\begin{figure}[!t] 
    \centering
        \begin{tikzpicture}
            \node[latent] (c) {$\rvc$};
            \node[latent, above=of c] (u) {$u$}; 
            \node[latent, draw=none, left=15pt of u] (muc) {$\bmu_k$};
            \node[latent, draw=none, left=15pt of c] (sigmac) {$\bsigma_k$};
            \node[rectangle, draw=none, right=10pt of u] (pic) {$\bpi_k$};

            \node[draw, diamond, right=35pt of c] (tc) {$\bar{\rvc}$};
            \node[obs, right=15pt of tc] (x) {$\rvx$};


            \node[latent, right=15pt of x] (v) {$\rvv$};
            \node[latent, above=of v] (uv) {$v$}; 
            
            \node[latent, draw=none, right=10pt of v] (sigmav) {$\bsigma_v$};
            \node[latent, draw=none, above=5pt of sigmav] (muv) {$\bmu_v$};
            \node[rectangle, draw=none, right=15pt of uv] (piv) {$\bpi_v$};

            \plate[] {kplate} {(muc)(sigmac)}{$K$};
            \plate[] {nplate} {(c)(u)}{$N$};
            
            \node[draw, rounded corners, fit={(v) (x) (tc) (uv)},  inner xsep=9pt, inner ysep=8pt, yshift=-5pt] (vplate) {};
    
            \node[above left] at (vplate.south east) {$V$};
            
            
            \draw[->, shorten <=7pt] (muc.center) -- (c);
            \draw[->, shorten <=7pt] (sigmac.center) -- (c);
            \draw[->, shorten <=7pt] (muv.center) -- (v);
            \draw[->, shorten <=7pt] (sigmav.center) -- (v);
             \draw [->] (v) to [out=120, in=60] node[midway, above] {$\bar{\rvc} = \gT_{\theta^v}(\rvc)$} (tc);

            \edge{u}{c}
            \edge{c}{tc}
            \edge{tc}{x}
            \edge{v}{x}
            \edge{uv}{v}
            \edge{pic}{u}
            \edge{piv}{uv}
        \end{tikzpicture}
    \caption{ 
    \textbf{Graphical model for multi-view probabilistic slot attention:} 
    For every image in a dataset a view $\rvv \in \mathbb{R}^{d_v} \sim p(\rvv)$, this view is used to compute transformation $\gT_{\theta^v}$. 
    Similarly, desired number ($< K$) of content representations $\rvc \in \mathbb{R}^{N \times d_s}$ are sampled content distribution $p(\rvc)$. Finally, the image $\rvx$ is generated using the transformed content $\gT_{\theta^v}(\rvc)$ and view $\rvv$.
    }
    \label{fig:generative_model}
    \vspace{-20pt}
\end{figure}
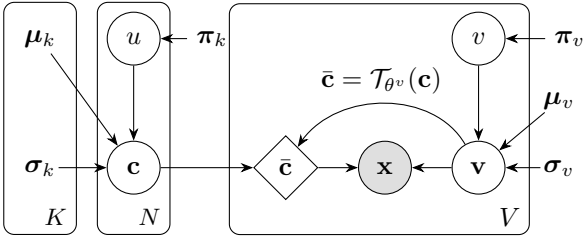

Let $\rvx^{1:V} = \{\rvx^1, \dots \rvx^V\} \in \gX = \gX^1 \times \dots \times \gX^V$, $V$ views of the same scene observed from different viewpoints with an observational space $\gX \subseteq \mathbb{R}^{V \times H \times W \times C}$. 
We consider $[V]$ as a shorthand notation for $\{1, \dots, V\}$.
Let $\gO^e = \gO^1 \cup \dots \cup \gO^V$ correspond to an abstract notion of object sets of an environment, while $\gO^v, \forall v \in [V]$ is a set of objects present in a considered viewpoint $v$.
Importantly, we consider that the number of objects per viewpoint can vary, \emph{i.e.,} $|\gO^1 \cup \dots \cup \gO^V | \geq |\gO^v| \; \forall \; v \in [V]$, allowing for partial or full occlusion in some viewpoints. 
Let $\rvv^{1:V} \in \gV = \gV^1\times\dots\times\gV^V \subseteq \mathbb{R}^{V \times d_v}$ be inferred viewpoint-specific information\footnote{We abuse the terminology by considering viewpoint, lighting, object dimension, to be encoded in a representation $\rvv$. 
Note that the $\rvv$ is inferred by the model.}, while $\rvs^{1:V}_{1:K}\in \gS = \gS^1\times\dots\times\gS^V \subseteq \mathbb{R}^{V \times K \times d_s}$ correspond to a viewpoint-specific slot representation. 
Let $\rvc_{1:K} \in \gC \subseteq \mathbb{R}^{K \times d_c}$ capture the notion of an \emph{aggregate}, effectively accumulating the object knowledge across viewpoints.
For any subset $A$ of $[V]$, we represent scene observations as $\mathbf{x}^A = \{\mathbf{x}^i : \forall i \in A\} \in \times_{i \in A} \mathcal{X}^i$. 
The inferred viewpoints and the view specific slots are denoted as $\mathbf{v}^A = \{\mathbf{v}^i: \forall i \in A\} \in \times_{i \in A} \mathcal{V}^i$, and $\mathbf{s}^A_{1:K} = \{\mathbf{s}^i_{1:K}: \forall i \in A\} \in \times_{i \in A} \mathcal{S}^i$, respectively.  
We define $p_A(\mathbf{c})$ as the distribution of $\mathbf{c}$ over $A$.  
A more comprehensive summary of notations and terminologies is provided in App. \ref{app:notations}.

In modelling, \texttt{w.l.o.g}, we consider access to a certain subset $A \subseteq [V]$, ensuring the model's applicability across different scenarios.  
Furthermore, to simplify notation, we sometimes do not include the superscript denoting the full set of views, thereby using $\mathbf{x} = \mathbf{x}^A$, $\mathbf{s}_{1:K} = \mathbf{s}^A_{1:K}$, and $\mathbf{v} = \mathbf{v}^A$ interchangeably.  
Likewise, if we do not specify the subscripts for $\mathbf{c}$ and $\mathbf{s}$, it implies they represent the entire collection of objects, specifically as $\mathbf{s} = \mathbf{s}^A_{1:K}$ and $\mathbf{c} = \mathbf{c}_{1:K}$. 
Lastly, for any function $f$ that operates on two distinct inputs $\rvx = f(\rvz, \rvv)$, its inverse is denoted by $\rvz = f^{-1}(\mathbf{x}; \mathbf{v})$, which signifies the reversal of $f$ conditioned on a variable $\mathbf{v}$.
%
%
%
%
%
%
%
In the rest of this section, we introduce all the components involved in our model. We also introduce assumptions, examples, and intuition wherever necessary.
Considering the generative model Eqn. \ref{eqn:likelihood}, which is overviewed in graphical model Fig. \ref{fig:generative_model}, any scene $\rvx$ is generated using view $\rvv$ and content $\rvc$. Here, both $\rvc$ and $\rvv$ are latent variables learned with variational inference\citep{kingma2013auto}.

\begin{align}
    p(\rvx) = \iint p(\rvx^v \mid   \gT_{\theta^v}(\rvc), \rvv^v) \; p(\rvc) \; p(\rvv^v) \; d\rvv^v \; d\rvc
    \label{eqn:likelihood}
\end{align}



\textbf{View model. } Given that the view property remains consistent across all objects, we treat the view as a global, image-level property as opposed to \citet{yuan2024unsupervisedocloc}, where view is treated as an object-level property.
Assuming access to a discrete set of viewpoints denoted by $A$, we consider prior over a view distribution to be a GMM represented by $p(\rvv) = \sum_{v=1}^{|A|} \bpi^v \gN(\rvv; \bmu_v, \bsigma^2_v )$.
To learn the parameters of this GMM, we consider the posterior of the form $q_{\phi}(\rvv \mid \rvx^v) \; \forall \; v \in A$\footnote{We consider the parametric form of $q$ to be Gaussian.}.
In both prior and posterior, we consider the covariance to be diagonal, implicitly making an ICA assumption \citep{khemakhem2020variational}.
The sampled variable $\rvv \sim q_{\theta}(\rvv \mid \rvx^v)$ is used to estimate transformation parameters $\theta^v \in \mathbb{R}^{3 \times 2}$ as in \citet{jaderberg2015spatial} which makes an affine transformation map $\gT_{\theta^v}$, which is later applied on content $\rvc$ and on view-specific slots $\rvs$.
It is important to note that we use the same set of parameters $\phi$ across all viewpoints in $A$ for inferring view information $\rvv$.

\textbf{Viewpoint specific slots.} 
As illustrated in Fig. \ref{fig:overview} the inference of $\rvc$ depends on the view-specific slots $\rvs$. 
For a considered image $\rvx^v, v \in A$, we first apply an inverse view transformation $\gT_{\theta^v}^{-1}$ and model the slot distribution as a spatial mixture model represented by $q(\rvs^A_{1:K} \mid \gT_{\theta^v}^{-1}(\rvx^A))$.
The inverse transformation makes sure that the estimated object representations across all view in $A$ are in a common implicit representation space.
As this is an intermediate variable which does not show up in our generative model in Eqn. \ref{eqn:likelihood}, we update the corresponding parameters with closed-form equations via expectation maximisation algorithm as in \cite{kori2024identifiable}.
The resulting slot posterior is a conditional GMM as described in Eqn. \ref{eqn:slot_posterior}, where $\bar{\rvx}^v = \gT_{\theta^v}^{-1}(\rvx^v)$ is a transformed inputs, $(\mu_k(\bar{\rvx}^v), \sigma^2_k(\bar{\rvx}^v), \pi_k(\bar{\rvx}^v))$ are mean, diagonal covariance, and mixing coefficients for the considered a view and object.
\vspace{-20pt}

\begin{align}
    q(\rvs^v \mid \bar{\rvx}^v) = \sum_{k=1}^K \bpi_{k}(\rvx^v) \gN \Big(\rvs^v_k; \bmu_{k}(\bar{\rvx}^v), \bsigma_{k}^2(\bar{\rvx}^v) \Big)
    \label{eqn:slot_posterior}
\end{align}


\textbf{Representation matching.} Given the permutation equivariance property of slot representations, we use a matching function with a permutation matrix $\mP^v$, $m_s: \gS^A \rightarrow \gS^A$ such that $m_s(\rvs^v_{1:K}) = \mP^v \rvs^v_{1:K}$ mapping representation axis \texttt{w.r.t} $\mP^v$.
The permutation matrix $\mP^v$ is estimated by considering the slots of the first viewpoint $\rvs^1$ as a base representation, and other representations $\rvs^v \, \forall v \in A$ are matched to align with it.
We utilise Hungarian matching, as illustrated in \cite{locatello2020object, wang2023slot}, to estimate this permutation matrix $\mP^v$, to control the noise in the matching algorithm, we introduce view-warmup strategy, which we detail in App. \ref{app:viewwarmup}.

\textbf{Content aggregator.} We consider $g:\gS \rightarrow \gC$ as a content aggregator function, which marginalises the effect of view conditioning. 
To achieve this, we consider a convex combination of all the aligned slot representations (aligned to a base representation), considering mixing coefficients $\bpi_{k}(\rvx^v)$ (we use $\bpi^v_k$ for simplicity) in Eqn. \ref{eqn:slot_posterior} as a combination weight. 
The convex combination accounts for potential object occlusions, which may cause objects to be absent in particular views ensuring only active representations are combined (refer to an intuition below), resulting in a content posterior ($q(\rvc \mid \rvs)$), which is a GMM  with mixing coefficients $ \tilde{\bpi}_k = \left(\sum_{v = 1}^{|A|} \bpi^v_k \right)/|A| $ and the parameters described in Eqn. \ref{eqn:GMM_parameters} (\texttt{w.l.o.g} we consider $\rvs, \bpi$ to represent aligned representations), refer to Lemma \ref{lemma:GMMconvexcombination}, with $w_i=1/|A| \, \forall i \in A$. 
Additionally, algorithm \ref{alg:the_alg} details the entire forward process.


\begin{tcolorbox}[
    enhanced,
    colframe=blue!50!white, 
    colback=blue!5!white,   
    coltitle=white,         
    colbacktitle=blue!50!white,, 
    title=Intuition: Content aggregation,        
    fonttitle=\bfseries,    
    boxrule=1mm,            
    rounded corners,        
    drop shadow,            
    sharp corners=northwest, 
    fontupper=\small\itshape, 
    left=5pt, right=5pt, top=3pt, bottom=3pt
]
Based on illustrated example in Fig. \ref{fig:overview}, for images $\rvx^1, \rvx^2, \rvx^3$, 
the resulting matched slots and mixing coefficients correspond to $\rvs^1 = \{\rvs^1_{r}, \rvs^1_{r}, \rvs^1_{\gO_3}, \rvs^1_{\gO_4}, \rvs^1_{b}\}, \rvs^2 = \{\rvs^2_{\gO_1}, \rvs^2_{r}, \rvs^2_{\gO_3}, \rvs^2_{\gO_4}, \rvs^2_{b}\}, \rvs^3 = \{\rvs^3_{\gO_1},  \rvs^3_{\gO_2}, \rvs^3_{\gO_3}, \rvs^3_{\gO_4}, \rvs^3_{b}\}$, where $\rvs^v_{\gO_i}, \rvs^v_{r},$ and $\rvs^v_{b}$ correspond to slot representation for object $\gO_i$, random slot representation and background information, respectively, with mixing coefficients $\bpi^1 = \{0, 0, 1, 1, 1\}, \bpi^2 = \{1, 0, 1, 1, 1\},$ and $\bpi^1 = \{1, 1, 1, 1, 1\}$.
Proposed aggregation merges the slots ignoring the random slots $\rvs^v_{r}$, resulting in $\rvc_{\gO_1} = (\rvs^2_{\gO_1} + \rvs^3_{\gO_1})/2, \rvc_{\gO_2} = \rvs^3_{\gO_2}$ and so on.
\end{tcolorbox}

\begin{align}
    g(\rvs^{1:V}_{1:K}, \bpi^{1:V}) = \sum_{v=1}^{|A|} \frac{\bpi_{1:k}^v}{|A|\tilde{\bpi}^v_k} \rvs^v_{1:K} ;
    \label{eqn:convex_combination} \\
    \tilde{\bmu}_k(\rvx^A) = \sum_{v=1}^{|A|} \frac{\bpi_k^v}{|A|\tilde{\bpi}^v_k} \bmu_k(\rvx^v) ; \nonumber \\
    \tilde{\bsigma}^2(\rvx^A) = \sum_{v=1}^{|A|} \left( \frac{\bpi_k^v}{|A|\tilde{\bpi}^v_k} \right)^2 \bsigma^2_k(\rvx^v) ; \,  
    \label{eqn:GMM_parameters}
\end{align}

\textbf{Mixing function and training objective. }
We consider both \emph{additive and non-additive} (ref. definition \ref{dfn:additivemodels}) mixing functions $f_{d}: \gC \times \gV^v \rightarrow \gX^v$.
For additive decoders, we use a spatial-broadcasting \citep{greff2019multi} and MLP decoders, and for non-additive mixing function, we use auto-regressive transformers \citep{vaswani2017attention}. 
We use the shared decoder $f_{d}$ for all views and objects, modelling the conditional distribution $p(\rvx^v \mid \gT_{\theta^v}(\rvc), \rvv^v)$.
%
To train our model in an end-to-end fashion, we maximise the log-likelihood of the joint $p(\rvx^A)$, which results in the evidence lower bound (ELBO), Eqn. \ref{eqn:elbo}, check Lemma \ref{lemma:elbo_derivation}. 
Here, we consider the distribution form of $p(\rvx^v \mid \rvc, \rvv^v)$ to be Gaussian with learnable mean with isotropic covariance. 
\vspace{-20pt} 

\begin{align}
    \mathbb{E}  \log p(\rvx \mid \gT_{\theta^v}(\rvc), \rvv) - \mathrm{KL}\left(q(\rvv \mid \rvx)  \parallel p(\rvv) \right)
    \label{eqn:elbo}
\end{align}




\paragraph{Computational complexity:}  VISA achieves $\mathcal{O(VTNK}d)$ with the added complexity of $2\mathcal{O(VN}d)$ for inverse and forward view point transformation given by $\mathcal{T}_{\theta}$, while it retains the complexity per view to be $\mathcal{O(TNK}d)$ which is the same as slot attention and probabilistic slot attention. Additionally, the representation matching function contributes $\mathcal{O(VK}^3d)$: this term does not alter the dominant term, in the general case when $\mathcal{K << N}$. Similar to PSA, when VISA is combined with an additive decoder, the complexity of the decoder can be lowered due to the property of automatic relevance determination (ARD), eliminating the need to decode inactive slots.

\section{Theoretical Analysis}
\label{sec:identifiability}

In this section, we leverage the properties of the proposed model to theoretically demonstrate the learning of identifiable representations under challenging spatial ambiguities.
In this work, we consider our data-generating process to satisfy a viewpoint sufficiency assumption (refer to~\ref {ass:viewpoint_sufficiency}).

\begin{assumption}(View-point sufficiency) For any set $A \subseteq [V]$, we consider set $A$ to be view-point sufficient iff $|\gO^A| = |\gO^e|$. This basically means that all the objects are visible across all the considered views $A$, even when an individual view may not contain all the objects.
\label{ass:viewpoint_sufficiency}
\end{assumption}

\begin{tcolorbox}[
    enhanced,
    colframe=blue!50!white, 
    colback=blue!5!white,   
    coltitle=white,         
    colbacktitle=blue!50!white,, 
    title=Example: Viewpoint-sufficiency,   
    fonttitle=\bfseries,    
    boxrule=1mm,            
    rounded corners,        
    drop shadow,            
    sharp corners=northwest, 
    fontupper=\small\itshape, 
    left=5pt, right=5pt, top=3pt, bottom=3pt
]
\begin{example}
    Based on illustrated example in Figure \ref{fig:overview}, the scene is composition of  four objects $\gO^e = \{\gO_1, \gO_2, \gO_3, \gO_4 \}$, view point subset $A=[V] = \{1, 2, 3\}$ is considered to be view point sufficient since $\bigcup_{v \in A} \gO^v = \{\gO_3, \gO_4 \} \cup \{\gO_1, \gO_3, \gO_4 \} \cup \{\gO_1, \gO_2, \gO_3, \gO_4 \} = \gO^e$.
    \label{exm:sufficiency_illustration}
\end{example}
\end{tcolorbox}

Given that we learn the parameters of our view-specific spatial GMM with closed-form updates, we do not use an explicit prior minimising KL divergence. Instead, we rely on the fact that marginalising the effect of data points from posterior (\emph{aggregate posterior}) is an optimal prior \citep{hoffman2016elbo,kori2024identifiable}, resulting in $p(\rvc) = \iint q(\rvc | \rvs^A, \rvx^A) d\rvs^A d\rvx^A$.
Given that GMMs are universal density approximates given enough components (even GMMs with diagonal covariances), the resulting aggregate posterior $q(\rvc) = p(\rvc)$ is highly flexible and multi-modal. 
It often suffices to approximate it using a sufficiently large subset of the dataset if marginalising out the entire dataset becomes computationally restrictive.

\begin{lemma}[Optimal Prior]
   For $A \in [V]$, given the a local content distribution $q(\rvc_{1:K} \mid \rvs^A_{1:K}, \rvx^A)$ (per-scene $\rvx^A \in \{ \rvx^A_i\}_{i=1}^M$), which can be expressed as a GMM with $K$ components, the aggregate posterior $q(\rvc)$ is obtained by marginalizing out $\rvx, \rvs$ is a non-degenerate global Gaussian mixture with $MK$ components:
    \begin{align}
        p(\rvc) = q(\rvc) &= \frac{1}{M} \sum_{i=1}^M \sum_{k=1}^K \widehat{\bpi}_{ik} \gN\left(\rvc; \widehat{\bmu}_{ik}, \widehat{\bsigma}^2_{ik} \right).
    \end{align}
    \label{prp:aggregate_poterior1}
\end{lemma}
\begin{sproof}

The result is obtained by integrating the product of involved posterior densities  $q(\rvc \mid \rvs)q(\rvs \mid \rvx)p(\rvx)$. Further, we verify if the mixing coefficients sum to one in the new mixture, proving the aggregate to be well-defined. 
\end{sproof}

With this, we show three main results: firstly, we show that aggregate content representations ($\rvc$) are identifiable without supervision (up to $\sim_s$). Secondly, we show that these representations are invariant to the choice of viewpoints under assumption \ref{ass:viewpoint_sufficiency}. Finally, we show that the model exhibits in an approximate view equivariance. 

\begin{theorem}(Affine Equivalence)
    For any subset $A \subseteq [V]$, such that $|A| > 0$ , given a set of images $\rvx^A \in \gX^A$ and a corresponding aggregate content $\rvc \in \gC$ and a non-degenerate content posterior $q(\rvc \mid \rvs^A)$, considering two mixing function $f_{d}, \tilde{f}_d$ satisfying assumption \ref{ass:weak_inj}, with a shared image, then $\rvc$ are identifiable up to $\sim_s$ equivalence.
    \label{thm:affine_equivalence}
\end{theorem}
\begin{tcolorbox}[
    enhanced,
    colframe=blue!50!white, 
    colback=blue!5!white,   
    coltitle=white,         
    colbacktitle=blue!50!white,, 
    title=Intuition: Affine equivalence,   
    fonttitle=\bfseries,    
    boxrule=1mm,            
    rounded corners,        
    drop shadow,            
    sharp corners=northwest, 
    fontupper=\small\itshape, 
    left=5pt, right=5pt, top=3pt, bottom=3pt
]
    Considering an example \ref{exm:sufficiency_illustration}, with two perfectly trained models $f_d$ and $\tilde{f}_d$. Resulting aggregate contents are described as $\rvc = f_d^{-1}(\rvx^A; \rvv^A) = \{\rvc_{\gO_1}, \rvc_{\gO_2}, \rvc_{\gO_3}, \rvc_{\gO_4}, \rvc_{\gO_b}\}$ and $\tilde{\rvc} = \tilde{f}_d^{-1}(\rvx^A; \rvv^A) = \{\tilde{\rvc}_{\gO_2}, \tilde{\rvc}_{\gO_4}, \tilde{\rvc}_{\gO_3}, \tilde{\rvc}_{\gO_1}, \tilde{\rvc}_{\gO_b}\}$ for $A=[V]=\{1, 2, 3\}$.
    $\sim_s$ equivalence states that there exists a permutation matrix $\mP$ which aligns the object order in $\tilde{\rvc}$ to match with $\rvc$ and there exists and invertible affine mapping $\mA$ such that $\tilde{\rvc}_{\gO_k} = \mA \rvc_{\gO_k} \forall k \in \{1, 2, 3, 4\}$.
\end{tcolorbox}
\begin{sproof}
    To prove the following result, we follow multiple steps as described below:
    \begin{enumerate*}[label=(\roman*).]
        \item We demonstrate the distribution $p(\rvc)$ obtained as a result of lemma \ref{prp:aggregate_poterior1} is non-degenerate and a valid distribution,
        \item With the above results, we demonstrate invertibility restrictions on mixing functions,
        \item Finally, we constrain the subspace to affine, demonstrating $\sim_s$ of aggregate content $\rvc$.
    \end{enumerate*}
\end{sproof}

\begin{theorem}(Invariance of aggregate content)
    For any subset $A, B \subseteq [V]$, such that $|A| > 0, |B| > 0$ and both $A, B$ satisfy an assumption \ref{ass:viewpoint_sufficiency}, we consider aggregate content to be invariant if $f_A \sim_s f_B$ for data $\gX^A \times \gX^B$.
    \label{theorem:invariance}
\end{theorem}
\begin{tcolorbox}[
    enhanced,
    colframe=blue!50!white, 
    colback=blue!5!white,   
    coltitle=white,         
    colbacktitle=blue!50!white,, 
    title=Intuition: Invariant slots,   
    fonttitle=\bfseries,    
    boxrule=1mm,            
    rounded corners,        
    drop shadow,            
    sharp corners=northwest, 
    fontupper=\small\itshape, 
    left=5pt, right=5pt, top=3pt, bottom=3pt
]
    Considering an example \ref{exm:sufficiency_illustration}, with $A=\{1, 3\}, B=\{2, 3\}$, such that sets $A, B$ are viewpoint sufficient.
    Let $f_A$ and $f_B$, be trained models on $\gX^A$ and $\gX^B$ respectively. Resulting in $\rvc = f_A^{-1}(\rvx^A; \rvv^A) = \{\rvc_{\gO_1}, \rvc_{\gO_2}, \rvc_{\gO_3}, \rvc_{\gO_4}, \rvc_{\gO_b}\}$ and $\tilde{\rvc} = \tilde{f}_B^{-1}(\rvx^B; \rvv^B) = \{\tilde{\rvc}_{\gO_2}, \tilde{\rvc}_{\gO_4}, \tilde{\rvc}_{\gO_3}, \tilde{\rvc}_{\gO_1}, \tilde{\rvc}_{\gO_b}\}$.
    Thm. \ref{theorem:invariance} states that the representations $\gT_{\theta^B}^{-1}(\tilde{\rvc}_{\gO_k})$ can be mapped to $\gT_{\theta^A}^{-1}(\rvc_{\gO_k})$ by permuting object indices and an affine transformation. 
\end{tcolorbox}
\begin{sproof}
To prove this, we extend the proof of Thm. \ref{thm:affine_equivalence}, and establish that there exist two inevitable affine functions $h_A, h_B$ for mixing functions $f_A, f_B: \gC \times \gV \rightarrow \gX$ to map representations $\rvc$ with a given view set $\rvv^A$ to observations $\rvx^A$. 
Later, we show that, in the case of invariance, an affine mapping exists from $h_A$ to $h_B$.
\end{sproof}

\begin{theorem} (Approximate representational equivariance)
    For a given aggregate content $\rvc$, for any two views $\rvv, \tilde{\rvv} \sim p_A(\rvv)$, resulting in respective scenes $\rvx \sim p_A(\rvx \mid \rvv, \rvc)$ and $\tilde{\rvx} \sim p_A(\rvx \mid \tilde{\rvv}, \rvc)$, for any homeomorphic transformation $h_x \in \gH_x$ such that $h_x(\rvx) = \tilde{\rvx}$, their exists another homeomorphic transformation $h_v \in \gH_v$ such that $\gH_v \subseteq \gH_x \subseteq \mathbb{R}^{\mathrm{dim}(\rvx)}$ and  $\rvv = h_v^{-1}\left(f^{-1}_d(h_x(\rvx); \rvc)\right)$.
    \label{theorem:equivariance}
\end{theorem}
\begin{remark}
    Note that the theorem only says that the transformation function transforming the view representations $\rvv$ as an effect of the homeomorphic transformation of $\rvx$ lies in the same subspace of input transformations.
\end{remark}
\begin{tcolorbox}[
    enhanced,
    colframe=blue!50!white, 
    colback=blue!5!white,   
    coltitle=white,         
    colbacktitle=blue!50!white,, 
    title=Intuition: Approximate equivariance,   
    fonttitle=\bfseries,    
    boxrule=1mm,            
    rounded corners,        
    drop shadow,            
    sharp corners=northwest, 
    fontupper=\small\itshape, 
    left=5pt, right=5pt, top=3pt, bottom=3pt
]
    In the scenario when the cameras are positioned such that they have overlapping fields of view, and their relative pose (rotation and translation) must avoid degeneracies like aligning on the same plane or mapping points to infinity.
    This results in the transformation between views being smooth, invertible, and consistent. 
    If the scene is planar or depth variations are minimal, the homography can capture the transformation accurately without the need for inverse rendering. 
    Notably, the cameras should have non-zero rotation and translation to avoid collapsing the scene, and their intrinsic parameters must be known or identical to prevent distortions.
    When the scenario satisfies all the above properties, the 2D homography transformation $\mH$ between two camera views can be learned as a homeomorphic transformation \citep{hartley2003multiple}.
\end{tcolorbox}
\begin{sproof}
   We prove the following result by following the steps in Thm. \ref{theorem:invariance}, over a view distribution $p(\rvv)$ but for a fixed content vector $\rvc$.
\end{sproof}

\section{Empirical Evaluation}
\label{sec:experiments}

\begin{figure*}[t]
    \centering
    \includegraphics[width=\textwidth]{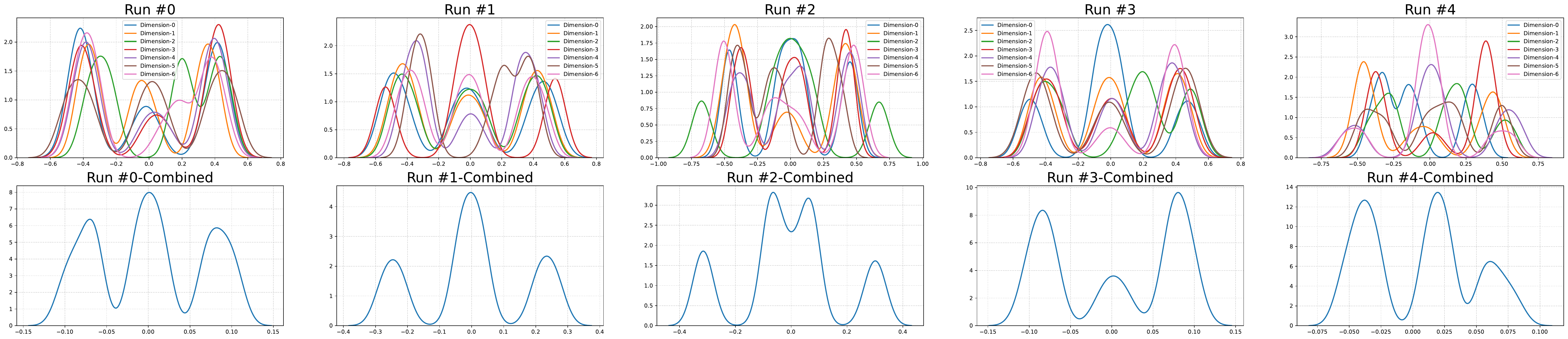}
    \caption{\textbf{Identifiability of $q(\rvc)$}. The top row indicates individual feature distribution across five different runs. The bottom row reflects the feature feature distribution, which we use as a proxy for multi-dimensional features given Lemma \ref{lemma:projected_GMM}.
    As observed, mean feature distribution across runs is either scaled, shifted, or split (increase in number of modes); this provides strong evidence of recovery of the latent space up to affine transformations, empirically verifying our claims in Thm. \ref{thm:affine_equivalence}. 
    } 
    \label{fig:identifiability_pc}
    \vspace{-10pt}
\end{figure*}


\begin{table*}[t]
\scriptsize
\centering
\caption{Comparing identifiability of $q(\rvs)$, $q(\rvc)$, and $p(\rvv)$ scores wrt existing OCL methods.}
\begin{tabular}{@{}lccc|ccc|ccc@{}}
\toprule
\textsc{Method} & \multicolumn{3}{c}{\textsc{CLEVR-mv}} & \multicolumn{3}{c}{\textsc{GQN}} & \multicolumn{3}{c}{\textsc{GSO}} 
\\ 
\cmidrule(l){2-10} 
            & SMCC $\uparrow$ & \textsc{inv-SMCC} $\uparrow$ & MCC $\uparrow$ & SMCC $\uparrow$ & \textsc{inv-SMCC} $\uparrow$ & MCC $\uparrow$ & SMCC $\uparrow$ & \textsc{inv-SMCC} $\uparrow$ & MCC $\uparrow$ \\ 
\midrule
AE             & $0.32 \pm .02$ & - & -  &  $0.29 \pm .02$ & - & - & $0.24 \pm 0.08$ & - & -  \\ 
SA             & $0.47 \pm .03$ & - & -  &  $0.38 \pm .02$ & - & - & $0.28 \pm 0.06$ & - & - \\ 
PSA            & $0.49 \pm .02$ & - & -  &  $0.38 \pm .02$ & - & - & $0.30 \pm 0.04$ & - & - \\ 
MulMON         & $0.61 \pm .03$ & $0.62 \pm .02$ & -  &  $0.59 \pm .06$ & $0.61 \pm .02$ & - & $0.56 \pm 0.04$ & $0.48 \pm 0.06$ & -\\
OCLOC & $0.63 \pm .02$ & $0.64 \pm .01$ & $0.48 \pm .04$  &  $0.60 \pm .03$ & $0.60 \pm .01$ & $0.42 \pm .08$ & $0.58 \pm 0.04$ & $0.54 \pm 0.03$ & $0.46 \pm 0.04$ \\
\midrule
\textbf{\textsc{VISA}} & $0.67 \pm .01$ & $0.66 \pm .01$ & $0.60 \pm .04$ &  $0.59 \pm .01$ & $0.63 \pm .01$ & $0.52 \pm .03$ &  $0.60 \pm .03$ & $0.61 \pm .02$ & $0.58 \pm .03$\\
\bottomrule
\end{tabular}
\label{table:simple_datasets}
\vspace{-10pt}
\end{table*}

Given the work's theoretical focus,  experimentally, we aim to provide strong empirical evidence of our identifiability, invariance, and equivariance claims in a multiview setting. 
We also extend our experiments to standard imaging benchmarks, including \textsc{CLEVR-mv, CLEVR-aug, GQN} \citep{li2020learning}; we additionally demonstrate the framework's scalability to highly diverse setting with \textsc{GSO}~\citep{downs2022google} and proposed datasets \textsc{mv-MoViC, mv-MOViD} which are multiview versions of MoViC dataset with fixed and varying scene-specific cameras~\citep{greff2022kubric}. 

\textbf{Experimental setup.\;} To verify our claims on (i) identifiability claim, we train our model on a given view subset $A \subseteq [V]$ and compare view averaged slot mean correlation coefficient (SMCC) measure as defined  \cite{kori2024identifiable} ($\text{SMCC}(\rvs, \tilde{\rvs})\coloneqq \frac{1}{Kd}\sum_{i=0}^d \sum_{j=0}^d \rho(\rvs_{ij}, \mA \tilde{\rvs}_{\tau(i)j})$ for some permutation map $\tau$ and affine transformation $\mA$), (ii) invariance claim, we train multiple models on different subsets of viewpoints $A, B \subseteq [V]$ and compare the aggregate content representations across models, quantifying the similarities with SMCC, we consider this measure to be invariant SMCC (\textsc{inv-SMCC}), and finally, (iii) for subspace equivariance, we consider a trained model with a view subset $A \subseteq [V]$ and compute MCC of view information $\rvv$ by applying random homeomorphic transformations on samples $\rvx^A \sim \gX^A$ (which can also be done by considering samples $\rvx^B \sim \gX^B$, where cameras relative position satisfy the required constraints \ref{theorem:equivariance}, and analyse $p(\rvv^A)$ and $p(\rvv^B)$). 

\textbf{Models \& baselines.\;} We consider two ablations with two types of decoders: (i) additive with MLPs and spatial broadcasting CNNs and (ii) non-additive decoders, which include transformer models. 
In all cases, we use LeakyReLU activations to satisfy the weak injectivity conditions (Assumption~\ref{ass:weak_inj}). 
In terms of object-centric learning baselines, we compare with standard additive autoencoder setups following~\citep{brady2023provably}, slot-attention (SA)~\citep{locatello2020object}, probabilistic slot-attention (PSA)~\citep{kori2024identifiable}, MulMON~\citep{li2020learning}, and OCLOC~\citep{yuan2024unsupervisedocloc}.

\textbf{Architectures:} As detailed in the paper we use two different classes of decoder architectures: (i) additive and (ii) non-additive; within the additive architecture we use both spatial broadcasting and MLP decoders, for the non-additive architecture we use transformer decoders. Concretely, we follow SA \citep{locatello2020object} for spatial broadcasting decoders and DINOSAUR \citep{seitzer2022bridging} for both MLP and transformer decoder. In detail we use:

\begin{enumerate}
    \item \textbf{Spatial broadcasting decoders} -  Input/Output: The generated slots are $\rvs \in \mathbb{R}^{K \times d}$, each slot representation is broadcasted onto a 2D grid of dimension $8 \times 8 \times d$ and augmented with position embeddings. Similar to slot attention, each such grid is decoded using a shared CNN to produce an output of size $W \times H \times 4$, where $W$ and $H$ are width and height of the image, respectively. The output channels encode RGB color channels and an (unnormalized) alpha mask. Further, we normalize the alpha masks with Softmax and perform convex combinations to obtain reconstruction.
    Shared CNN architecture: $3 \times [\text{Conv} (kernel = 5\times 5, stride=2), \text{LeakyReLU}(0.02)] + \text{Conv} (kernel = 3\times 3, stride=1), \text{LeakyReLU}(0.02)$ 

    \item \textbf{MLP decoders}- Input/Output: similar to the spatial broadcasting decoder, here, each slot representation is broadcasted onto $N$ tokens (resulting in $N \times d$) and augmented with position embeddings. Then the individual slot representation is transformed with a shared MLP decoder to generate a representation corresponding to feature dimension along with additional alpha mask, which is further normalised with Softmax and used in creating convex combinations to obtain reconstruction.
    Shared MLP architecture: $[\text{Linear} (d, d, bias = False), \text{LayerNorm}(d)] + 3 \times [\text{Linear} (d, d_{hidden}), \text{LeakyReLU}(0.02)] +  \text{Linear} (d_{hidden}, d+1) $
    
    \item \textbf{Transformer decoders}: Input/Output- transformer consists of linear transformers encoder output $(N \times d)$ and extracted slots $(K \times d)$ as input, while returning the slot conditioned feature as output with a dimension of $(N \times d)$. Transformer architecture: is made up of 4 transformer blocks, where each transformer block consists of a self-attention on input tokens, cross-attention with the set of slots, and residual two-layer MLP with hidden size $4\times d$. Before the Transformer blocks, both the initial input and the slots are linearly transformed to $d$, followed by a layer norm.
\end{enumerate}
 
\begin{figure}[!t]
    \centering
    \includegraphics[width=\columnwidth]{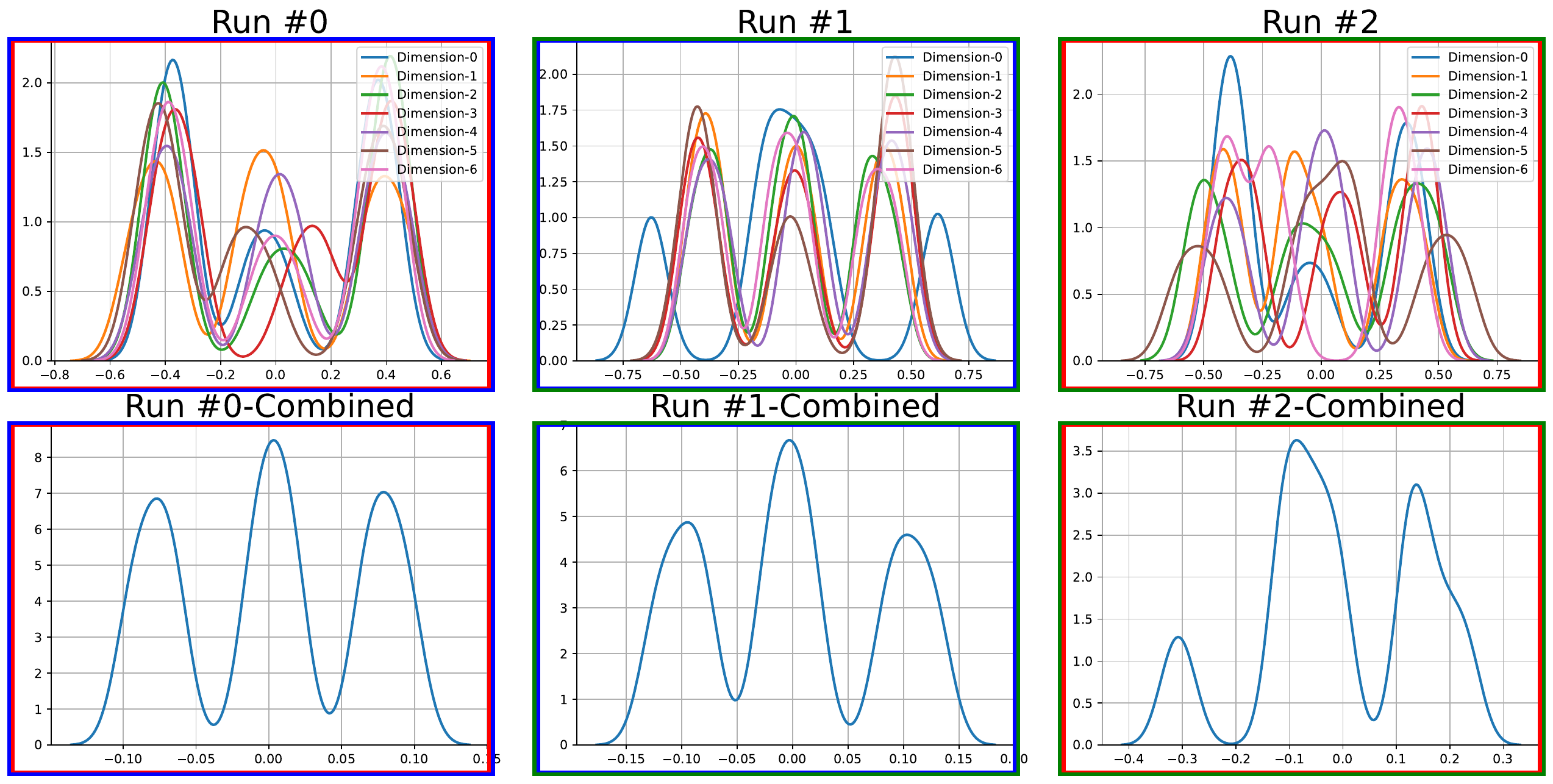}
     \caption{\textbf{Viewpoint invariance for $q(\rvc)$}. The top and bottom row indicates individual feature levels and mean feature distributions, respectively. Each columns reflect  marginalised aggregate content distribution $q(\rvc)$ when trained with different view pairs $\{$(\textcolor{blue}{blue}, \textcolor{red}{red}), (\textcolor{green}{green}, \textcolor{blue}{blue}), and (\textcolor{green}{green}, \textcolor{red}{red})$\}$, respectively. As the resulting distributions with different datasets only vary by an affine transformation, providing strong evidence for Thm. \ref{theorem:invariance}. 
    } 
    \label{fig:invariance_pc}
\end{figure}

\begin{table*}[!t]
\scriptsize
\centering
\caption{Identifiability and generalisability analysis on \textsc{mv-MoViC} dataset. 
}
\resizebox{\textwidth}{!}{
\begin{tabular}{@{}lcccc|cccc@{}}
\toprule
\textsc{Method} & \multicolumn{4}{c}{\textsc{In-domain Results}} & \multicolumn{4}{c}{\textsc{Out-of-domain Results}} \\ 
\cmidrule(l){2-9} 
& mBO $\uparrow$ & SMCC $\uparrow$   & \textsc{inv-SMCC} $\uparrow$ 
 & MCC $\uparrow$ & mBO $\uparrow$ & SMCC $\uparrow$   & \textsc{inv-SMCC} $\uparrow$ 
 & MCC $\uparrow$ \\ 
\midrule
\textsc{SA-MLP}             & $0.28 \pm 0.091$ 
                            & $0.36 \pm 0.004$
                            & -
                            & -
                            & $0.26 \pm 0.08$ 
                            & $0.38 \pm 0.006$
                            & -
                            & -

\\
\textsc{PSA-MLP}            & $0.30 \pm 0.022$ 
                            & $0.38 \pm 0.002$
                            & -
                            & -
                            & $0.30 \pm 0.03$ 
                            & $0.40 \pm 0.005$
                            & -
                            & -
\\
\midrule
\textbf{\textsc{VISA-MLP}}          &   $0.28 \pm 0.021$
                            &   $0.52 \pm 0.021$
                            &   $0.61 \pm 0.023$
                            &   $0.54 \pm 0.026$ 
                            &   $0.27 \pm 0.02$
                            &   $0.51 \pm 0.029$
                            &   $0.58 \pm 0.031$
                            &   $0.52 \pm 0.021$
\\
\midrule
\textsc{SA-Transformer}     &  $0.34 \pm 0.014$     
                            & $0.36 \pm 0.016$
                            & -
                            & -
                            & $0.33 \pm 0.041$
                            & $0.36 \pm 0.043$
                            & -
                            & -
                            \\
\textsc{PSA-Transformer}    &  $0.37 \pm 0.021$     
                            & $0.38 \pm 0.007$
                            & - 
                            & -
                            & $0.37 \pm 0.033$
                            & $0.39 \pm 0.016$
                            & -
                            & -
                            \\
\midrule
\textbf{\textsc{VISA-Transformer}}  &   $0.38 \pm 0.008$    
                            & $0.44 \pm 0.003$
                            & $0.46 \pm 0.001$
                            & $0.53 \pm 0.011$
                            & $0.36 \pm 0.017$
                            & $0.46 \pm 0.033$
                            & $0.46 \pm 0.018$
                            & $0.55 \pm 0.082$
                            \\
\bottomrule
\end{tabular}
}
\label{table:movi-results}
\vspace{-10pt}
\end{table*}

\textbf{\textsc{Case Study 1: Illustration of Identifiability}.\,}
To definitively show the validity of our claims about identifiability (Thm~\ref{thm:affine_equivalence}, Thm~\ref{theorem:invariance}, and Thm~\ref{theorem:equivariance}), we created a synthetic unconfounded scenario for modelling. 
This provides us with two data modalities, Fig, \ref{fig:exampleDGP} (i) projected point cloud data, and (ii) corresponding imagery data, we detail point cloud illustrations in appendix \ref{appendix:toysetup}.
Additionally, this dataset also provides us with the ground truth object and viewpoint features for evaluation.
To visualise the aggregate mixture, following Lemma \ref{lemma:projected_GMM}, we use the projected GMM to interpret the distribution of random variables in $\mathbb{R}^d$.

The data-generating process is thoroughly explained in the App. \ref{subsec:illustrative_dataset}. 
In Fig. \ref{fig:identifiability_pc}, we display the distributions of marginalized aggregate content distribution $q(\rvc)$, comparing individual features and a mean feature across different runs that are either scaled, shifted, or split (increase in number of modes), which is reflective of affine transformation of features across runs. 
To quantitatively measure the same, we computed SMCC and observed it to be $\mathbf{0.72} \pm 0.04$, empirically verifying our Thm. \ref{thm:affine_equivalence}. 
Furthermore, to illustrate the invariance of distribution $q(\rvc)$ across viewpoints (Thm. \ref{theorem:invariance}), we consider three different viewpoints.
We use all possible pairs to learn $q(\rvc)$ distributions as illustrated in Fig. \ref{fig:invariance_pc}, where the distributions are described \texttt{w.r.t} viewpoints described by $\{$\textcolor{green}{g}, \textcolor{red}{r}$\}$, $\{$\textcolor{red}{r}, \textcolor{blue}{b}$\}$, and $\{$\textcolor{green}{g}, \textcolor{blue}{b}$\}$,  respectively. 
These distributions were also found to have similar properties as before, with an observed SMCC of $\mathbf{0.71} \pm 0.11$, further confirming the claims in Thm. \ref{theorem:invariance}.
Additionally, Fig. \ref{fig:motivation2} demonstrates the improvement in identifiability as the number of viewpoints increases.

%

\textbf{\textsc{Case Study 2: Imaging Applications}.\,} 
We first evaluate the framework on standard benchmarks, specifically focusing on \textsc{CLEVR-mv}, \textsc{CLEVR-aug}, \textsc{GQN}, and \textsc{GSO} with simple objects. 
Given the \emph{true generative factors} are unobserved, we derive our quantitative assessments from multiple runs.
The results are shown in Table \ref{table:simple_datasets}, confirming the validity of our theory on imaging datasets. 
Regarding the baseline comparisons that utilize a single viewpoint, the \textsc{inv-SMCC} mirrors the SMCC due to its inherent design (\emph{i.e.,} aggregation of a set with a single element is the same element). 
Moreover, in the case of \textsc{AE, SA, PSA}, and \textsc{MulMON}, the models do not estimate view information but either treat them independently or use the observed view conditioning, rendering the MCC metric inapplicable.
Fig. \ref{fig:|A|_identifiability} showcases how the number of viewpoints impacts the identifiability of the $\rvs, \rvv$, and $\rvc$ variables; the involved experiments reflect the increase in performance with an increase in the number of views, across all benchmark datasets.

Additionally, we demonstrate our methodology on proposed complex datasets, \textsc{mv-MoViC} and \textsc{mv-MoViD}, the latter dataset enables us to examine the model performs when the assumption \ref{ass:viewpoint_sufficiency} is not satisfied. 
To evaluate model behaviour in an environment with consistent objects but with different viewpoints, we conducted in-domain and out-of-domain (OOD) evaluations.
For in-domain analysis, the model is trained and assessed on the same viewpoint group $A=[1,2,3]$. 
Conversely, for OOD evaluation, we consider the previously trained model but test it against a new set of viewpoints $B=[3, 4, 5]$. 
The findings presented in Table \ref{table:movi-results} regarding the \textsc{mv-MoViC} dataset reveal that the SMCC, \textsc{inv-SMCC}, and MCC metrics show similar performance across both domains. 
This indicates that the distributional characteristics remain unchanged when both the training and testing environments contain the same objects. The \textsc{mv-MoViD} dataset analysis can be found in App. \ref{appendix:exps}.

\section{Conclusion \& Discussion}
Understanding when object-centric representations are both unambiguous and identifiable is essential for developing large-scale models with provable correctness guarantees. Unlike most existing work on identifiability, which largely focuses on single-view setups, we offer identifiability guarantees in multi-view scenarios.
We use distributional assumptions for latent slot and view representations, drawing inspiration from mixture model-based structures. To achieve this, we propose a model that is viewpoint-agnostic and does not require additional view-conditioning information.

Our model specifically guarantees the identifiability of view-specific slot representations, viewpoint-invariant content representations, and view representations, all without the need for additional supervision (up to an equivalence relation). We visually validate our theoretical claims with unconfounded synthetic dataset with illustrative 2D data plots. We then empirically demonstrate the model's identifiability properties on multiple object-centric benchmarks, highlighting its ability to resolve view ambiguities in imaging applications.
Furthermore, we showcase the scalability of our approach on large-scale datasets and more complex decoders using realistic datasets and transformer decoders, respectively, demonstrating its capacity to scale effectively with both data volume and decoder complexity.

\textbf{Limitations \& future work.}\, We recognize that our assumptions, particularly regarding the \textit{viewpoint sufficiency}, are strong and may not always hold in practice. 
However, we did not observe limiting effects of this assumption on the proposed \textsc{mv-MoViD} dataset. A more extensive analysis of this assumption and its implications in real-world applications is left for future work. 
We would also highlight that the \textit{weak injectivity} of the mixing function may not always hold for different types of architectures. While generally applicable, the piecewise-affine functions we use may not always capture valid assumptions for real-world problems, \emph{e.g.}, when the model is misspecified. 
Nevertheless, to the best of our knowledge, our theoretical results on multi-object, multi-view identifiability are unique 
and capture key concepts in object-centric representation learning, opening various new avenues for future research along the lines of generalisability, world-modelling, and planning.

\section*{Impact Statement}
This paper proposes a view invariant slot attention algorithm, addressing spatial ambiguities
with identifiability guarantees. 
The work extends theoretical advancements in
the field of OCL, and as such, it has little immediate societal or ethical consequences. 
Our method
might be a step towards interpretable, equivariant, and aligned models, which are desired properties
of trustworthy AI.

\section*{Acknowledgements}
A. Kori is supported by UKRI (grant number EP/S023356/1), as part of the UKRI Centre for Doctoral
Training in Safe and Trusted AI. B. Glocker received support from the Royal Academy of Engineering as part of his Kheiron/RAEng Research Chair, and acknowledges the support of the UKRI AI programme, and the EPSRC, for CHAI - EPSRC Causality in Healthcare AI Hub (grant no. EP/Y028856/1).
Additionally, we thank Fabio De Sousa Ribeiro for insightful discussions and providing feedback on the initial draft of the paper.
\newpage
\bibliographystyle{icml2024}
\bibliography{main}

\newpage
\appendix
\onecolumn

\appendix

\section{Notations}
\label{app:notations}

\renewcommand{\arraystretch}{1.5} 
\begin{tabularx}{\textwidth}{L{4cm} X}
    $\mathcal{O}^v$ & $: $ Abstract object set as observed from viewpoint $v$. \\ 
    $[V] = \{1, \dots, V\}$ & $: $ Exhaustive set of viewpoints, representing all possible views. \\
    $A, B \subset [V]$ & $: $ Subsets of viewpoints, used for training. \\
    $\gX = \times_{v \in A} \gX^v$ & $: $ Data space, formed by the Cartesian product of data spaces for each view in subset $A$. \\
    $\rvx^A = \{\rvx^v : \forall v \in A\} \in \gX$ & $: $ Data sample, where $\rvx^v$ is the data from view $v$, and $\rvx^A$ represents the set of data across all views in $A$. \\
    $f_e$ & $: $ Encoder model maps input data to a latent space or feature representation. \\
    $\rvz$ & $: $ Spatial latent features, representing inferred spatial properties from the data. \\
    $\gS$ & $: $ View-specific slot space, a space for features that are tied to particular viewpoints. \\
    $\gC$ & $: $ View-invariant content space, representing features that are constant across different viewpoints. \\
    $\rvs \in \gS$ & $: $ Samples from the view-specific slot space, representing view-dependent latent features. \\
    $\rvc \in \gC$ & $: $ Samples from the view-invariant content space, representing features that remain consistent across views. \\
    $f_s, \tilde{f}_s$ & $: $  Slot attention module, responsible for attending to and disentangling different parts of the input related to different views. \\
    $f_d, \tilde{f}_d$ & $: $ Mixing function, which combines view-specific and view-invariant features into a unified representation. \\
    $\gV$ & $: $ View information space, a space that encodes information specific to each viewpoint (e.g., angle, position). \\
    $\rvv \in \gV$ & $: $ A sample from the view information space representing a specific view or camera configuration. \\
    $f_v, \tilde{f}_v$ & $: $ View extractor function, which extracts viewpoint-related information from the data. \\
    $\bmu_c, \bmu_s, \bmu_v$ & $: $ Mean of invariant content, view-specific slots, and view distributions. \\
    $\bsigma_c, \bsigma_s, \bsigma_v$ & $: $  Standard deviation of invariant content, view-specific slots, and view distributions. \\
    $\bpi_c, \bpi_s, \bpi_v$ & $: $ Mixing coefficients of invariant content, view-specific slots, and view distributions. \\
    $A_{nk}$ & $: $ Assignment confidence of a slot $k$ getting mapped to token $n$. \\
    $\mP \in \gP \subseteq \{0, 1\}^{K \times K}$ & $:$ Permutation matrix. \\
    $m_s$ & $: $ Matching function, used to align object representations across views. \\
    $\Delta^K$ & $: $ Simplex in the space of dimension K.\\
    $\gH_x, \gH_v$ & $: $ Space of homeomorphic transformation. \\    
\end{tabularx}

\newpage 
\section{Extended Related Works}
\label{app:ext_related_works}

\paragraph{Identifiable representation learning.}
Learning meaningful representations from unlabeled data has long been a primary objective of deep learning \citep{bengio2013representation}.
Several approaches, such as those proposed by \citep{higgins2017betavae, pmlr-v80-kim18b, eastwood2018a, pmlr-v97-mathieu19a}, relied on independence assumptions between latent variables to learn disentangled representations. However, \cite{hyvarinen1999nonlinear,locatello2019challenging} demonstrated the provable impossibility of unsupervised methods for learning independent latent representations from i.i.d. data.
Which is tackled by restricting mixing functions to conformal maps \citep{buchholz2022function} or volume-preserving transformations \citep{yang2022nonlinear}, or with additional data assumptions \citep{zimmermann2021contrastive, locatello2020weakly, brehmer2022weakly, ahuja2022interventional, von2021self}, or by imposing structure in the latent space as in nonlinear Independent Component Analysis (ICA) \citep{pmlr-v89-hyvarinen19a,khemakhem2020variational, Khemakhem2020_ice}, resulting in identifiable models.
In the context of nonlinear ICA, \cite{dilokthanakul2016deep} introduced a VAE model with a GMM prior, and \cite{willetts2021don} empirically demonstrated the effectiveness of the GMM prior, which was later rigorously proven by \cite{kivva2022identifiability}. 
\cite{kori2024identifiable} use this notion of latent GMM in the context of OCL, achieving identifiability guarantees for object-centric representations.
Here, we use this notion in the context of multi-view object-centric representations, tackling the issues with spatial ambiguities and uncertainties in bindings.

\paragraph{Multi-view Object-centric learning.}
Recent progress in multi-view object-centric learning has seen notable contributions from methods like \textsc{MulMON} \citep{li2020learning}, \textsc{ROOTS} \citep{chen2021roots}, \textsc{SlotLifter}\citep{liu2025slotlifter}, and \textsc{uOCF}\citep{luo2024unsuperviseduocf}, each offering distinct approaches to compositional representation learning. However, these methods rely heavily on viewpoint annotations, which limit their applicability in fully unsupervised settings.
\textsc{MulMON} refines object representations iteratively using annotated viewpoint-image pairs, while \textsc{ROOTS, SlotLifter, uOCF} estimates 3D object positions performing an inverse rendering operation within a grid and projects them into image space via viewpoint transformations.
In contrast, we deal with fully unsupervised framework without the need of viewpoint annotations while providing approximate viewpoint equivariance for object representations.

\paragraph{Temporal Object-centric learning.}
An alternative approach to bypass the need for viewpoint annotations leverages temporal information. Methods for learning from single-viewpoint video sequences, such as Relational \textsc{N-EM} \citep{van2018relational}, \textsc{SQAIR} \citep{kosiorek2018sequential}, \textsc{SILOT} \citep{crawford2020exploiting}, and \textsc{SAVi} \citep{kipf2021conditional}, focus on modeling object motion, interactions, and identity tracking across frames, even under occlusion. However, these methods assume fixed viewpoints, making them unsuitable for multi-view scenarios where objects appear in different spatial configurations. Additionally, object motion affects individual objects independently, unlike viewpoint changes, which influence the entire scene.
Recent advances such as \textsc{DyMON} \citep{li2021objectdymon} extend multi-view approaches like \textsc{MulMON} \citep{li2020learning} to dynamic scenes by disentangling object motion and viewpoint changes, assuming one dominates in adjacent frames. However, \textsc{DyMON} relies on viewpoint annotations, limiting its utility in unsupervised settings. Temporal methods such as \textsc{SIMONe} \citep{luo2024unsuperviseduocf} address this by leveraging temporal coherence across multi-view videos, using spatial and temporal positional embeddings to disentangle object and viewpoint representations. Yet, SIMONe’s reliance on temporal continuity restricts its generalizability to scenarios where such coherence is absent. In contrast, our framework does not assume temporal dependencies.

\section{Algorithm}
Here we illustrate all the steps involved in the of proposed method \textsc{VISA}, refer \ref{alg:the_alg}.
\begin{algorithm}[h]
  \caption{View Invariant Slot Attention \textsc{VISA}}
  \label{alg:the_alg}
  \begin{algorithmic}[1]
    \STATE {\textbf{Input:} $A \in [V]$, $\rvz^A = \{ f_e(\rvx^v) \; \forall v \in A\} \in \mathbb{R}^{|A| \times N \times d}$ \hfill {\color{gray}$\triangleright$ input representations}} 
    \STATE {\textbf{View:} $\rvv^A = \{ \rvv^v \sim \gN(\rvv^v; \bmu(\rvz^v), \bsigma^2(\rvz^v)) \; \forall v \in A\} \in \mathbb{R}^{|A| \times d}$ \hfill {\color{gray}$\triangleright$ view representations}} 
    \STATE {\textbf{View Transformation:} $\theta^A = \{ \theta^v = \mathrm{STN}(\rvv^v) \; \forall v \in A\} \in \mathbb{R}^{|A| \times 2 \times 3}$ \hfill {\color{gray}$\triangleright$ transformation parameters}} 
    
    \STATE $\mathrm{key}^A \gets \mW_k \gT_{\theta^v}^{-1}(\rvz^A) \in \mathbb{R}^{|A| \times N \times d}, \mathrm{value}^A \gets \mW_v \gT_{\theta^v}^{-1}(\rvz^A) \in \mathbb{R}^{|A| \times N \times d}$ \hfill {\color{gray}$\triangleright$ optional $\mathrm{value} \coloneqq \mathrm{key}$} 
    
    \STATE $\rvs \gets \emptyset$; $\hat{\bpi} \gets \emptyset$
    
    \FOR{$v \in A$}
      \STATE $\forall k$, $\bpi(0)_k \gets 1/K$, $\bmu(0)_k \sim \mathcal{N}(0, \mathbf{I}_d)$, $\bsigma(0)_k^2 \gets \mathbf{1}_d$
      \FOR{$t=0 \to T-1$}
        \STATE \quad $A_{nk} \gets \frac{\bpi(t)_k\mathcal{N}\left(\mathrm{key}_n;\mW_q \bmu(t)_k, \bsigma(t)_k^2\right)}{\sum_{j=1}^{K}\bpi(t)_j\mathcal{N}\left(\mathrm{key}_n;\mW_q \bmu(t)_j, \bsigma(t)_j^2\right)}$ \hfill {\color{gray}$\triangleright$ compute attention}
        \STATE \quad $\hat{A}_{nk} \gets \frac{A_{nk}}{\sum_{l=1}^N A_{lk}}$ \hfill {\color{gray}$\triangleright$ normalize attention}
        \STATE \quad $\bmu(t+1)_k \gets \sum_{n=1}^N \hat{A}_{nk} \cdot \mathrm{value}_n$ \hfill {\color{gray}$\triangleright$ update slot mean}
        \STATE \quad $\bsigma(t+1)_k^2 \gets \sum_{n=1}^N \hat{A}_{nk} \cdot \left(\mathrm{value}_n - \bmu(t+1)_k\right)^2$ \hfill {\color{gray}$\triangleright$ update slot variance}
        \STATE \quad $\bpi(t+1)_k \gets \frac{1}{N} \sum_{n=1}^N A_{nk}$ \hfill {\color{gray}$\triangleright$ update mixing coefficient}
      \ENDFOR
      \STATE $\rvs \gets \rvs \cup \{(\bmu_{1:K}(T), \bsigma^2_{1:K}(T))\}$; $\hat{\bpi} \gets \hat{\bpi} \cup \{\bpi_{1:K}(T)\}$ \hfill {\color{gray}$\triangleright$ slot collection}
    \ENDFOR
    \STATE \textbf{return} $\mathrm{ConvexCombination}(\rvs, \hat{\bpi})$ \hfill {\color{gray}$\triangleright$ $K$ view invariant content}
  \end{algorithmic}
\end{algorithm}

\section{Datasets}
\subsection{Illustrative dataset}
\label{subsec:illustrative_dataset}

\begin{figure}
    \centering
    \includegraphics[trim={0 0 2.5cm 0},clip, width=\columnwidth]{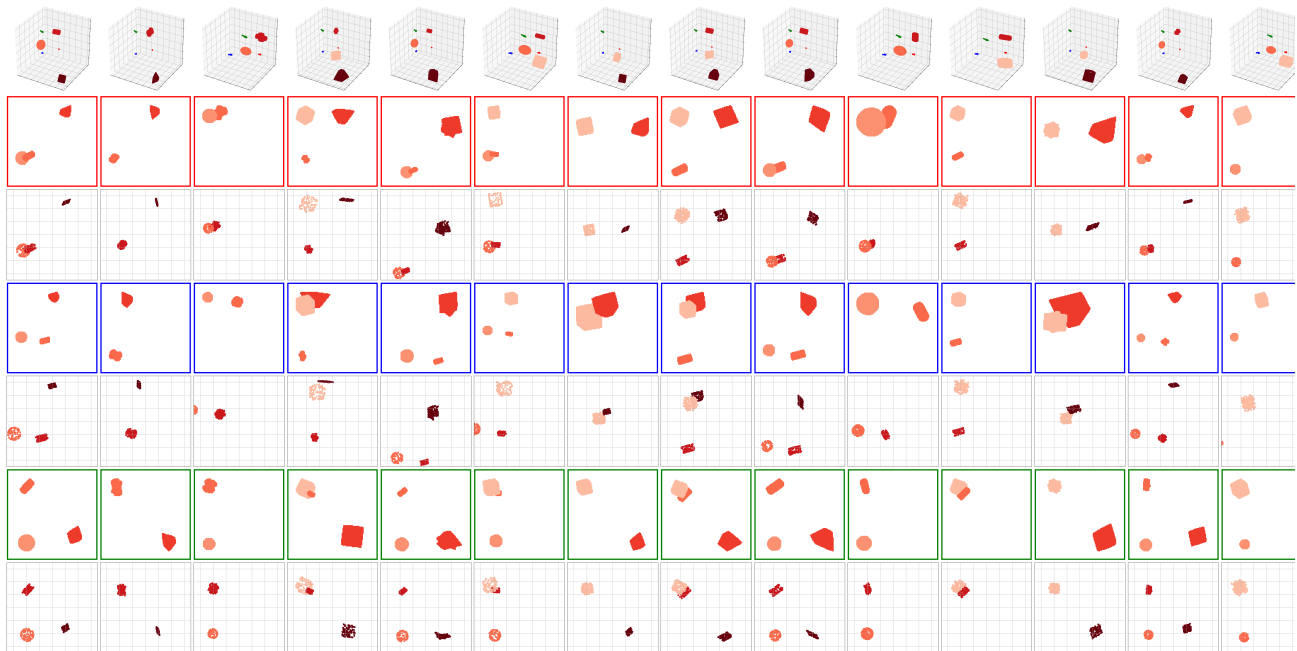}
    \caption{
    \textbf{Data generating process:} The figure illustrates 3D point cloud data in the first row, with camera location highlighted in \textcolor{red}{red}, \textcolor{blue}{blue}, and \textcolor{green}{green} arrow. Following rows indicates projected images and point cloud as observed from \textcolor{red}{red}, \textcolor{blue}{blue}, and \textcolor{green}{green} cameras, respectively.
    }
    \label{fig:exampleDGP}
\end{figure}

To visually illustrate the effectiveness of our theory we experiment with two dimensional illustrative dataset.
For this, similar to \cite{kori2024identifiable}, we defined a $K=5$ component GMM, with differing mean parameters $\bmu = \{\bmu_1, \dots, \bmu_5\}$, and shared isotropic covariances, which we use to sample locations for an object. 
For a given location, we randomly select one object from \texttt{\{`cube', `cylinder', `pyramid', `sphere'\}} and generate 1000 random points on the surface of the selected shape uniformly covering it.
To create a single data point, we randomly select three of the five locations and place a randomly selected object at the location.
To include multiple viewpoints, we consider $V$ camera location and project the objects, creating $V$ different scenes.
We then fill this by considering convex hull operation resulting in projected images as illustrated in Fig. \ref{fig:exampleDGP}.
To maintain uniformity, we only use imaging modality in the main paper while also demonstrating point cloud illustrations here in the appendix.
We use different colours representing different objects in Fig. \ref{fig:identifiability_results_analysis}, \ref{fig:invariance_results_analysis} and used $10,000$ data points in total to train our toy models. 
Unlike existing benchmark datasets, here we remove all the confounding effects caused by lighting and depth. This provides an ideal test bed to validate all our theoretical claims.

\subsection{Proposed dataset} 
In this work, we introduce the \textsc{mv-MoVi} datasets, created using Kubric \cite{greff2022kubric}, which feature multi-view scenes with segmentation annotations. 
We propose two variants of the dataset: \textsc{mv-MoViC}, where the camera locations for every viewpoint remain fixed across all scenes, and \textsc{mv-MoViD}, where the camera locations dynamically change for each scene.

Both \textsc{mv-MoViC} and \textsc{mv-MoViD} primarily consist of scenes generated by randomly selecting a background from a set of \textbf{458} available options and choosing $K$ objects, where $3 \leq K \leq 6$, from a pool of \textbf{930} objects. 
In total, a significantly high number of images can be generated in general. In contrast, for this work, we generate \textbf{72,000} scenes, each captured from 5 different viewpoints, with object segmentation masks for every view to facilitate the evaluation of model performance.
In the case of \textsc{mv-MoViC}, the locations of all five cameras are fixed across the $72,000$ scenes, while in \textsc{mv-MoViD}, the camera positions are dynamically sampled and vary across scenes.

\section{Mask Generation}
\label{appendix:mask_generation}
In the case of additive decoders, the decoder outputs $K$ three channelled tensors along with $K$ single channelled mask.
We consider normalising these masks with \texttt{softmax} transformation along slot dimension, ensuring that each pixel only contributes to a single slot.
The resulting softmaxed masks are used in composing ($\mathrm{image} = \sum_k \mathrm{mask}_k \cdot \mathrm{image}_k$) the slots to reconstruct an image for training.
During inference, we normalise masks with \texttt{sigmoid} transformation, allowing us to estimate occluded objects visually, resolving the spatial ambiguities with occluded objects. 
In a later section, we illustrate the results with both \texttt{softmax} and \texttt{sigmoid} transformations.

\subsection{Additivity Implications}
\begin{definition}(Additive models)
\label{dfn:additivemodels}  
Function $f$ is considered to be an additive decoder if, for any object decoders $f_\mathrm{obj}$ and masking mechanism $m_{\mathrm{obj}}$, if they can be expressed as:

\begin{align}
    f(\rvz) = \sum_{k \in [K]} m_{\mathrm{obj}}(\rvz_k) \odot f_{\mathrm{obj}}(\rvz_k)
\end{align}
\end{definition}

As pointed out in \cite{lachapelle2023additive}, \texttt{softmax}-based masks do not truly fall under the category of additive decoders due to the competition between masks for groups of pixels. This implies that the additive decoders studied in \cite{lachapelle2023additive} are not expressive enough to represent the “masked decoders” typically employed in object-centric representation learning. The issue arises from the normalization of alpha masks, and care must be taken when extrapolating the findings from \cite{lachapelle2023additive} to the models used in practice.

Although \texttt{sigmoid}-based masks satisfy the condition of additivity during inference, it is important to note that the model is still trained using \texttt{softmax} normalization in our setting. The effect of using \texttt{sigmoid} masks during inference can be visually observed in App. \ref{appendix:exps}.

\section{Proofs}
\label{appendix:proofs}

\begin{lemma}[ELBO ]
    With prior distributions $p(\rvv)$ and $p(\rvc)$ for view and content latent random variables, the likelihood $p(\rvx)$ can be maximised by maximising the following expression:  

    \begin{align}
         \log p(\rvx)
        & \geq \mathbb{E}  \log p(\rvx \mid \gT_{\theta^v}(\rvc), \rvv) - \mathrm{KL}\left(q(\rvv \mid \rvx)  \parallel p(\rvv) \right) := \mathrm{ELBO}(\rvx)
    \end{align}
    \label{lemma:elbo_derivation}
\end{lemma}

\begin{proof}

    Considering the generative model in Eqn. \ref{eqn:likelihood} respecting the graphical model in Fig. \ref{fig:generative_model}, we get:
    
    \begin{align}
        p(\rvx) &= \iint p(\rvx^A \mid   \gT_{\theta^v}(\rvc), \rvv^A) \; p(\rvc) \; p(\rvv^A) \; d\rvv \; d\rvc \\
        \log p(\rvx) &= \log \iint p(\rvx^A \mid \rvc_{1:K}, \rvv^A) \; p(\rvc_{1:K}) \; p(\rvv^A) \frac{q(\rvv, \rvc \mid \rvx^A)}{q(\rvv, \rvc \mid \rvx^A)} \; d\rvv \; d\rvc_{1:K}  \\
        & \geq \iint q(\rvv^A \mid \rvx^A) q(\rvc_{1:K} \mid \gT_{\theta^v}^{-1}(\rvx^A)) \log p(\rvx^A \mid \gT_{\theta^v}(\rvc)_{1:K}, \rvv^A)  \frac{p(\rvv^A_{1:K})}{q(\rvv^A \mid \rvx^A)} \frac{p(\rvc^A_{1:K})}{q(\rvc_{1:K} \mid \rvx^A)} \;  \; d\rvv^A \; d\rvc_{1:K} \\
        &= \sum_{v \in A} \iint  q(\rvv^v \mid \rvx^v) q(\rvc_{1:K} \mid \gT_{\theta^v}^{-1}(\rvx^v))) \log p(\rvx^v \mid \gT_{\theta^v}(\rvc)_{1:K}, \rvv^v)  
        \frac{p(\rvv^A_{1:K})}{q(\rvv^A \mid \rvx^v)}
        \frac{p(\rvc^A_{1:K})}{q(\rvc_{1:K} \mid \rvx^v)} \;  \; d\rvv^v \; d\rvc_{1:K} 
    \end{align}

    Given the iterative update for $\rvc$ with EM algorithm, ideally we expect posterior to converge to prior, which results in:

    \begin{align}
        \log p(\rvx)  &= \sum_{v \in A} \iint  q(\rvv^v \mid \rvx^v) q(\rvc_{1:K} \mid \gT_{\theta^v}^{-1}(\rvx^v))) \log p(\rvx^v \mid \gT_{\theta^v}(\rvc)_{1:K}, \rvv^v)  
        \frac{p(\rvv^A_{1:K})}{q(\rvv^A \mid \rvx^v)} \;  \; d\rvv^v \; d\rvc_{1:K} \\
        & = \sum_{v \in A} \mathbb{E}_{\rvc, \rvv}  \log p(\rvx^v \mid \gT_{\theta^v}(\rvc), \rvv) - \mathrm{KL}\left(q(\rvv \mid \rvx^v)  \parallel p(\rvv) \right)
    \end{align}

    Given the subscript notation, the above expression can also be expressed as:
    \begin{align}
        \mathbb{E}_{\rvc, \rvv}  \log p(\rvx \mid \gT_{\theta^v}(\rvc), \rvv) - \mathrm{KL}\left(q(\rvv \mid \rvx)  \parallel p(\rvv) \right) := \mathrm{ELBO}(\rvx)
    \end{align}
\end{proof}

\begin{lemma}[Mean GMM]
Let \(\rvz \in \mathbb{R}^{N \times d}\) be a random variable drawn from a GMM with \(K\) components:
\begin{equation}
\rvz \sim \sum_{k=1}^K \bpi_k \mathcal{N}(\rvz; \bmu_k, \bSigma_k),
\end{equation}
where \(\bpi_k\) are the mixture weights, \(\bmu_k \in \mathbb{R}^d\) are the mean vectors, and \(\bSigma_k \in \mathbb{R}^{d \times d}\) are the covariance matrices. Assuming the mixture satisfies the ICA assumption, such that the components of \(\rvz\) are statistically independent.
A projected random variable \(\bar{z}\) as the average over the dimensions of \(\rvz\):
\begin{equation}
\bar{z} = \frac{1}{d} \sum_{j=1}^d \rvz_{j},
\end{equation}
is also distributed according to a GMM with \(K\) components, with appropriately transformed means and variances.

\label{lemma:projected_GMM}
\end{lemma}

\begin{proof}

Given the random variable \(\rvz\) follows a GMM, so its density can be expressed as:
\begin{equation}
p(\rvz) = \sum_{k=1}^K \bpi_k \mathcal{N}(\rvz; \bmu_k, \bSigma_k),
\end{equation}
where:
\begin{align}
&& \bmu_k = [\mu_{k,1}, \mu_{k,2}, \dots, \mu_{k,d}]^\top; &&
\bSigma_k = \text{diag}([\sigma_{k,1}^2, \sigma_{k,2}^2, \dots, \sigma_{k,d}^2]). &&
\end{align}

Considering, the projection of \(\rvz\) onto \(\bar{z}\) is defined as:
\begin{equation}
\bar{z} = \frac{1}{d} \sum_{j=1}^d \rvz_{j}.
\end{equation}
Given the ICA assumption, the components \(\rvz_{:,j}\) are independent. For a fixed component \(k\), the projected mean and variance of \(\bar{z}\) can be derived as:
\begin{align}
&& \mathbb{E}[\bar{z}] = \frac{1}{d} \sum_{j=1}^d \mu_{k,j}; && 
\text{Var}(\bar{z}) = \frac{1}{d^2} \sum_{j=1}^d \sigma_{k,j}^2. &&
\end{align}

Since the projection \(\bar{z}\) is a linear combination of independent Gaussian variables, \(\bar{z}\) remains Gaussian for each component \(k\). Thus, the overall distribution of \(\bar{z}\) is also a GMM:
\begin{equation}
p(\bar{z}) = \sum_{k=1}^K \bpi_k \mathcal{N}(\bar{z}; \mu_{\bar{z}, k}, \sigma_{\bar{z}, k}^2),
\end{equation}
where:
\begin{align}
&& \mu_{\bar{z}, k} = \frac{1}{d} \sum_{j=1}^d \mu_{k,j}; &&
\sigma_{\bar{z}, k}^2 = \frac{1}{d^2} \sum_{j=1}^d \sigma_{k,j}^2. &&
\end{align}

This concludes the proof.

\end{proof}

\begin{lemma}[Convex Combination of GMMs]
    Let $\rvs^1 = \{\rvs^1_1, \dots, \rvs^1_K\}$ and $\rvs^2 = \{\rvs^2_1, \dots, \rvs^2_K\}$ be two sets 
    of $K$ random vectors in $\mathbb{R}^d$, each distributed according to GMMs:
    \begin{align}
        && \rvs^1 \sim \sum_{k=1}^K \bpi_{1,k} \mathcal{N}(\bmu_{1,k}, \bSigma_{1,k}) ; \quad \rvs^2 \sim \sum_{k=1}^K \bpi_{2,k} \mathcal{N}(\bmu_{2,k}, \bSigma_{2,k}) &&
    \end{align}
    where $\bmu_{i,k} \in \mathbb{R}^d$, $\bSigma_{i,k} \in \mathbb{R}^{d \times d}$, and $\bpi_{i,k}$ are 
    the means, covariances, and mixing coefficients respectively.
    
    Then for any weights $w_1, w_2 \in \mathbb{R}$ such that $w_1 + w_2 = 1$, the convex combination 
    $\rvs = w_1\rvs^1 + w_2\rvs^2$ is also distributed according to a GMM with $K$ components.
    \label{lemma:GMMconvexcombination}
\end{lemma}

\begin{proof}
    Without loss of generality, assume the components of both GMMs are aligned. 
    For each component $k$, we derive the parameters of the resulting mixture:

    The mixing coefficients of the resulting GMM are weighted combinations of the original coefficients:
    \begin{align}
        \tilde{\bpi}_k = w_1\bpi_{1,k} + w_2\bpi_{2,k}
    \end{align}

    For each component $k$, the convex combination of Gaussians results in a Gaussian distribution. 
    The mean of the resulting Gaussian is:
    \begin{align}
        \tilde{\bmu}_k = \frac{w_1\bpi_{1,k}\bmu_{1,k} + w_2\bpi_{2,k}\bmu_{2,k}}{\tilde{\bpi}_k}
    \end{align}

    The covariance of the resulting Gaussian for each component $k$ can be derived as follows. 
    Firstly, recall that for a random variable $X$, the covariance is:
    \begin{align}
        \mathrm{Var}(X) = \mathbb{E}[(X - \mathbb{E}[X])(X - \mathbb{E}[X])^\top] = \mathbb{E}[XX^\top] - \mathbb{E}[X]\mathbb{E}[X]^\top
    \end{align}

   First lets compute $\mathbb{E}[\rvs_k\rvs_k^\top]$:
    \begin{align}
        \mathbb{E}[\rvs_k\rvs_k^\top] &= \mathbb{E}\left[\left(\frac{w_1\bpi_{1,k}\rvs^1_k + w_2\bpi_{2,k}\rvs^2_k}{\tilde{\bpi}_k}\right)\left(\frac{w_1\bpi_{1,k}\rvs^1_k + w_2\bpi_{2,k}\rvs^2_k}{\tilde{\bpi}_k}\right)^\top\right] \\
        &= \frac{w_1^2\bpi_{1,k}^2\mathbb{E}[\rvs^1_k(\rvs^1_k)^\top] + w_2^2\bpi_{2,k}^2\mathbb{E}[\rvs^2_k(\rvs^2_k)^\top]}{(\tilde{\bpi}_k)^2} \\
        &\quad + \frac{w_1w_2\bpi_{1,k}\bpi_{2,k}}{(\tilde{\bpi}_k)^2}\mathbb{E}[\rvs^1_k(\rvs^2_k)^\top + \rvs^2_k(\rvs^1_k)^\top]
    \end{align}

    Then, substitute known expectations:
    \begin{align}
        \mathbb{E}[\rvs^i_k(\rvs^i_k)^\top] &= \bSigma_{i,k} + \bmu_{i,k}\bmu_{i,k}^\top \\
        \mathbb{E}[\rvs^1_k(\rvs^2_k)^\top] &= \bmu_{1,k}\bmu_{2,k}^\top
    \end{align}

    Finally, by subtract $\mathbb{E}[\rvs^i_k
    ]\mathbb{E}[\rvs^i_k]^T =  \tilde{\bmu}_k\tilde{\bmu}_k^\top$ we get the covariance:
    \begin{align}
        \tilde{\bSigma}_k &= \frac{w_1^2\bpi_{1,k}^2\bSigma_{1,k} + w_2^2\bpi_{2,k}^2\bSigma_{2,k}}{(\tilde{\bpi}_k)^2}
    \end{align}

    To verify this forms a non degenerate GMM, we show the mixing coefficients sum to 1:
    \begin{align}
        \sum_{k=1}^K \tilde{\bpi}_k &= \sum_{k=1}^K (w_1\bpi_{1,k} + w_2\bpi_{2,k}) \\
        &= w_1\sum_{k=1}^K \bpi_{1,k} + w_2\sum_{k=1}^K \bpi_{2,k} \\
        &= w_1 \cdot 1 + w_2 \cdot 1 = 1
    \end{align}

    Therefore, the convex combination results in a valid Gaussian mixture model with $K$ components, where each component has mean $\tilde{\bmu}_k$, covariance $\tilde{\bSigma}_k$, and mixing coefficient $\tilde{\bpi}_k$.
\end{proof}

\paragraph{Lemma \ref{prp:aggregate_poterior1}.}(Optimal Content Mixture)
    For $A \in [V]$, given the a local content distribution $q(\rvc_{1:K} \mid \rvs^A_{1:K}, \rvx^A)$ (per-scene $\rvx^A \in \{ \rvx^A_i\}_{i=1}^M$), which can be expressed as a GMM with $K$ components, the aggregate posterior $q(\rvc)$ is obtained by marginalizing out $\rvx, \rvs$ is a non-degenerate global Gaussian mixture with $MK$ components:
    \begin{align}
        p(\rvc) = q(\rvc) &= \frac{1}{M} \sum_{i=1}^M \sum_{k=1}^K \widehat{\bpi}_{ik} \gN\left(\rvc; \widehat{\bmu}_{ik}, \widehat{\bsigma}^2_{ik} \right).
    \end{align}


\begin{proof}
We extend the proof in \cite{kori2024identifiable}, by incorporating hierarchical slot to aggregate content formalisation.
For which, we begin by noting that the aggregate posterior $q(\rvc)$ is the optimal prior $p(\rvc)$ so long as our posterior approximation $q(\rvc \mid \rvs^A,\rvx^A)$ is close enough to the true posterior $p(\rvc \mid \rvs^A, \rvx^A)$, since for a dataset $\rvx^A \in \{ \rvx^A_i\}_{i=1}^M$, for which we start with $q(\rvs^A \mid \rvx^A)$, wlog, given view point transformation is deterministic, we consider $\rvx^A = \gT_{\theta^A}(\rvx^A)$ we have that:
\begin{align}
    p(\rvs^A) & = \int p(\rvs^A \mid \rvx^A) p(\rvx^A) d\rvx^A 
    \\[0pt] & 
    = \mathbb{E}_{\rvx^A \sim p(\rvx^A)} \left[ p(\rvs^A \mid \rvx^A) \right] 
    \\[0pt] & 
    \approx \frac{1}{M}\sum_{i=1}^M p(\rvs^A \mid \rvx^A_i) \quad \text{(empirical approximation)} 
    \\[0pt] & 
    \approx \frac{1}{M}\sum_{i=1}^M q(\rvs^A \mid \rvx^A_i) \quad \text{(posterior approximation)} 
    \\[0pt] & 
    \eqqcolon q(\rvs^A),
\end{align}
We further extend this to $q(\rvc)$, with the result from Lemma \ref{lemma:GMMconvexcombination}, we know that the $q(\rvc \mid \rvs^A)$ is a GMM with same number of components as $q(\rvs^v \mid \rvs^v)$ for any $v \in [V]$ as follows
\begin{align}
    p(\rvc) & = \int p(\rvc \mid \rvs^A) p(\rvs^A) d\rvs^A 
    \\[0pt] & 
    = \mathbb{E}_{\rvs^A \sim p(\rvs^A)} \left[ p(\rvc \mid \rvs^A) \right] 
    \\[0pt] & 
    \approx \frac{1}{M}\sum_{i=1}^M p(\rvc \mid \rvs^A_i)
    \\[0pt] & 
    \approx \frac{1}{M}\sum_{i=1}^M q(\rvc \mid \rvs^A_i)
    \\[0pt] & 
    \eqqcolon q(\rvc),
\end{align}
where we approximated $p(\rvx)$ using the empirical distribution, then substituted in the approximate posterior, marginalizing $\rvx$ to get $p(\rvs)$, we later consider the distributional form of $p(\rvs)$ and marginalise $\rvs^A$ to get $p(\rvc)$. 
This observation was first made by~\cite{hoffman2016elbo} and was used in \cite{kori2024identifiable} we use it to motivate our setup.
Given our model fits a local GMM to each latent variable sampled from the approximate posterior: $\rvz^A \sim q(\rvz^A \mid \rvx^A_i)$, for $i=1,\dots,M$. Let $f_s(\rvz^A)$ denote the (local) the product of GMM density, its expectation is given by:
\begin{align}
    \mathbb{E}_{p(\rvx^A), q(\rvz^A\mid \rvx^A)} \left[ f_s(\rvz^A)\right] 
    & = \iint p(\rvx^A) q(\rvz^A \mid \rvx^A)f_s(\rvz^A) d\rvx^A d\rvz^A
    \\[5pt] & \approx \iint \frac{1}{M} \sum_{i=1}^M \delta(\rvx^A - \rvx^A_i) q(\rvz^A \mid \rvx^A)f(\rvz^A) d\rvx^A d\rvz^A
    \\[5pt] & = \int \frac{1}{M} \sum_{i=1}^M q(\rvz^A \mid \rvx^A_i)f(\rvz^A) d\rvz^A
    \\[5pt] \nonumber & = \int \frac{1}{M} \sum_{i=1}^M \gN \big(\rvz^A; \bmu(\rvx^A_i), \bsigma^2(\rvx^A_i)\big) \cdot 
    \sum_{k=1}^K \bpi_{k}(\rvx^A_i) \gN\left(\rvz^A; \bmu_{k}(\rvx^A_i), \bsigma^2_{k}(\rvx_i^A \right) d\rvz^A
    \\[5pt] & \approx \int \frac{1}{M} \sum_{i=1}^M \delta(\rvz^A - \bmu(\rvx^A_i)) \cdot \sum_{k=1}^K \bpi_{k}(\rvx^A_i) \gN\left(\rvz^A; \bmu_{k}(\rvx^A_i), \bsigma^2_{k}(\rvx_i^A \right) d\rvz^A \label{eq: dap}
    \\[5pt] & = \frac{1}{M} \sum_{i=1}^M \sum_{k=1}^K \bpi_{k}(\rvx^A_i) \gN\left(\rvz^A; \bmu_{k}(\rvx^A_i), \bsigma^2_{k}(\rvx_i^A \right)\label{eq:mc}
    \\[5pt] & \eqqcolon q(\rvz^A),
\end{align}
where we again used the empirical distribution approximation of $p(\rvx)$, and the following basic identity of the Dirac delta to simplify: $\int \delta(\rvx - \rvx') f_e(\rvx) d\rvx = f_e(\rvx')$. 

For the general case, however, we must instead compute the product of $q(\rvz^A \mid \rvx^A)$ and $f_s(\rvz^A)$ rather than use a Dirac delta approximation as in Eqn.~\ref{eq: dap}. To that end we may proceed as follows w.r.t. to each datapoint $\rvx^A_i$:
\begin{align}
     q(\rvz^A \mid \rvx^A_i) \cdot f_s(\rvz^A) & = \gN\big(\rvz^A; \bmu(\rvx^A_i), \bsigma^2(\rvx^A_i)\big) \cdot \sum_{k=1}^K \bpi_{k}(\rvx^A_i) \gN\left(\rvz^A; \bmu_{k}(\rvx^A_i), \bsigma^2_{k}(\rvx_i^A \right)
     \\ & = \sum_{k=1}^K \bpi_{k}(\rvx^A_i) \left[\gN\big(\rvz^A; \bmu(\rvx^A_i), \bsigma^2(\rvx^A_i)\big) \cdot \gN\left(\rvz^A; \bmu_{k}(\rvx^A_i), \bsigma^2_{k}(\rvx_i^A \right) \right]
\end{align}
Given that means across all views are aligned, similar to Lemma \ref{lemma:GMMconvexcombination}, we know the resulting combined GMM has same number of components:
\begin{align}
    q(\rvz^A \mid \rvx^A_i) \cdot f_s(\rvz^A) & = \prod_{v=1}^{|A|}\sum_{k=1}^K \bar{\bpi}^v_{ik} \gN \left(\rvz; \bar{\bmu}_{ivk}, \bar{\bsigma}^2_{ivk} \right),
\end{align}
Given the product of GMM is a GMM with the number of components equal to the product of a number of components in individual GMM, however in our setting we consider all the components in individual GMM across viewpoints to be aligned resulting in GMM with a number of components equal to the sum of individual components which in our case correspond to $K$.
The posterior parameters of the resulting mixture are given in closed form by:
\begin{align}
    &&\bar{\bsigma}^2_{ivk} = \left(\frac{1}{{\bsigma}^2_{k}(\rvx^v_i)} + \frac{1}{{\bsigma}^2(\rvx^v_i)}\right)^{-1},
    && \bar{\bmu}_{ivk} = \bar{\bsigma}^2_{ivk} \left( \frac{\bmu(\rvx^v_i)}{\bsigma^2(\rvx^v_i)} + \frac{\bmu_{k}(\rvx^v _i)}{\bsigma^2_{k}(\rvx^v_i)}\right),&&
\end{align}
The resulting GMM is still on the view-specific slots, the aggregation of these slots to obtain content vectors marginalises the viewpoint-level information with convex combination of parameters across all the viewpoints considered as described in cf. \ref{lemma:GMMconvexcombination}, results in:
\begin{align}
    \prod_{v=1}^{|A|} \sum_{k=1}^K \bar{\bpi}^v_{ik} \gN \left(\rvz; \bar{\bmu}_{ivk}, \bar{\bsigma}^2_{ivk} \right)  = \sum_{k=1}^K \widehat{\bpi}_{ik} \gN \left(\rvz; \widehat{\bmu}_{ik}, \widehat{\bsigma}^2_{ik} \right),
\end{align}
\begin{align}
    & \widehat{\bsigma}^2_{ik} = g(\bar{\bsigma}_{ik}, \bar{\bpi}_{ik}) = \sum_{v=1}^{|A|}\left( \frac{\bar{\bpi}^v_{ik}}{\sum_{v=1}^{|A|} \bar{\bpi}_{ik}^v} \right)^2 \bar{\bsigma}^2_{ik}, \\
    & \widehat{\bmu}_{ivk} = g(\bar{\bmu}_{ik}, \bar{\bpi}_{ik}) = \sum_{v=1}^{|A|} \frac{\bar{\bpi}^v_{ik}}{\sum_{v=1}^{|A|} \bar{\bpi}_{ik}^v} \bar{\bmu}_{ik},
\end{align}

Now to show that the resulting GMM is non-degenerate we need to show $\sum_{k=1}^K \widehat{\bpi}_{ik} = 1, \mathrm{for} \ i=1,2,\dots, M$. Based on Eqn. \ref{eq:mc}:
\begin{align}
    &\implies \frac{1}{M} \sum_{i=1}^{M} \sum_{k=1}^K \widehat{\bpi}_{ik} = \frac{1}{M|A|} \sum_{i=1}^{M} \sum_{k=1}^K \sum_{v=1}^{|A|} \bar{\bpi}^v_{ik} =  \frac{1}{M|A|} \sum_{i=1}^{M} |A| = \frac{1}{M|A|} \cdot M |A| = 1,\\
    & \implies \frac{1}{M} \sum_{i=1}^{M} \sum_{k=1}^K \widehat{\bpi}_{ik} = 1. &&
\end{align}

based on the above equation we can say that the scaled sum of the mixing proportions of all $K$ components in all $M$ GMMs when the components are aligned must equal $1$, show that the resulting aggregate posterior is non-degenerate and a valid probability distribution.
\end{proof}

\begin{assumption}[\textit{Weak} Injectivity] Let $f: \mathcal{Z} \to \mathcal{X}$ be a mapping between latent space and image space, where $\mathrm{dim}(\mathcal{Z}) \leq \mathrm{dim}(\mathcal{X})$. The mapping $f_d$ is weakly injective if there exists $\mathbf{x}_0 \in \mathcal{X}$ and $\delta > 0$ such that $|f^{-1}(\{\mathbf{x}\})| = 1$, $\forall \mathbf{x} \in B(\mathbf{x}_0, \delta) \cap f(\mathcal{Z})$, and $\{\mathbf{x} \in \mathcal{X} : |f^{-1}(\{\mathbf{x}\})| = \infty\} \subseteq f(\mathcal{Z})$ has measure zero w.r.t. to the Lebesgue measure on $f(\mathcal{Z})$ (cf.~\cite{kivva2022identifiability}).
    \label{ass:weak_inj}
\end{assumption}
    
\begin{remark}
    In words, Assumption~\ref{ass:weak_inj} says that a mapping $f_d$ is weakly injective if: (i) in a small neighbourhood around a specific point $\mathbf{x}_0 \in \mathcal{X}$ the mapping is injective -- meaning each point in this neighbourhood maps to exactly one point in the latent space $\mathcal{Z}$; and (ii) while $f_d$ may not be globally injective, the set of points in $\mathcal{X}$ that map back to an infinite number of points in $\mathcal{Z}$ (non-injective points) is almost non-existent in terms of the Lebesgue measure on the image of $\mathcal{Z}$ under $f_d$.
\end{remark}
\begin{theorem}[Mixture of Concatenated Slots]
    Let $f_s$ denote a permutation equivariant probabilistic slot attention function such that $f_s( \rvz^v, P \rvs^v) = P f_s(\rvz^v, \rvs^v)$, where $P \in \{0, 1\}^{K \times K}$ is an arbitrary permutation matrix. Let $\rvc = (g(\rvs^A_1, .), \dots, g(\rvs^A_K, .)) \in \mathbb{R}^{Kd}$ be the concatenation of $K$ individual content vectors, where each vector is an aggregate of all the slots obtained from considered viewpoints in a viewpoint-set $A \subseteq [V]$ (cf. \cite{kori2024identifiable}).
    Due to the nature of the aggregator function, the individual content vector is Gaussian distributed within a $K$-component mixture: $\rvc_k \sim \mathcal{N}(\bmu_k, \bSigma_k) \in \mathbb{R}^d, \forall k \in \{1, \dots K\}$. Then, $\rvc$ is also GMM distributed with $K!$ mixture components:
    \begin{align}
        p(\rvc) = \sum_{p=1}^{K!} \bpi_p \gN(\rvc; \bmu_p, \bSigma_p), \text{ where } \bpi \in \Delta^{K!-1}, \bmu_p \in \mathbb{R}^{Kd}, \bSigma_p \in \mathbb{R}^{Kd \times Kd}.
    \end{align}
    \label{thm:concatenated_mixture}
\end{theorem}

We additionally borrow some theorems and definitions from \cite{kivva2022identifiability} which are essential for our proofs.
First, we restate the definition of a \textit{generic point} as outlined by \cite{kivva2022identifiability} below.
\begin{definition}(Generic point)
    A point $\rvx\in f_d(\mathbb{R}^m)\subseteq \mathbb{R}^n$ is generic if there exists $\delta>0$, such that $f_d:B(\rvs, \delta)\rightarrow \mathbb{R}^n$ is affine for every $\rvs\in f_d^{-1}(\{\rvx\})$
\end{definition}
\begin{theorem}[Kivva et al.~\cite{kivva2022identifiability}]
    \label{thm:genericfunctions}
    Given $f_d:\mathbb{R}^m\rightarrow \mathbb{R}^n$ is a piecewise affine function such that $\{\rvx\in \mathbb{R}^n\, :\, |f_d^{-1}(\{\rvx\})| = \infty\}\subseteq f_d(\mathbb{R}^m)$ has measure zero with respect to the Lebesgue measure on $f_d(\mathbb{R}^m)$, this implies $\dim (f_d(\mathbb{R}^m)) = m$ and almost every point in $f_d(\mathbb{R}^m)$ (with respect to the Lebesgue measure on $f_d(\mathbb{R}^m)$) is generic with respect to~$f_d$. 
\end{theorem}
\begin{theorem}[Kivva et al.~\cite{kivva2022identifiability}]
\label{thm:local-gmm-iden}
Consider a pair of finite GMMs  in $\mathbb{R}^m$:
\begin{align}
    &&\rvy =\sum_{j=1}^{J} \bpi_{j}\gN(\rvy; \bmu_{j}, \bSigma_{j}), &&\text{and} &&\rvy'=\sum_{j=1}^{J'} \bpi_{j}'\gN(\rvy'; \bmu'_{j}, \bSigma'_{j}).&&
\end{align}
Assume that there exists a ball $B(\rvx, \delta)$ such that $\rvy$ and $\rvy'$ induce the same measure on $B(\rvx, \delta)$. Then $\rvy \equiv \rvy'$, and for some permutation $\tau$ we have that $\bpi_i = \bpi'_{\tau(i)}$ and $(\bmu_i, \bSigma_i) = (\bmu'_{\tau(i)}, \bSigma'_{\tau(i)})$.
\end{theorem}
\begin{theorem}[Kivva et al.~\cite{kivva2022identifiability}]
    \label{thm:genericfunctions2}
        Given $\rvz~\sim \sum_{i=1}^{J} \bpi_i\gN(\rvz; \bmu_i, \bSigma_i)$ and $\rvz' \sim \sum_{j=1}^{J'} \bpi_j'\gN(\rvz'; \bmu_j', \bSigma_j')$ and $f_d(\rvz)$ and $\tilde{f}_d(\rvz')$ are equally distributed. We can assume for $\rvx \in \mathbb{R}^n$ and $\delta>0$, $f_d$ is invertible on $B(\rvx, 2\delta)\cap f_d(\mathbb{R}^{m})$.
        This implies that there exists $\rvx_1\in B(\rvx, \delta)$ and $\delta_1>0$ such that both $f_d$ and $\tilde{f}_d$ are invertible on $B(\rvx_1, \delta_1)\cap f_d(\mathbb{R}^{m})$.
\end{theorem}

\paragraph{Theorem \ref{thm:affine_equivalence}}(Affine Equivalence of aggregate content)
    For any subset $A \subseteq [V]$, such that $|A| > 0$ , given a set of images $\rvx^A \in \gX^A$ and a corresponding aggregate content $\rvc \in \gC$ and a non-degenerate content posterior $q(\rvc \mid \rvs^A)$, considering two mixing function $f_{d}, \tilde{f}_d$ satisfying assumption \ref{ass:weak_inj}, with a shared image, then $\rvc$ are identifiable up to $\sim_s$ equivalence.
    

\begin{proof}
    Based on the results of \cite{kori2024identifiable} we know that when $p(\rvs)$ is aggregate posterior of $q(\rvs \mid \rvx)$, $p(\rvs)$ is identifiable up to $\sim_s$ equivalence. Additionally, based on lemma \ref{prp:aggregate_poterior1} we know that both $q(\rvs \mid \rvx)$ and $q(\rvc \mid \rvs)$ are a non-degenerate GMM with valid probability distribution. 
    Using similar arguments in \cite{kori2024identifiable, kivva2022identifiability} we show that $p(\rvc)$ and $p(\rvs)$ are identifiable up to $\sim_s$ equivalence.
    \texttt{W.l.o.g}, given view point transformation is deterministic, we consider $\rvx^A = \gT_{\theta^A}(\rvx^A)$.
    
    We know that 
    \begin{align}
        p(\rvs^A) &= \int q(\rvs^A_{1:K} \mid \rvx^A) p(\rvx^A) d\rvx^A \\
        & = \int \prod_{v \in A} q(\rvs^v \mid \rvx^v) p(\rvx^v) d\rvx^A \\
        & = \int \prod_{v \in A} \left(\sum_{k=1}^K \bpi^v_k \gN\left(\rvs^v; \bmu_{k}(\rvx^v), \bsigma^2_{k}(\rvx^v)\right)\right)  p(\rvx^v) d\rvx^A  \\
        & = \prod_{v \in A} \frac{1}{|\gX|} \int \left(\sum_{k=1}^K \bpi^v_k \gN\left(\rvc^v; \bmu_{k}(\rvx^v), \bsigma^2_{k}(\rvx^v)\right)\right) \delta(\rvx^v - \rvx^v_i) d\rvx^A  \\
        & = \prod_{v \in A} \left(\sum_{k=1}^{|\gX| K} \frac{1}{|\gX|} \hat{\bpi}^v_{ik}\gN\left(\rvs^v; \hat{\bmu}_{ivk}, \tilde{\bsigma}^2_{ivk}\right)\right)
        \label{eqn:content_prior}
    \end{align}

    Change of variables from $\rvs$ to $\rvc$ to get prior over random variable $\rvc$, with matching function $g$, results in:
    \begin{align}
        p(\rvc_{1:K}) &= \int p(\rvs_{1:K}^A)\delta \left(\rvs^A_{1:K} - g(\rvs^A_{1:K}, \bpi^A_{1:K}) \right) d \rvc^A_{1:K}
        \label{eqn:change_of_variables}
    \end{align}

    Given the transformation $g$ is linear, resulting us with the distribution with mean given by:
    \begin{align}
        \mathbb{E}_{\rvc}\left(\rvc_{1:K}\right) &= \mathbb{E}_{\rvs}\left(g(\rvs^A_{1:K}, \bpi_{A,1:K}) \right)\\ 
                                                 &= g\left(\mathbb{E}_{\rvs}(\rvs^A_{1:K}), \bpi^A_{1:K} \right)\\
                                                 &= \sum_{v \in A} \frac{\bpi^v_{ 1:K}}{\sum_{v \in A}\bpi^v_{1:K}}\mathbb{E}_{\rvs}(\rvs^A_{1:K})
        \label{eqn:prior_s_mean}
    \end{align}
    
    and the covariance follows the diagonal structure as in $p(\rvc)$, which can be described as follows: 
    \begin{align}
        \mathbb{V}\mathrm{ar}(\rvc_{1:K}) = \sum_{v \in A} \left(\frac{\bpi^v_{1:K}}{\sum_{v \in A}\bpi^v_{1:K}}\right)^2 \mathbb{V}\mathrm{ar}_{\rvc}(\rvc^A_{1:K})
         \label{eqn:prior_s_variance}
    \end{align}
    
    Finally, the mixture components can be expressed as:
    \begin{align}
        \tilde{\bpi}_{1:K} = \frac{\sum_{v \in A} {\bpi}^v_{1:K}}{|A|}
        \label{eqn:prior_s_pis}
    \end{align}

    With distribution parameters described in equations \ref{eqn:prior_s_mean}, \ref{eqn:prior_s_variance}, and \ref{eqn:prior_s_pis}, we define the aggregate content distribution as GMM expressed as follows:
        \begin{equation}
            p(\rvc) = \sum_{k=1}^{|\gX| K} \frac{1}{|\gX|} \tilde{\bpi}^v_{k} \gN\left(\rvv; \mathbb{E}(\rvc)_k), \mathbb{V}\mathrm{ar}(\rvc)_k\right)
            \label{eqn:prior_s}
        \end{equation}

    \paragraph{Validity of $p(\rvc)$:}  The outer summation in equation \ref{eqn:prior_s} can be split into two one for image samples and other for original mixing coefficients, which results in the equation:
    \begin{equation}
             p(\rvc) = \sum_{i=1}^{|\gX| } \sum_{k=1}^{K} \frac{1}{|\gX|} \tilde{\bpi}^v_{ik} \gN\left(\rvv; \mathbb{E}(\rvc)_{ik}), \mathbb{V}\mathrm{ar}(\rvc)_{ik}\right)
    \end{equation}
    Based on this we can observe the each component in our GMM corresponds to particular slots for a given image in a given viewpoint, triple describing each component is:
        
    \begin{align}
        && \left\{\tilde{\bpi}^v_{ik}, \tilde{\bmu}_{vik}, \tilde{\bsigma}^2_{vik} \right\},  && \mathrm{for} \quad v = 1,\dots,|A| \quad i = 1,2,\dots,|\gX|, \quad \mathrm{and} \quad k = 1,2,\dots,K. &&
    \end{align}
    To verify that $p(\rvc)$ is a non-degenerate mixture, we observe the following implication:
    \begin{align}
         &&\sum_{i=1}^{|\gX| } \sum_{k=1}^{K} \frac{1}{|\gX|} \frac{\sum_{v \in A} \tilde{\bpi}^v_{ik}}{|A|}   = 1, &&
    \end{align}
    \begin{align}
         && \implies \frac{1}{|\gX|}  \frac{1}{|A|} \sum_{i=1}^{|\gX| } \sum_{v \in A}  \sum_{k=1}^{K} \bpi^v_{ik} = \frac{1}{|\gX|}  \frac{1}{|A|}  |\gX| \cdot |A| \cdot 1 = 1 &&
    \end{align}
    
    similar to lemma \ref{prp:aggregate_poterior1}, this says that the scaled sum of the mixing proportions of all $K$ components in all $|\gX|$ GMMs must equal 1, proving that the associated aggregate posterior mixture $p(\rvc)$ is a well-defined and non degenerate probability distribution.

    \paragraph{Invertibility restrictions:}
    Given two piece-wise affine compositional functions $f_d, \tilde{f}_d: \gC \times \gV  \rightarrow \gX$, for a given set of views $\rvv^A$, let $\rvc = (\rvc_1, \dots, \rvc_K), \ni \rvc_k \sim \mathcal{N}(\rvc_k; \bmu_k, \bSigma_k)$ and $\rvc' = (\rvc'_1, \dots, \rvc'_K), \ni \rvc'_k \sim  \mathcal{N}(\rvc'_k; \bmu_k', \bSigma_k')$ be a pair of aggregate content representations, result of sampling a concatenated higher dimensional GMM distribution in $\mathbb{R}^{Kd}$, as shown in Theorem \ref{thm:concatenated_mixture}, \cite{kori2024identifiable}. 
    In the case when, $f_{d\sharp}(\gC, \{\rvv^A\})$ and $\tilde{f}_{d\sharp}(\gC', \{\rvv^A\})$\footnote{$f_{d\sharp}$ correspond to push forward operation, applying function $f_d$ on all the elements of the given set.} are equally distributed. 
    Now assume that there exists $\rvx^A \in \gX$ and $\delta>0$ such that $f_d$ and $\tilde{f}_d$ are invertible and piecewise affine on $B(\rvx^A, \delta)\cap f_d(\gS)$, for a given set of views $\rvv^A$, which implies $\dim f_d(\gC, \{\rvv^A\}) = |\gC|$.

    \paragraph{Affine subspace:} We now restrict the space $B(\rvx^A, \delta)$ to a subspace $B(\rvx'^A, \delta')$ where $\rvx^A \in B(\rvx'^A, \delta')$ such that $f_d$ and $\tilde{f}_d$ are now invertible and affine on $B(\rvx'^A, \delta')\cap f_d(\gC \times \{\rvv^A\})$.
    With $L \subseteq \gX^A$ be an $|\gC|$-dimensional affine subspace (assuming $|\gX^A| \geq |\gC|$), such that $B(\rvx'^A, \delta')\cap f_{d\sharp}(\gC,\{\rvv^A\}) = B(\rvx'^A, \delta')\cap L$. 
    We also define $h_f, h_{\tilde{f}} : \gC \rightarrow L$ to be a pair of invertible affine functions where $h^{-1}_{f\sharp}(B(\rvx'^A, \delta')\cap L) = f^{-1}_{d\sharp} (B(\rvx'^A, \delta')\cap L; \rvv^A)$ and $h_{\tilde{f}\sharp}^{-1}(B(\rvx'^A, \delta')\cap L) = \tilde{f}_{d\sharp}^{-1}(B(\rvx'^A, \delta')\cap L; \rvv^A)$. 
    Implying $h_f(\rvc)$ and $h_{\tilde{f}}(\rvc')$ are finite GMMs that coincide with $B(\rvx'^A, \delta')\cap L$ and $h_f(\rvc) \equiv h_{\tilde{f}}(\rvc')$, theorem \ref{thm:local-gmm-iden}, \cite{kivva2022identifiability}. 
    Given, $h = h_{\tilde{f}}^{-1}\circ h_f$ and $h_f(\rvc)$ and $h_{\tilde{f}}(\rvc')$ then $h$ is an affine transformation such that $h(\rvc) = \rvc'$.   

    \paragraph{$\sim_s$ equivalence:} 
    Given Theorems~\ref{thm:genericfunctions} and \ref{thm:genericfunctions2}, there exists a point $\rvx \in f_{d\sharp}(\gC, \{\rvv^A\})$ that is generic with respect $f_d$ and $\tilde{f}_d$ and invertible on $B(\rvx, \delta)\cap f_{d\sharp}(\gC, \{\rvv^A\})$. 
    Having established that there is an affine transformation $h(\rvc) = \rvc'$ and invertiblility of piece-wise affine functions $f_d$ and $\tilde{f}_d$ on $B(\rvx^A, \delta)\cap f_{d\sharp}(\gC, \{\rvv^A\})$, this implies that $\rvc$ is identifiable up to an affine transformation and permutation of $\rvc_k \in \rvc$, concluding our proof.

    \paragraph{Remark:} Given Theorem \ref{thm:local-gmm-iden}, we know that for each higher dimensional mixture component in $p(\rvc)$ induces the same measure on $B(\rvx^A, \delta)$ and hence for some permutation $\tau$ we have that $(\bmu_{\pi(i)}, \bSigma_{\pi(i)}) = (\bmu'_{\tau(\pi(i))}, \bSigma'_{\tau(\pi(i))})$. 
    Therefore, each mixture component $\rvc_{\pi(i)}$ is identifiable up to affine transformation, and permutation of aggregate content representations in $\rvc$. 
    Now, given sampling $\rvc_k$ is equivalent to obtaining $K$ samples from the GMM, $q(\rvz)$ and concatenating, this makes $q(\rvz)$ identifiable up to affine transformation, $h$ and permutation of slot representations in $\rvc$. It now trivially follows that each of the aggregate content representation $\rvc_k \sim \gN(\rvc_k; \bmu_k, \bSigma_k) \in \mathbb{R}^d, \forall \, k \in \{1, \dots, K\}$ is identifiable up to affine transformation, $h$ based on the following observed property of GMMs: 
    \begin{align}
        \sum_{k=1}^{K} {\bpi}_{k}h_{\sharp} \left(\gN(\rvs_k; \bmu_{k},\bSigma_{k})\right)
        \sim
        h_{\sharp}\Big(\sum_{k=1}^{K} {\bpi}_{k}\gN(\rvs'_k;\bmu'_{k},\bSigma'_{k})\Big),
    \end{align}
\end{proof}

\paragraph{Theorem \ref{theorem:invariance}}(Invariance of aggregate content)
    For any subset $A, B \subseteq [V]$, such that $|A| > 0, |B| > 0$ and both $A, B$ satisfy an assumption \ref{ass:viewpoint_sufficiency}, we consider aggregate content to be invariant to viewpoints if $f_A \sim_s f_B$ for data $\gX^A \times \gX^B$.
    
\begin{proof}
    Based on equation \ref{eqn:prior_s}, $p_A(\rvs)$ and $p_B(\rvs)$ can be expressed as follows:
    \begin{equation}
        p_A(\rvc) = \sum_{k=1}^{|\gX| K} \frac{1}{|\gX|} \frac{\sum_{v \in A} \bpi^v_{k}}{|A|} \gN\left(\rvc; \sum_{v \in A} \frac{\bpi^v_{k}}{\sum_{v \in A}\bpi^v_{k}}\bmu_{vk}, \sum_{v \in A} \left(\frac{{\bpi}_{vk}}{\sum_{v \in A}\bpi^v_{k}}\right)^2 \bsigma^2_{vk}\right)
    \end{equation}

    \begin{equation}
        p_B(\rvc) = \sum_{k=1}^{|\gX| K} \frac{1}{|\gX|} \frac{\sum_{u \in B} {\bpi}^u_{k}}{|B|} \gN\left(\rvc; \sum_{u \in B} \frac{{\bpi}_{uk}}{\sum_{u \in B}{\bpi}^u_{k}}{\bmu}^u_{k}, \sum_{u \in B} \left(\frac{{\bpi}^u_{k}}{\sum_{u \in B}{\bpi}^u_{k}}\right)^2 {\bsigma}^2_{uk}\right)
    \end{equation}

    Given the assumption of viewpoint sufficiency \ref{ass:viewpoint_sufficiency} we know the objects observed in viewpoint set $A$ are same as the object observed in set $B$.
    Following the results of Theorem \ref{thm:affine_equivalence}, we know that both $p_A(\rvs)$ and $p_B(\rvs)$ are independently identifiable up to $\sim_s$ equivalence, which means $f_A$ and $f_B$ are invertible for a given views $\rvv^A$ and $\rvv^B$ respectively.

    \paragraph{Affine mapping.} Without loss of generality, for a given set of views $\rvv^A$, there exists some $L \subseteq \gX^A$ be an $|\gS|$-dimensional affine subspace, such that $B(\rvx'^A, \delta) \cap f_{A\sharp}(\gC,\{\rvv^A\}) \cap f_{B\sharp}(\gC,\{\rvv^A\}) = B(\rvx'^A, \delta) \cap L$. This implies their exists an affine map between $\rvc = f_A^{-1}(\rvx^A; \rvv^A)$ and $\tilde{\rvc} = f_B^{-1}(\rvx^B; \rvv^A)$. Let $h_A: \gC \rightarrow L$ to be an invertible affine functions where $h^{-1}_{A\sharp}(B(\rvx'^A, \delta')\cap L) = f^{-1}_{A\sharp} (B(\rvx'^A, \delta')\cap L; \rvv^A) = f^{-1}_{B\sharp} (B(\rvx'^B, \delta')\cap L; \rvv^A)$ resulting in $h_A(\rvc) = \rvc'$. 
    Similarly, we can show their exists an affine map between $\tilde{\rvc} = f_A^{-1}(\rvx^A; \rvv^B)$ and $\tilde{\rvc}' = f_B^{-1}(\rvx^B; \rvv^B)$, such that $h_B(\tilde{\rvc}) = \tilde{\rvc}'$.

    \paragraph{Invariance setup.} In the case when representations are invariant, $p_A(\rvc)$ and $p_B(\rvc)$ are equally distributed, which means aggregate content domain in both cases are same or similar $\gC_A = \gC_B$.

    \begin{align}
        & \rvc' = h(\tilde{\rvc}') \\
        & \implies  h_A(\rvc) = (h \circ h_B)(\tilde{\rvc}) \\
        & \implies  \rvc = (h_A^{-1} \circ h \circ h_B)(\tilde{\rvc})
    \end{align}

    Given composition of affine maps is affine, we can consider the mapping $(h_A^{-1} \circ h \circ h_B)$ to be an affine, resulting in an $\sim_s$ equivalence between $f_A$ and $f_B$.

\end{proof}


\paragraph{Theorem \ref{theorem:equivariance}} (Approximate representational equivariance)
    For a given aggregate content $\rvc$, for any two views $\rvv, \tilde{\rvv} \sim p_A(\rvv)$, resulting in respective scenes $\rvx \sim p_A(\rvx \mid \rvv, \rvc)$ and $\tilde{\rvx} \sim p_A(\rvx \mid \tilde{\rvv}, \rvc)$, for any homeomorphic, monotonic  transformation $h_x \in \gH_x$ such that $h_x(\rvx) = \tilde{\rvx}$, their exists another homeomorphic and monotonic transformation $h_v \in \gH_v$ such that $\gH_v \subseteq \gH_x \subseteq \mathbb{R}^{\mathrm{dim}(\rvx)}$ and  $\rvv = h_v^{-1}\left(f^{-1}_d(h_x(\rvx); \rvc)\right)$.

\begin{proof}
    For a given view $\rvv$ and a mixing function $f_d$ that satisfy assumptions \ref{ass:weak_inj} and  is piecewise affine, from theorem \ref{thm:affine_equivalence} we know the latent view representations are identifiable up to $\sim_s$ equivalence for a given aggregate content vector. 
    We know that $p(\rvv)$ is expressed as GMM with a considered set of viewpoints, ideally learning each component for each viewpoint.
    $$p(\rvv) = \sum_{v = 1}^{|A|} \bpi^v \gN(\rvv; \bmu_v, \bsigma_v)$$

    Following similar arguments in Theorem \ref{thm:affine_equivalence} and \cite{kivva2022identifiability}, we can show that for a given content representation $\rvc$ the view distribution $p(\rvv)$ is identifiable up to affine transformation.
    This means, for any two considered models $f_d, \tilde{f}_d$, such that $f_{d\sharp}(\gV; \{\rvc\})$ and $\tilde{f}_{d\sharp}(\gV; \{\rvc\} )$ are equally distributed, then for any $\rvx^A \sim \gX$ with the corresponding content representations given by $\rvc$ the views $\rvv = f_d^{-1}(\rvx^v; \rvc), \; \rvv' = \tilde{f}_d^{-1}(\rvx^v; \rvc)$ are related in by an affine transformation $h(\rvv) = \rvv'$, results in:
    
    \begin{align}
        \sum_{v=1}^{|A|} {\bpi}^{v} h_{\sharp} \left(\gN(\rvv; \bmu_{v},\bsigma^2_{v})\right)
        \sim
        h_{\sharp}\left(\sum_{v=1}^{|A|} {\bpi}^{v}\gN(\rvv;\bmu_{v},\bsigma^2_{v})\right),
    \end{align}

    Without loss of generality we can consider any function $f: \gC \times \gV \rightarrow \gX$ is identifiable up to affine transformation, with this for given views $\rvv, \tilde{\rvv} \sim p(\rvv)$ and for any object representations $\rvc$, the resulting scenes are sampled by distributions learned with mixing function $f$ is given by $\rvx \sim p_f(\rvx \mid \rvc, \rvv), \tilde{\rvx} \sim p_f(\rvx \mid \rvc, \tilde{\rvv})$.
    As previously established for some affine transformation $h$, 
    \begin{align}
        h(\rvv) = f^{-1}(\tilde{\rvx}; \rvc) \implies \rvv = h^{-1} \left( f^{-1}(\tilde{\rvx}; \rvc) \right)
    \end{align}
    Given $h_x(\rvx) = \tilde{\rvx}$, when combined with above equation we know $\rvv = h^{-1} \left( f^{-1}(\rvx; \rvc)\right), \tilde{\rvv} = h'^{-1} \left( f^{-1}(h_x(\rvx); \rvc)\right)$, for some invertible affine transformations $h$ and $h'$.

    Given $h_x$ is homeomorphic and monotonic, and $f$ is piecewise linear, the inverse can be transferred resulting in $\tilde{\rvv} = h'^{-1} \left( \bar{h}_v(f^{-1}(\rvx; \rvc))\right)$, similarly we can also swap $h'^{-1}$ with $\bar{h}_v$, resulting in $\tilde{\rvv} =  \bar{h}_v \left( h'^{-1} \left(f^{-1}(\rvx; \rvc)\right)\right)$.
    Additionally combining the results from theorem \ref{thm:affine_equivalence} and \cite{kivva2022identifiability}, we know that $h'^{-1} \circ h$ is an affine transformation $\bar{h}$. 
    This results in:
     \begin{align}
        & \bar{h} = h'^{-1} \circ h \\
        & \implies  \tilde{\rvv} =  (\bar{h}_v \circ h \circ \bar{h}) \left(f^{-1}(\rvx; \rvc)\right) \\
        & \implies  \tilde{\rvv} = h_v (\rvv)
    \end{align}

    Given affine transformation preserves monotonicity and homeomophism, the resulting transformation $h_v \in \gH_v$ and $h_v \in \gH_x$, concluding the proof.

\end{proof}

\section{Experiments}
\label{appendix:exps}

\subsection{Toy Setting}
\label{appendix:toysetup}
\begin{figure*}[t]
    \centering
    \hfill
    \begin{subfigure}{.21\textwidth}
        \centering
        {\footnotesize $\mathrm{Run \ \#1}$} \\[1pt]
        \includegraphics[trim={50, 55, 0, 33.2},clip,width=.985\textwidth]{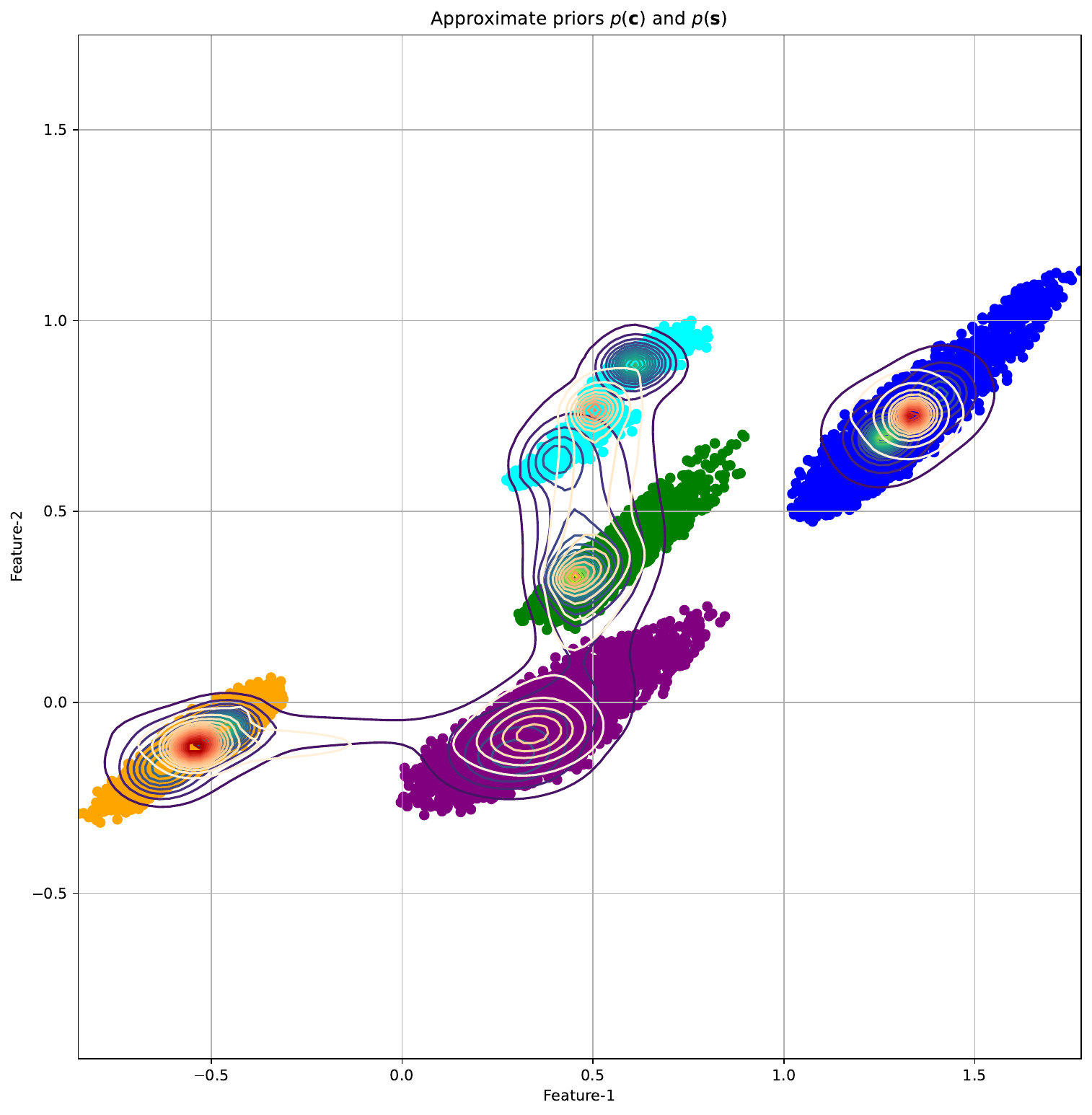}
        \vspace*{-18pt}
        \caption*{\footnotesize$\mathbf{z}_1$}
    \end{subfigure}
    \hfill
    \begin{subfigure}{.01\textwidth}
        \rotatebox{90}{\hspace{37pt} \footnotesize $\mathbf{z}_2$}
    \end{subfigure} 
    \begin{subfigure}{.21\textwidth}
        \centering
        {\footnotesize $\mathrm{Run \ \#2}$} \\[1pt]
        \includegraphics[trim={45, 55, 0, 33.2},clip,width=.985\textwidth]{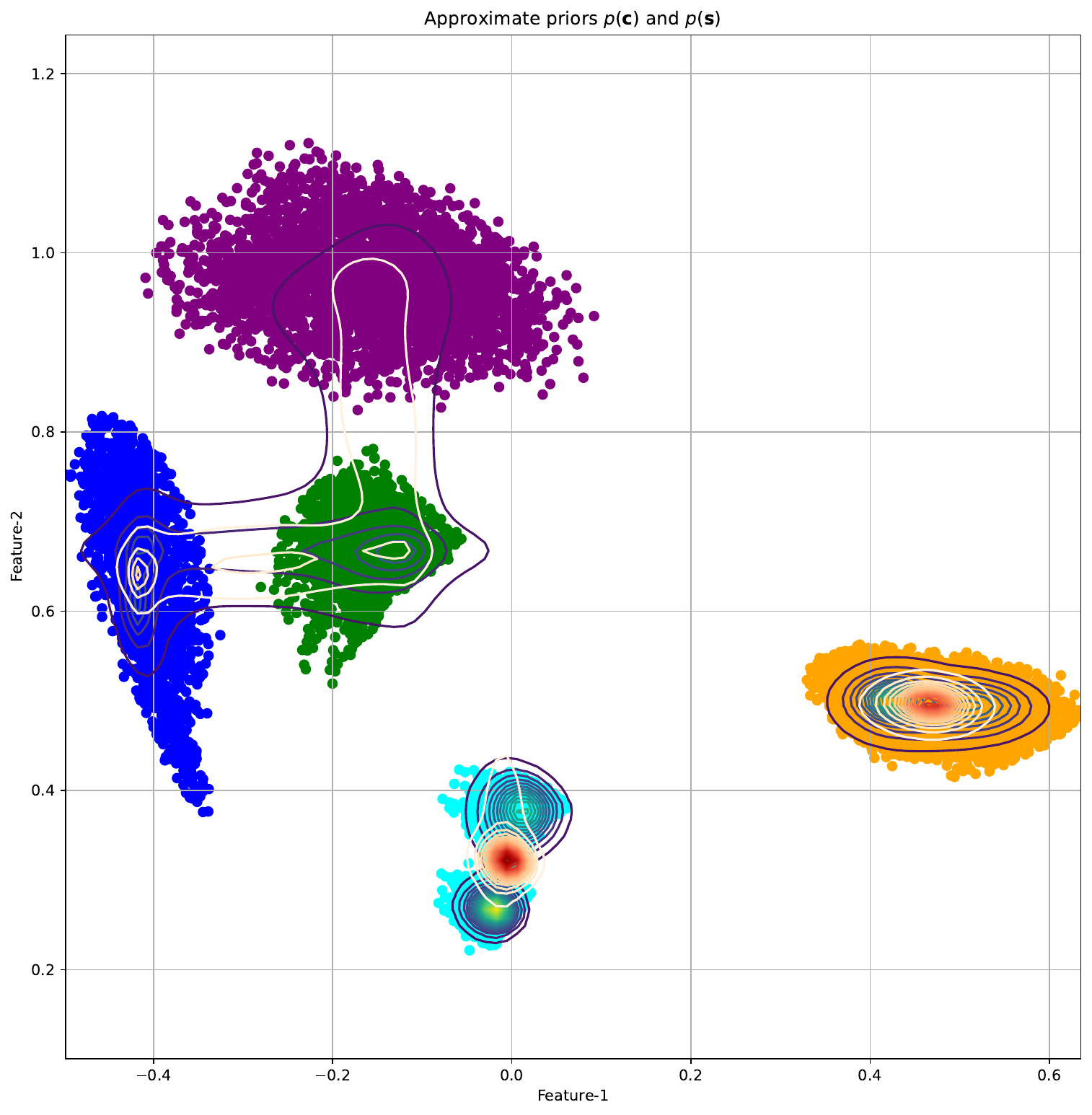}
        \vspace*{-18pt}
        \caption*{\footnotesize$\mathbf{z}_1$}
    \end{subfigure}
    \hfill
    \begin{subfigure}{.01\textwidth}
        \rotatebox{90}{\hspace{37pt} \footnotesize $\mathbf{z}_2$}
    \end{subfigure} 
    \begin{subfigure}{.21\textwidth}
        \centering
        {\footnotesize $\mathrm{Run \ \#3}$} \\[1pt]
        \includegraphics[trim={50, 55, 0, 33.2},clip,width=.985\textwidth]{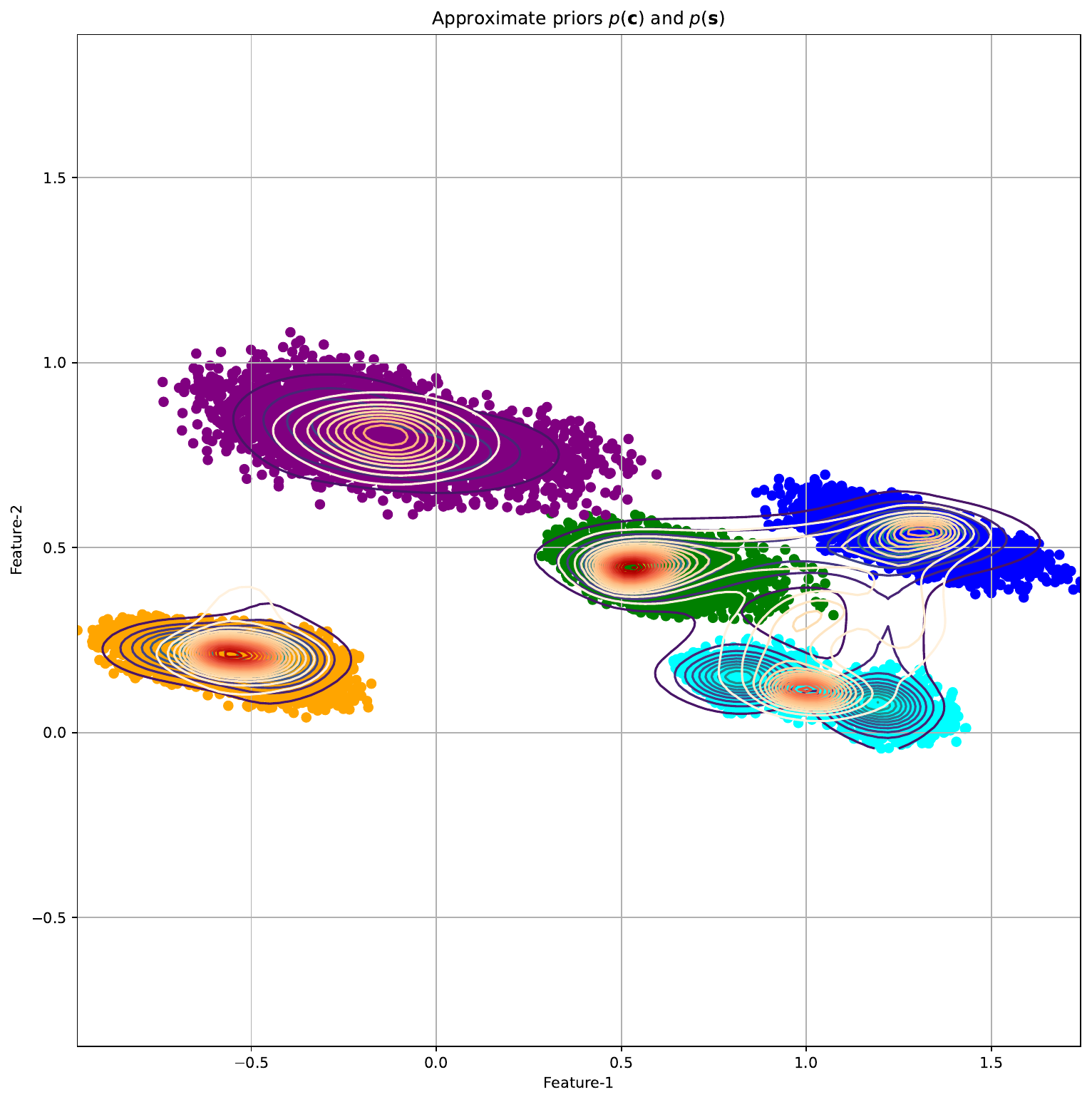}
        \vspace*{-18pt}
        \caption*{\footnotesize$\mathbf{z}_1$}
    \end{subfigure}
    \hfill
    \begin{subfigure}{.01\textwidth}
        \rotatebox{90}{\hspace{37pt} \footnotesize $\mathbf{z}_2$}
    \end{subfigure} 
    \begin{subfigure}{.21\textwidth}
        \centering
        {\footnotesize $\mathrm{Run \ \#4}$} \\[1pt] 
        \includegraphics[trim={50, 55, 0, 33.2},clip,width=.985\textwidth]{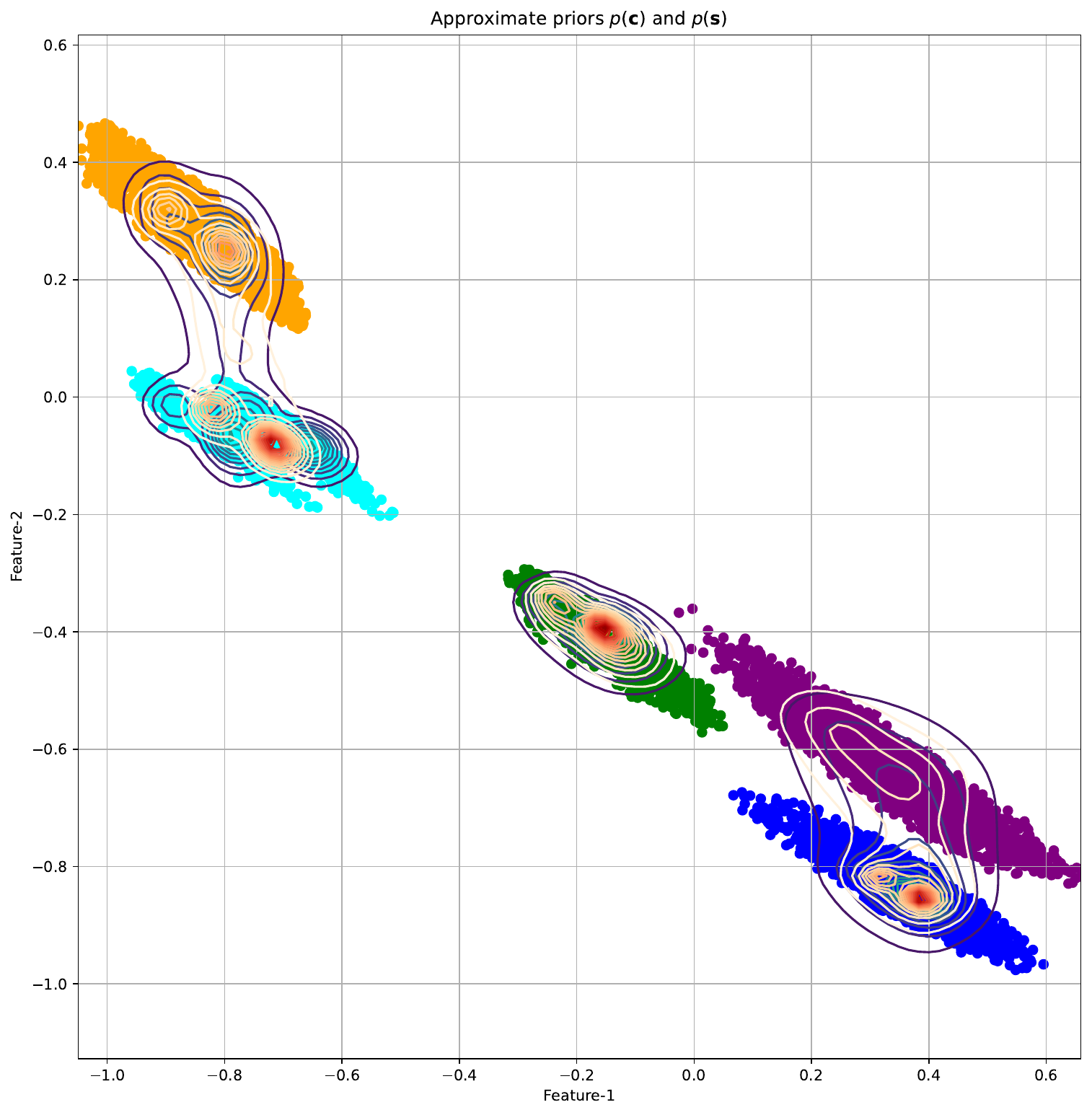}
        \vspace*{-18pt}
        \caption*{\footnotesize$\mathbf{z}_1$}
    \end{subfigure}
    \hfill
    \caption{\textbf{Identifiability of $q(\rvc)$ and $q(\rvs)$}. Estimated marginalised slot distribution ($q(\rvs)$--\textcolor{blue}{blue} contours) and marginalised content distribution ($q(\rvc)$--\textcolor{orange}{orange} contours, across 4 runs of \textsc{VISA}. 
    This provides strong evidence of recovery of the latent space up to affine transformations, empirically verifying our claims in Thm. \ref{thm:affine_equivalence}. 
    } \label{fig:identifiability_results_analysis}
\end{figure*}

\begin{figure*}[t]
    \centering
    \hfill
    \begin{subfigure}{.28\textwidth}
        \centering
        {\footnotesize $q_{A=\{\textcolor{red}{r}, \textcolor{green}{g}\}}(\rvc)$} \\[1pt]
        \includegraphics[trim={50, 55, 0, 33.2},clip,width=.985\textwidth]{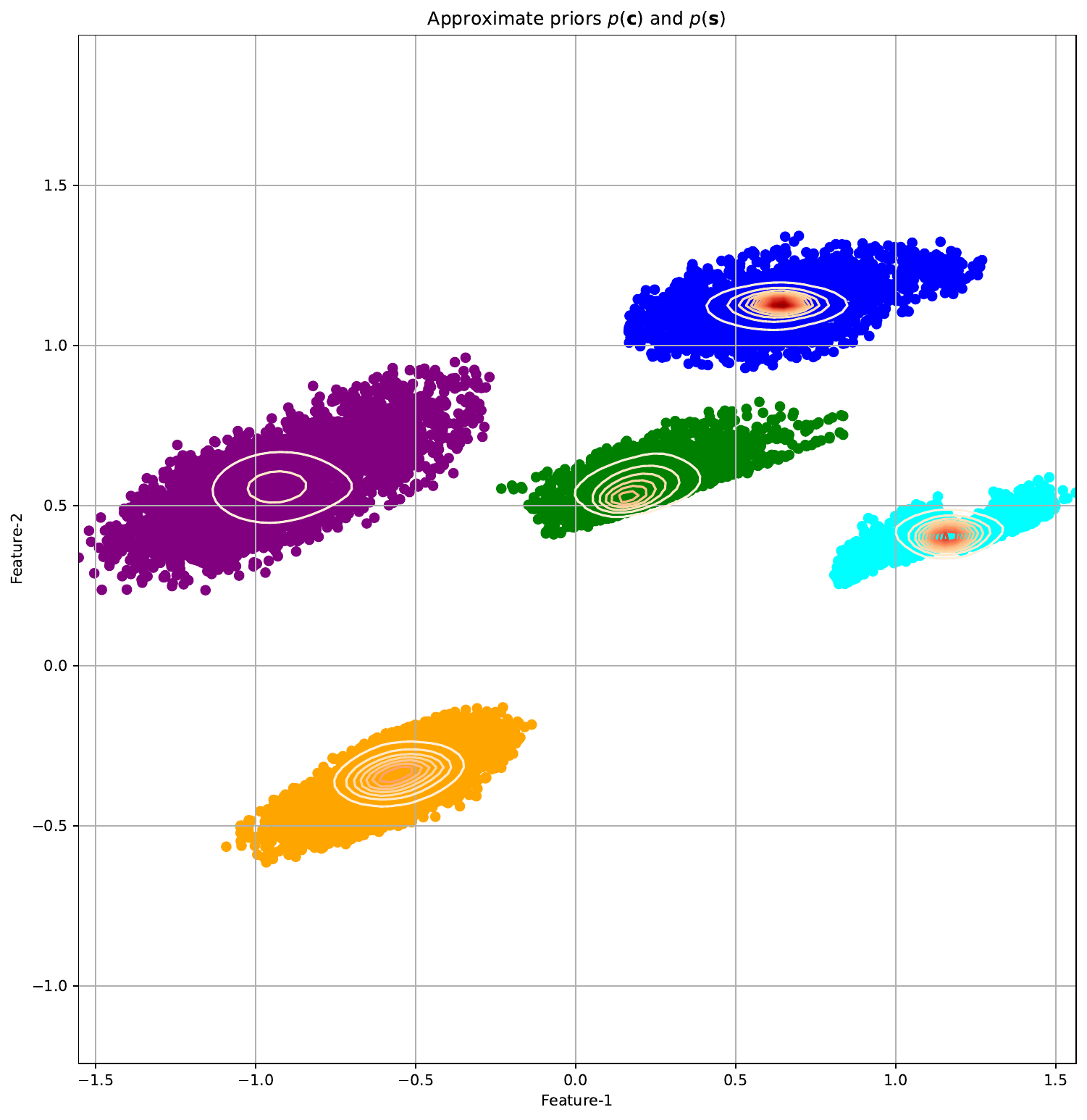}
        \vspace*{-18pt}
        \caption*{\footnotesize$\mathbf{z}_1$}
    \end{subfigure}
    \hfill
    \begin{subfigure}{.01\textwidth}
        \rotatebox{90}{\hspace{50pt} \footnotesize $\mathbf{z}_2$}
    \end{subfigure} 
    \begin{subfigure}{.28\textwidth}
        \centering
        {\footnotesize $q_{A=\{\textcolor{red}{r}, \textcolor{blue}{b}\}}(\rvc)$} \\[1pt]
        \includegraphics[trim={48, 55, 0, 33.2},clip,width=.985\textwidth]{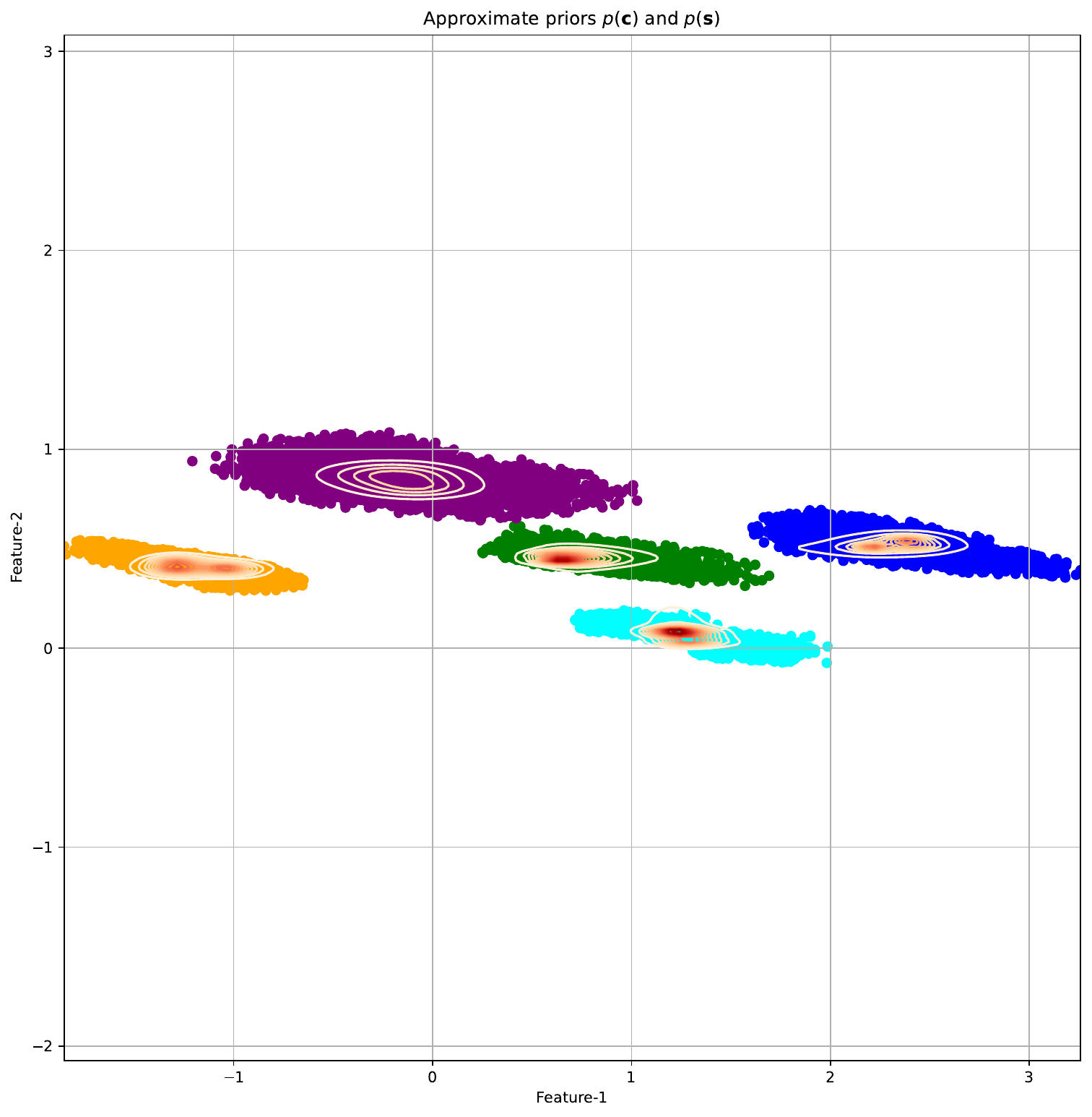}
        \vspace*{-18pt}
        \caption*{\footnotesize$\mathbf{z}_1$}
    \end{subfigure}
    \hfill
    \begin{subfigure}{.01\textwidth}
        \rotatebox{90}{\hspace{50pt} \footnotesize $\mathbf{z}_2$}
    \end{subfigure} 
    \begin{subfigure}{.28\textwidth}
        \centering
        {\footnotesize  $q_{A=\{\textcolor{blue}{b}, \textcolor{green}{g}\}}(\rvc)$} \\[1pt]
        \includegraphics[trim={50, 55, 0, 33.2},clip,width=.985\textwidth]{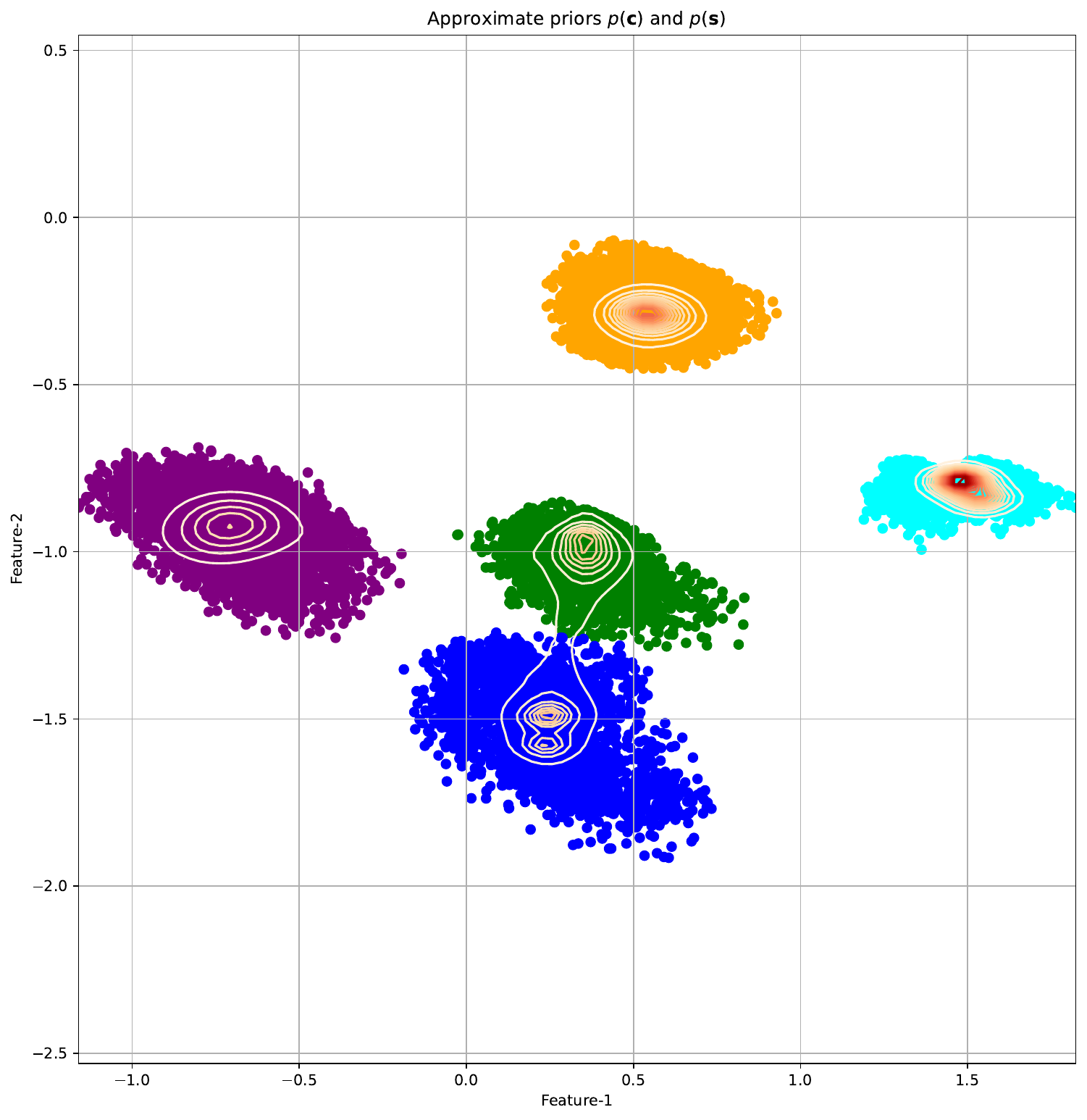}
        \vspace*{-18pt}
        \caption*{\footnotesize$\mathbf{z}_1$}
    \end{subfigure}
    \hfill
    \caption{\textbf{Viewpoint invariance for $q(\rvc)$}. Estimated  marginalised aggregate content distribution $q(\rvc)$ when trained with different view pairs $\{$(\textcolor{green}{green}, \textcolor{red}{red}), (\textcolor{red}{red}, \textcolor{blue}{blue}), (\textcolor{green}{green}, \textcolor{blue}{blue})$\}$ are illustrated in later figures. As the resulting distributions with different datasets only vary by an affine transformation, providing strong evidence for Thm. \ref{theorem:invariance}. 
    } 
    \label{fig:invariance_results_analysis2}
    \vspace{-15pt}
\end{figure*}

Here, we repeat the experiments in \textsc{case study 1} with point cloud giving us two dimensional distributions, which can analysed visually.
In Fig. \ref{fig:identifiability_results_analysis}, we display the distributions of marginalized aggregate content distribution $q(\rvc)$, comparing different runs that are either rotated, skewed, or mirrored with respect to each other, indicating identifiability up to affine transformation. 
To quantitatively measure the same, we computed SMCC and observed it to be $\mathbf{0.95} \pm 0.01$, empirically verifying our Thm. \ref{thm:affine_equivalence}. 
Furthermore, to illustrate the invariance of distribution $q(\rvc)$ across viewpoints (Thm. \ref{theorem:invariance}), we consider three different views.
We use all possible pairs to learn $q(\rvc)$ distributions as illustrated in Fig. \ref{fig:invariance_results_analysis2}, where the distributions from second to last sub-figures are learned wrt viewpoints described by $\{$\textcolor{green}{g}, \textcolor{red}{r}$\}$, $\{$\textcolor{red}{r}, \textcolor{blue}{b}$\}$, and $\{$\textcolor{green}{g}, \textcolor{blue}{b}$\}$,  respectively. 
Similar to our previous findings, these distributions were also found to be rotated, skewed, or mirrored relative to each other, with an observed SMCC of $\mathbf{0.87} \pm 0.11$, further confirming the claims in Thm. \ref{theorem:invariance}.

\subsection{Synthetic dataset results}
\begin{figure}[!h]
    \centering
    \subfloat[]{\includegraphics[width=.79\textwidth]{figures/cvAug-1.png}}\hfill
    \subfloat[]{\includegraphics[width=.79\textwidth]{figures/cvAug-2.png}}\hfill
    \subfloat[]{\includegraphics[width=.79\textwidth]{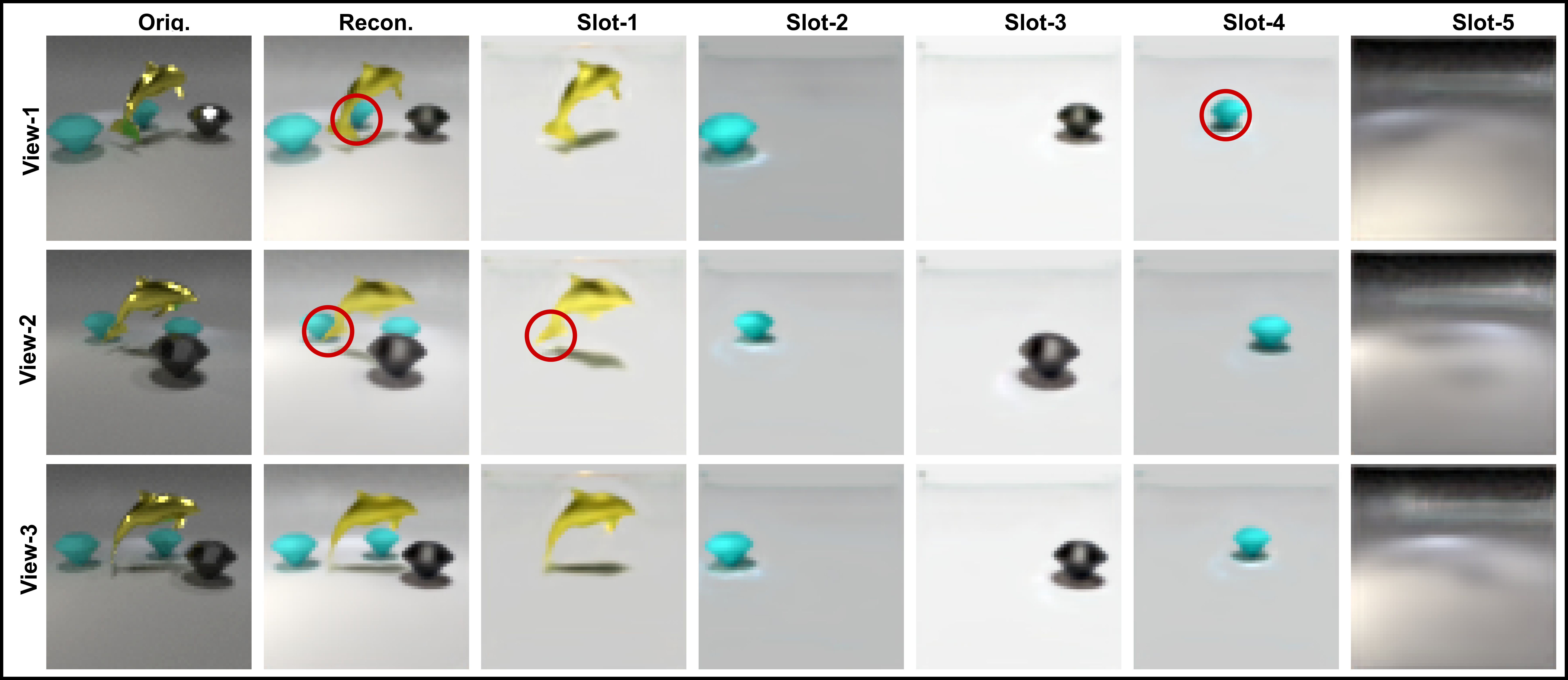}}\hfill
    \caption{Visual illustrations of benchmark results on \textsc{CLEVR-aug} dataset.}
    \label{fig:cvaug_results}
\end{figure}

Here, we illustrate visual results reflecting object binding in the case of view ambiguities.
Table \ref{table:clevr-aug_results}, demonstrates identifiability results on \textsc{CLEVR-aug} datasets.
In Fig. \ref{fig:cvaug_results}, we demonstrate the results of \textsc{VISA} across three different views.
We additionally highlight some of the occluded regions which seem to be better captured by our proposed model, which can be attributed to the multi-view setting and the \texttt{sigmoid} mask.

\begin{table*}[t]
\scriptsize
\centering
\caption{Comparing identifiability of $q(\rvs)$, $q(\rvc)$, and $p(\rvv)$ scores wrt existing OCL methods on \textsc{CLEVR-aug} dataset.}
\begin{tabular}{@{}lccc@{}}
\toprule
\textsc{Method}  
            & SMCC $\uparrow$ & \textsc{inv-SMCC} $\uparrow$ & MCC $\uparrow$ \\ 
\midrule
AE             & $0.26 \pm .01$ & - & - \\ 
SA             & $0.45 \pm .05$ & - & - \\ 
PSA            & $0.48 \pm .03$ & - & - \\ 
MulMON         & $0.56 \pm .04$ & $0.57 \pm .01$ & - \\
OCLOC          & $0.58 \pm .02$ & $0.60 \pm .01$ & $0.48 \pm .04 $
\\
\midrule
\textbf{\textsc{VISA}} & $0.64 \pm .01$ & $0.66 \pm .01$ & $0.57 \pm .04$ \\
\bottomrule
\end{tabular}
\label{table:clevr-aug_results}
\end{table*}

\begin{figure}[!h]
    \centering
    \subfloat[]{\includegraphics[width=.79\textwidth]{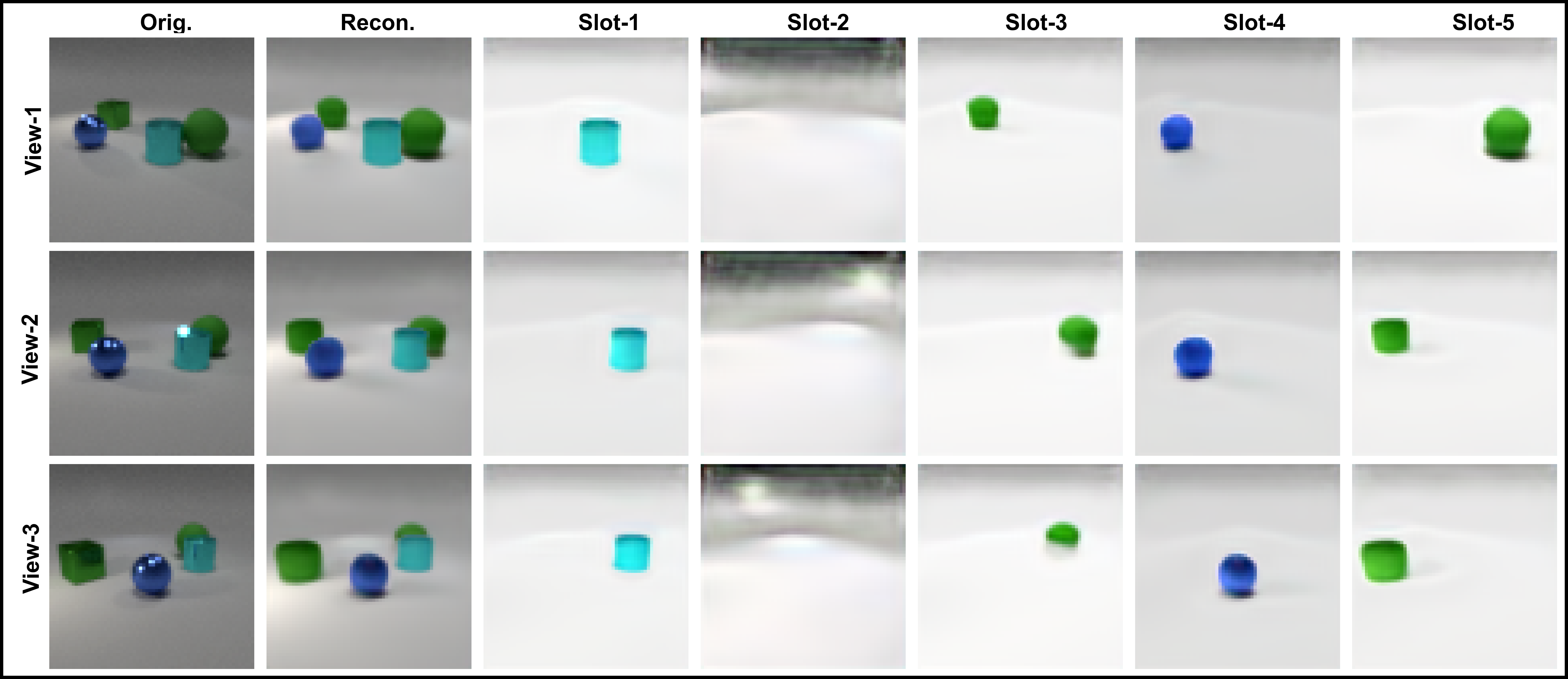}}\hfill
    \subfloat[]{\includegraphics[width=.79\textwidth]{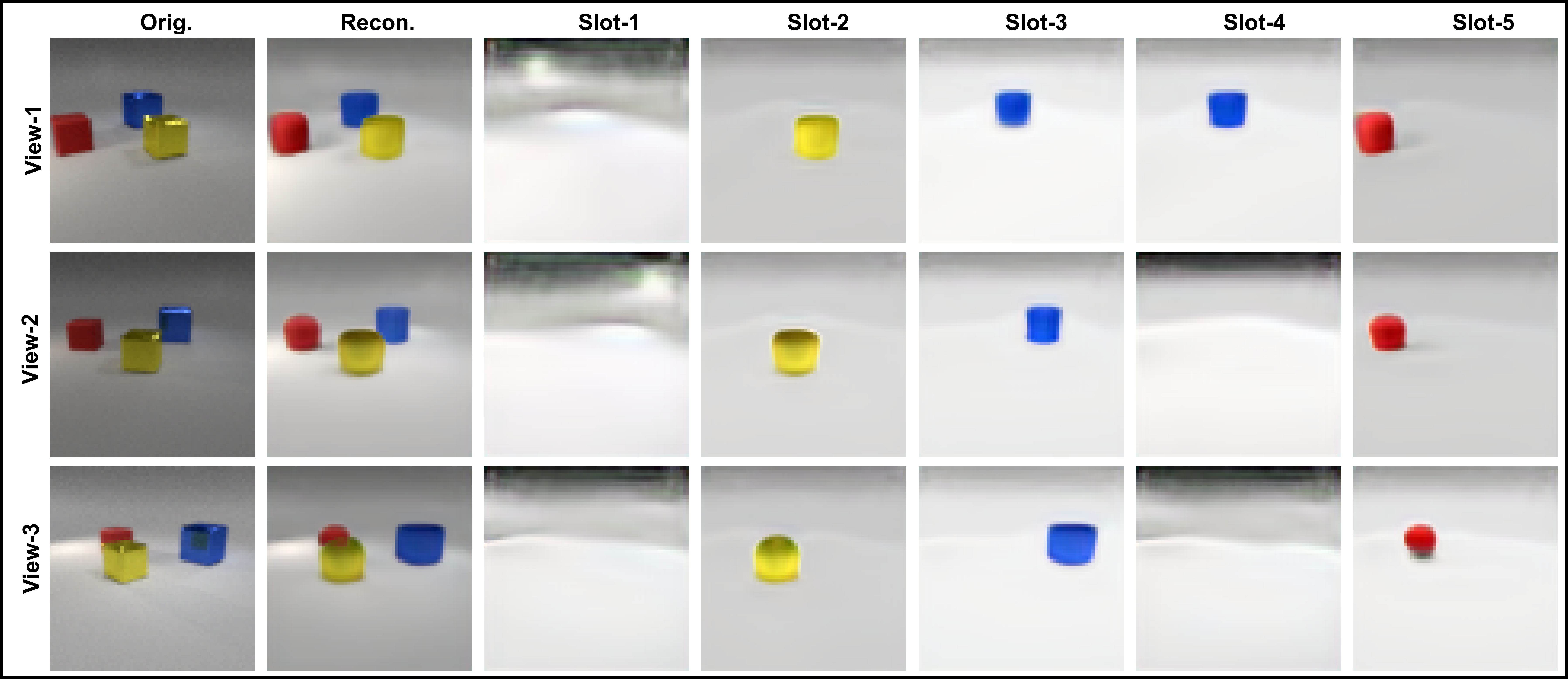}}\hfill
    \subfloat[]{\includegraphics[width=.79\textwidth]{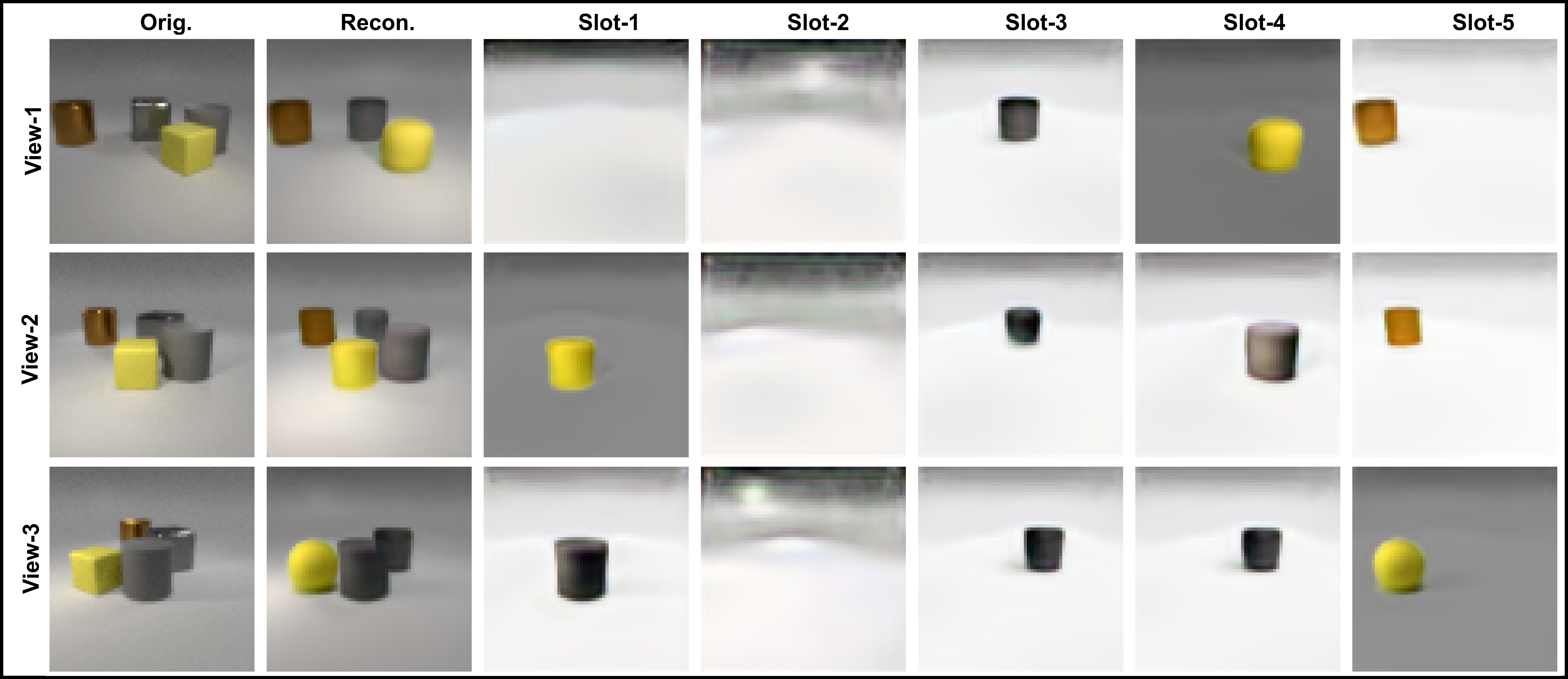}}\hfill
    \caption{Visual illustrations of benchmark results on \textsc{CLEVR-mv} dataset.}
    \label{fig:cvmv_results}
\end{figure}

Additionally, we also illustrate the results from \textsc{CLEVR-mv} dataset in figure \ref{fig:cvmv_results}. 

\subsection{Influence of Number of Views}

\begin{figure*}[h]
    \centering
    \subfloat{\includegraphics[width=.325\textwidth]{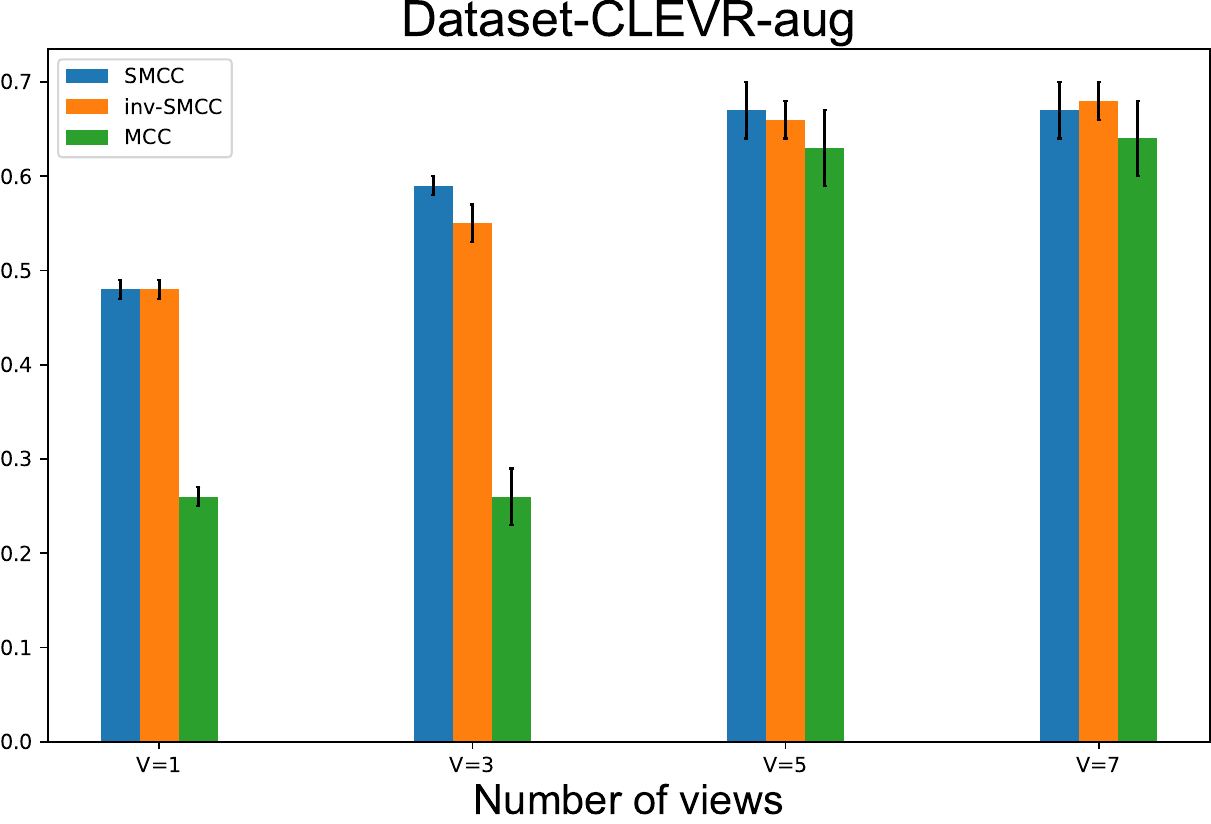}}\hfill
    \subfloat{\includegraphics[width=.325\textwidth]{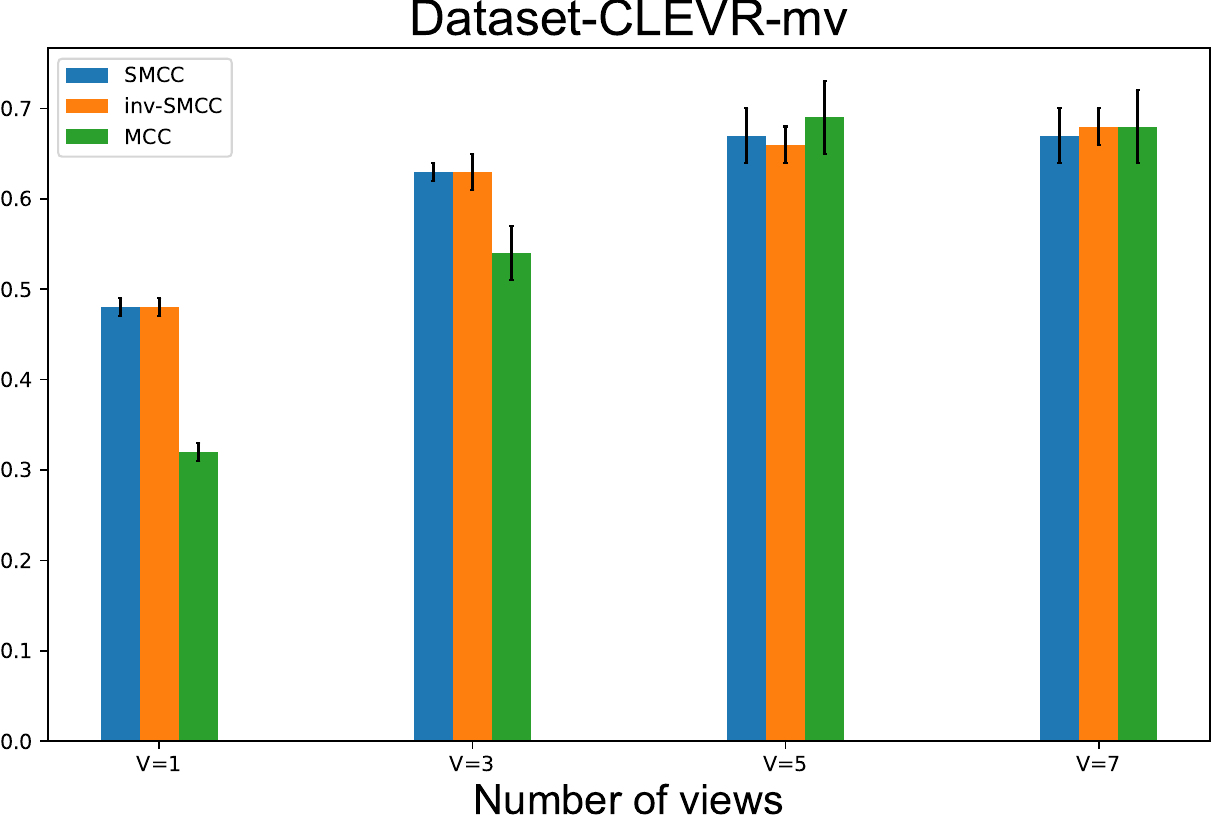}}\hfill
    \subfloat{\includegraphics[width=.325\textwidth]{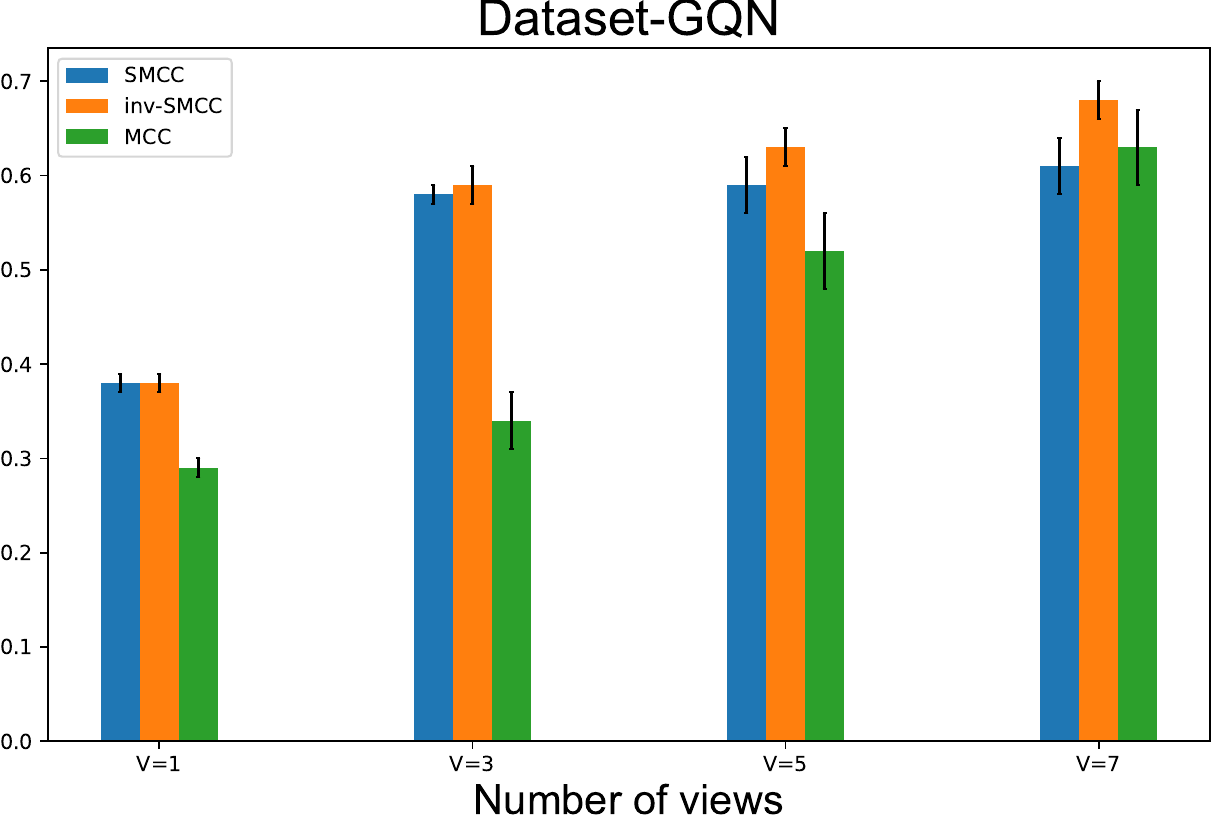}}\hfill
    \caption{Influence of Number of viewpoints on identifiability for synthetic datasets.}
    \label{fig:|A|_identifiability}
\end{figure*}

Here, we demonstrate the influence of the number of views on the overall identifiability of object-centric representations. Similar to Fig. \ref{fig:motivation2}, in Fig. \ref{fig:|A|_identifiability}, we observe an increasing number of views increase overall results.

\subsection{\textsc{mvMoVi} Results}

Here, we discuss the results obtained from the proposed dataset. 
To reiterate, \textsc{mvMoVi-C} is a variant where fixed camera positions are maintained for all viewpoints across all scenes in the dataset. 
This setup helps assign a fixed type of viewpoint conditioning for all images captured from a particular camera. 

The detection and binding quality of different models are illustrated in Table \ref{table:movi-results}. 
From these results, we can clearly observe that while the model demonstrates similar binding capabilities, the identifiability of object representations is improved in our proposed model. 
This suggests that the use of fixed camera positions in \textsc{mvMoVi-C} enhances the consistency and quality of object representation learning, leading to better detection performance across different viewpoints.

Figure \ref{fig:mvC_results-v2} \& \ref{fig:mvC_results-v3} showcases the object discovery capabilities of the \textsc{VISA}.
In the iteration of the \textsc{mvMoVi-D} dataset, we vary the camera position for each scene, making the dataset more dynamic and allowing for the potential violation of assumption \ref{ass:viewpoint_sufficiency} in certain cases. Table \ref{table:moviD-results} presents the binding and identifiability results for both in-domain and out-of-domain data, following a similar analysis as in Table \ref{table:movi-results}. 
We observe consistent trends and behaviours, suggesting that the impact of the assumption is minimal. A more detailed analysis of the assumption's effects will be left for future work.

\begin{figure*}[!h]
    \centering
    \subfloat[]{\includegraphics[width=.79\textwidth]{figures/mvC-1.png}}\hfill
    \subfloat[]{\includegraphics[width=.79\textwidth]{figures/mvC-2.png}}\hfill
    \caption{Visual illustrations of benchmark results on \textsc{mvMoVi-C} dataset with 2 views.}
    \label{fig:mvC_results-v2}
\end{figure*}

\begin{figure*}[!h]
    \centering
    \subfloat[]{\includegraphics[width=.79\textwidth]{figures/mvC-3.png}}\hfill
    \subfloat[]{\includegraphics[width=.79\textwidth]{figures/mvC-4.png}}\hfill
    \caption{Visual illustrations of benchmark results on \textsc{mvMoVi-C} dataset with 3 views.}
    \label{fig:mvC_results-v3}
\end{figure*}


\begin{table}[h]
\scriptsize
\centering
\caption{Identifiability and generalisability analysis on \textsc{mv-MoViD} dataset.}
\resizebox{\columnwidth}{!}{
\begin{tabular}{@{}lcccc|cccc@{}}
\toprule
\textsc{Method} & \multicolumn{4}{c}{\textsc{Indomain Analysis}} & \multicolumn{4}{c}{\textsc{Out of domain}} \\ 
\cmidrule(l){2-9} 
& mBO $\uparrow$ & SMCC $\uparrow$   & \textsc{inv-SMCC} $\uparrow$ 
 & MCC $\uparrow$ & mBO $\uparrow$ & SMCC $\uparrow$   & \textsc{inv-SMCC} $\uparrow$ 
 & MCC $\uparrow$ \\ 
\midrule
\textsc{SA-MLP}             & $0.24 \pm 0.031$
                            & $0.44 \pm 0.005$
                            & -
                            & -
                            & $0.24\pm 0.097$
                            & $0.45 \pm 0.008$
                            & -
                            & -

\\

\textsc{PSA-MLP}            & $0.26\pm 0.022$
                            & $0.44 \pm 0.006$
                            & -
                            & -
                            & $0.25 \pm 0.012$
                            & $0.42 \pm 0.006$
                            & -
                            & -
\\
\midrule
\textsc{VISA-MLP}          & $0.24 \pm 0.099$
                            & $0.48 \pm 0.009$
                            & $0.46 \pm 0.054$
                            & $0.57 \pm 0.021$
                            & $0.25 \pm 0.011$
                            & $0.48 \pm 0.006$
                            & $0.51 \pm 0.021$
                            & $0.55 \pm 0.021$
\\
\midrule
\textsc{SA-Transformer}     &  $0.34 \pm 0.017$     
                            & $0.40 \pm 0.041$
                            & -
                            & -
                            & $0.34 \pm 0.066$
                            & $0.38 \pm 0.031$
                            & -
                            & -
                            \\
\textsc{PSA-Transformer}    & $0.37 \pm 0.021$     
                            & $0.38 \pm 0.007$
                            & -
                            & -
                            & $0.36 \pm 0.024$
                            & $0.36 \pm 0.016$
                            & - 
                            & -
                            \\
\midrule
\textbf{\textsc{VISA-Transformer}}  &   $0.39 \pm 0.016$    
                            & $0.46 \pm 0.001$
                            & $0.48 \pm 0.001$
                            & $0.54 \pm 0.032$
                            & $0.37 \pm 0.051$
                            & $0.46 \pm 0.022$
                            & $0.45 \pm 0.010$
                            & $0.54 \pm 0.029$
                            \\
\bottomrule
\end{tabular}
}
\label{table:moviD-results}
\end{table}

\subsection{View warm-up}
\label{app:viewwarmup}
Given the stochasticity during the initial phase of training, to facilitate meaningful representation in content aggregator function, we consider view warm-up strategy.  
For the initial 100,000 iterations, we randomly use the view-specific slots for reconstruction instead of invariant content with a probability of 0.5.

This primary makes sure the feature extractor extracts meaningful representations before aggregation, which helps to stabilize the training process and allows the model to effectively bind and integrate information from different perspectives in later stages of training.

\subsection{Hyperparameters}
In Table \ref{tab:hparams}, we detail all the hyper-parameters used in our experiments. 
In the case of benchmark experiments, we use trainable CNN encoder as used in \cite{locatello2020object, kori2023grounded}, while in the case of proposed \textsc{mvMoVi} datasets we use DINO \citep{caron2021emerging} encoder to extract image features and change our objective to reconstruct these features rather than the original image as proposed in \cite{seitzer2022bridging}.
For most of hyperparameters we use the values suggested by \cite{locatello2020object, seitzer2022bridging}, based on their ablation results.

\renewcommand{\arraystretch}{1.} 
\begin{table}
    \caption{Experimental details w.r.t datasets}
    \centering
        \begin{tabular}{@{}llc@{}}
            \toprule
            \textsc{Datasets}($\downarrow$)
             & \textsc{Parameters} & \textsc{Values} 
        \\ \cmidrule(r){1-3}
        
        \multirow{12}{*}{\textsc{CLEVR},
        \textsc{GSO}}
             & No. Layers             & 4 \\
             & No. Views              & 10 (\textsc{GSO: 8)} \\
             & No. Slots              & 7   \\   
             & Training Epochs        & 5000 \\
             & Batch Size             & 32 \\
             & Optimizer              & \textsc{Adam} \\
             & Learning Rate          & $0.0002$ \\
             & Initial Slot $\bmu$    & $\gN(0, 1)$ \\
             & Initial Slot $\bsigma$ & $\mathbb{I}$ \\
             & Warmup Steps           & $10000$ \\
             & Decoder                & \textsc{Spatial Broadcasting-CNN} \\ 
             & $\rvx-$ likelihood     & $\gN(\bmu_x, \sigma^2_x \mathbb{I})$ \\
             
        \midrule
             
        \multirow{12}{*}{\textsc{GQN}}
             & No. Layers             & 4 \\
             & No. Views              & 10 \\
             & No. Slots              & 4   \\   
             & Training Epochs        & 5000 \\
             & Batch Size             & 64 \\
             & Optimizer              & \textsc{Adam} \\
             & Learning Rate          & $0.0002$ \\
             & Initial Slot $\bmu$    & $\gN(0, 1)$ \\
             & Initial Slot $\bsigma$ & $\mathbb{I}$ \\
             & Warmup Steps           & $10000$ \\
             & Decoder                & \textsc{Spatial Broadcasting-CNN} \\ 
             & $\rvx-$ likelihood     & $\gN(\bmu_x, \sigma^2_x \mathbb{I})$ \\
        \midrule             
        \multirow{13}{*}{\textsc{mvMoVi-C}, \textsc{mvMoVi-D}}
             & No. Layers             & 4 \\
             & No. Views              & 5 \\
             & No. Slots              & 7   \\   
             & Training Epochs        & 560 \\
             & Batch Size             & 64 \\
             & Optimizer              & \textsc{AdamW} \\
             & Learning Rate          & $0.0002$ \\
             & Initial Slot $\bmu$    & $\gN(0, 1)$ \\
             & Initial Slot $\bsigma$ & $\mathbb{I}$ \\
             & Warmup Steps           & $10000$ \\
             & Pretrained Encoder     & \textsc{dino\_vitb16} \\
             & Decoder                & \textsc{MLP}, \textsc{Transformer} \\ 
             & $\rvx-$ likelihood     & $\gN(\bmu_x, \mathbb{I})$ \\
     
        \bottomrule
        \end{tabular}
    \label{tab:hparams}
\end{table}

\subsection{Computational Resources}
We run all our experiments on a cluster with a Nvidia NVIDIA L40 48GB GPU cards. 
Our training usually takes between eight hours to a couple of days, depending on the model and the dataset. 
It is to be noted that speed might differ slightly with respect to the considered system and the background processes. 
All experimental scripts will be made available on GitHub at a later stage.
\end{document}